\documentclass[12pt,a4paper,%twoside=true,%,openany,% twoside=true zum drucken!
BCOR=12mm,%
DIV=14,%
headsepline,%
cleardoublepage=empty,%
chapterprefix,%
parskip,%
listof=totoc,%
index=totoc,%
bibliography=totoc]{scrbook}

\usepackage[ngerman, english]{babel} % Englisch

\PassOptionsToPackage{hyphens}{url}
\usepackage[utf8]{inputenc} % erlaubt direkte Eingabe von Unicode-Zeichen, insbesondere Umlaute
\usepackage[T1]{fontenc}
\usepackage[babel]{csquotes}
\usepackage[style=alphabetic, maxbibnames=999, backend=bibtex, backref=true]{biblatex}
\usepackage{mathptmx}
\usepackage[scaled=0.92]{helvet}
\usepackage{courier}
\usepackage[calc]{datetime2}
\usepackage{amsmath}

\usepackage{amsfonts}
\usepackage{amstext}
\usepackage{subfigure}
\usepackage{color}
\usepackage{graphicx}
\usepackage{ifpdf}
\usepackage{iflang}
\usepackage{makeidx}
\usepackage{nomencl}
\usepackage{setspace}
\usepackage{lscape}
\usepackage{tabularx}
\usepackage{booktabs}
\usepackage{todonotes}
\usepackage{listings}
\usepackage[%
    binary-units=true,
    per-mode=symbol,
    exponent-product=\cdot,
    range-units=single,
    exponent-to-prefix=true,
    zero-decimal-to-integer=true,
    list-units=single
    ]{siunitx}

\usepackage[plainpages=false,pdfpagelabels]{hyperref}
\usepackage[german]{cleveref}

\usepackage{xcolor}
\usepackage{soul}

\usepackage{algorithm}
\usepackage{algpseudocode}
\usepackage{longtable}

\def\mauve{0}
\def\graffi{1}

\newcommand{\graduationlevel}{A Master}

\newcommand{\professor}{\mauve}
\newcommand{\thesistitle}{Linguistic Structure Induction from Language Models}

\newcommand{\thesisauthor}{Omar Momen}

\newcommand{\thesisauthorbirthplace}{Cairo, Egypt}

\newcommand{\thesiskeywords}{}

\newcommand{\thesissubmissionday}{29}

\newcommand{\thesissubmissionmonth}{02}

\newcommand{\thesissubmissionyear}{2024}

\IfLanguageName{ngerman}{
    \newcommand{\thesistype}{\graduationlevel{}arbeit}
}{
    \newcommand{\thesistype}{\graduationlevel{}'s Thesis}
}

\ifpdf
    \definecolor{brown}{cmyk}{0, 0.81, 1, 0.60}
    \hypersetup{%
      pdftitle={\thesistitle{}},
      pdfsubject={\thesistype{}},
      pdfauthor={\thesisauthor{}},
      pdfkeywords={\thesiskeywords{}},
      colorlinks=false, urlcolor=blue, citecolor=brown,
      bookmarksnumbered=true,
    }
\else   
\fi

\DefineBibliographyStrings{ngerman}{%
    mathesis = {Masterarbeit},
    bathesis = {Bachelorarbeit},
    phdthesis = {Dissertation}
}

\DefineBibliographyStrings{english}{%
    mathesis = {Master's Thesis},
    bathesis = {Bachelor's Thesis},
    phdthesis = {PhD Thesis}
}

\makeindex

\renewbibmacro*{doi+eprint+url}{%
  \iftoggle{bbx:doi}
    {\printfield{doi}}
    {}%
  \newunit\newblock
  \iftoggle{bbx:eprint}
    {\usebibmacro{eprint}}
    {}%
  \newunit\newblock
  \iffieldundef{doi}{% Only show URL if DOI is undefined
    \iftoggle{bbx:url}
      {\usebibmacro{url+urldate}}
      {}}
    {}%
}

\RedeclareSectionCommand[
  beforeskip=-20pt, % Reduces the space above the chapter title
  afterskip=10pt, % Reduces the space above the chapter title
]{chapter}

\RedeclareSectionCommand[
  beforeskip=-1.8pt, % Reduces the space above the chapter title
  afterskip=1.8pt, % Reduces the space above the chapter title
]{section}

\RedeclareSectionCommand[
  beforeskip=-1.8pt, % Reduces the space above the chapter title
  afterskip=1.8pt, % Reduces the space above the chapter title
]{subsection}

\newcolumntype{T}[1]{>{\centering\arraybackslash}m{#1}}
\newcolumntype{M}[1]{>{\centering\arraybackslash}m{#1}}

\usepackage{mathtools}
\newcommand\Set[2]{\{\,#1\mid#2\,\}}

\begin{document}
\onehalfspacing

\frontmatter

% do not change this file
\pdfbookmark[0]{\IfLanguageName{ngerman}{Titelseite}{Front Page}}{frontpage}
\begin{titlepage}
    
%     \begin{tabular}{lT}
%         \includegraphics[width=7.8cm]{static/Logo_HHU_+Name_horizontal_4c_+Safezone} & 
%         \textsf{\textbf{\large INSTITUT FÜR INFORMATIK}}\\
%         & \textsf{\large Universitätsstr. 1 \hspace{1em} D-40225 Düsseldorf} 

% \\
%     \end{tabular}
    
%     % \begin{flushleft}
%     %     \begin{figure}[ht]
%     %         % \centering
            
%     %     \end{figure}
%     % \end{flushleft}
  % \vspace{-2em}
  \centering
  \includegraphics[width=0.9\linewidth]{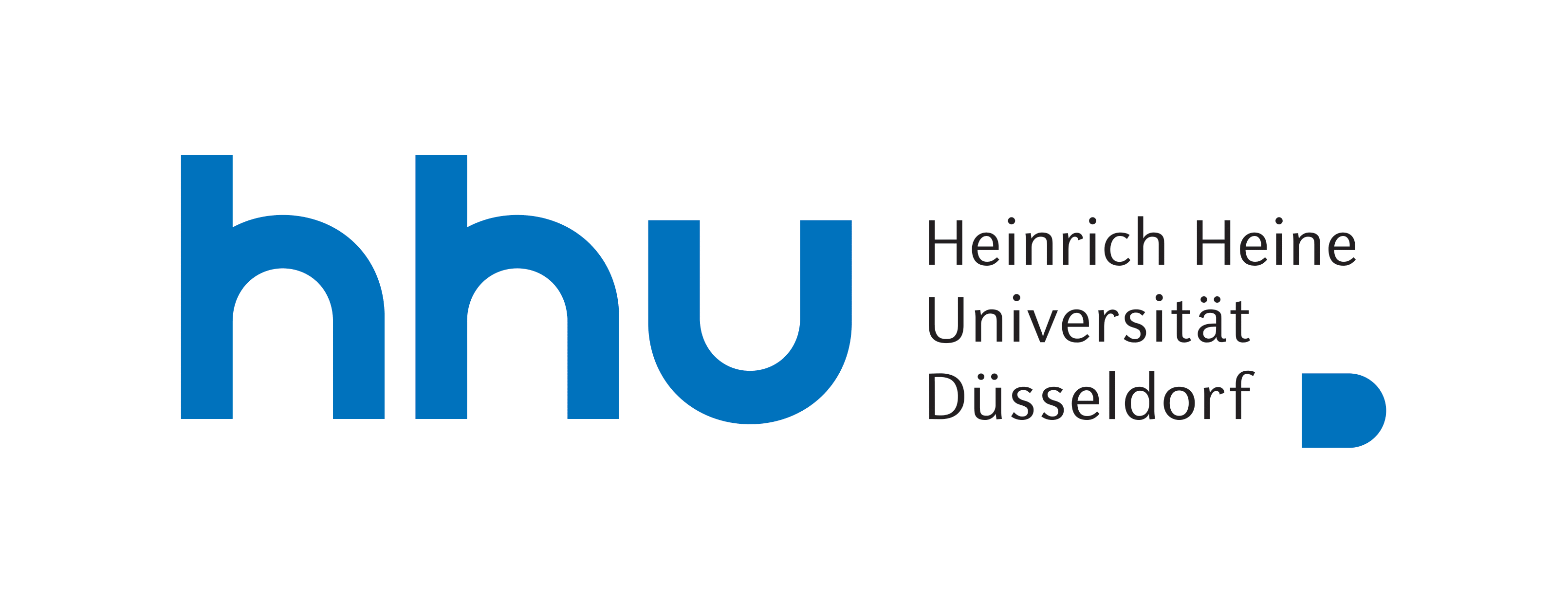}\\
  \vspace{-1.5em}
  \textsf{{\large INSTITUT FÜR INFORMATIK}}\\
  \textsf{\large Universitätsstr. 1 \hspace{1em} D-40225 Düsseldorf}

  \vfill
  \huge
  \thesistitle{}\\*[45pt]
  \normalsize

  \vfill
  \large
  \thesistype{}\\[0.25em]
  \normalsize
  \IfLanguageName{ngerman}{von}{by}\\
  \Large
  \thesisauthor{}\\

  \vspace{5mm}
  \normalsize
  \IfLanguageName{ngerman}{aus}{from}\\
  \thesisauthorbirthplace{}\\[1cm]

 \if\professor\mauve
    \IfLanguageName{ngerman}{vorgelegt am}{Submitted in partial fulfillment of the requirements for the Master of Science in Artificial Intelligence and Data Science}\\[5mm]

    \vspace{2em}
    
    \IfLanguageName{ngerman}{Lehrstuhl für Rechnernetze}%{Professorship for Computational Linguistics}\\
    %Prof.\ Dr.\ Martin Mauve\\
    \begin{tabular}{lc}
        First Reviewer: & Prof.\ Dr.\ Laura Kallmeyer \\
        Second Reviewer: &   David Arps, M.\,A.
    \end{tabular}

  \else
    \if\professor\graffi
      \IfLanguageName{ngerman}{vorgelegt bei}{submitted to}\\[5mm]
      \IfLanguageName{ngerman}{Technik Sozialer Netzwerke}{Technology of Social Networks Lab}\\
      Jun.-Prof.\ Dr.-Ing.\ Kalman Graffi\\
    \fi
  \fi

    \vspace{2em}
  % \IfLanguageName{ngerman}{Heinrich-Heine-Universität Düsseldorf}{Heinrich-Heine-University Düsseldorf}\\[0.5cm]
  \thesissubmissionday \hspace{0.1em} \DTMmonthname{\thesissubmissionmonth} \thesissubmissionyear{}\\[0.5cm]
  % \IfLanguageName{ngerman}{Betreuer}{Supervisor}:\\
  % \thesissupervisor{}
\end{titlepage}

\cleardoublepage

% do not change this
\pdfbookmark[0]{\IfLanguageName{ngerman}{Zusammenfassung}{Abstract}}{abstract}
\begin{center} 
    \huge \IfLanguageName{ngerman}{Zusammenfassung}{Abstract}
\end{center}

Linear sequences of words are implicitly represented in our brains by hierarchical structures that organize the composition of words in sentences. Linguists formalize different frameworks to model this hierarchy; two of the most common syntactic frameworks are Constituency and Dependency. Constituency represents sentences as nested groups of phrases, while dependency represents a sentence by assigning relations between its words. Recently, the pursuit of intelligent machines has produced Language Models (LMs) capable of solving many language tasks such as translation, sentiment analysis, and text generation with a human-level performance. Many studies now question whether LMs implicitly represent syntactic hierarchies. To this end, some studies assess LMs outputs against syntactic rules, while others probe LMs representations for syntactic knowledge evidence. This thesis instead focuses on producing constituency and dependency structures from LMs in an unsupervised setting. We review the critical methods in this field and highlight a line of work that utilizes a numerical representation for binary constituency trees (Syntactic Distance). We present a detailed study on StructFormer~\cite{shen-etal-2021-structformer}, which retrofits a transformer encoder architecture with a parser network to produce constituency and dependency structures. We present six experiments to analyze and address this field's challenges: 1) Reproduce the results of StructFormer, 2) Verify the consistency of its induced trees, 3) Investigate the effect of repositioning the parser network within the StructFormer architecture, 4) Evaluate subword-based induced trees against word-based reference trees, 5) Evaluate a subword-based model variant pretrained on a new dataset, and 6) Benchmark the models developed in these experiments on linguistic tasks. Models benchmarking is performed by participating in the BabyLM challenge, resulting in a published paper at CoNLL 2023~\cite{omar-2023}. The results of this thesis encourage further development in the direction of retrofitting transformer-based models to induce syntactic structures, supported by the relatively acceptable performance of StructFormer in different experimental settings and, most importantly, the highlighted limitations that require innovative solutions to advance the state of syntactic structure induction.

% I review the methods of inducing 

% representations    Words the  Analysing the syntax of texts requires hierarchical structured processing to 

% show human-level capabilities in comprehending and producing texts. These language models can solve many linguistic tasks as good as humans such as translation, sentiment analysis, and text generation. However, 

% Language 

\cleardoublepage

% do not change this
\pdfbookmark[0]{\IfLanguageName{ngerman}{Danksagung}{Acknowledgments}}{acknowledgments}
\begin{center} 
    \huge \IfLanguageName{ngerman}{Danksagung}{Acknowledgments}
\end{center}

First and foremost, I extend my deepest gratitude to my supervisor, Prof. Dr. Laura Kallmeyer, whose expertise, understanding, and patience added considerably to my graduate experience. Your guidance was invaluable in supervising my lab rotation, formulating my research topic, and writing this thesis. I am exceedingly grateful for the time and effort you invested in both my academic growth and the progression of this work. I would also like to thank my second supervisor, David Arps, for his insightful comments and tips. Your feedback has been instrumental in refining my research focus and methodology. My appreciation also goes out to Regina Stodden for the opportunity to engage in her research work and for introducing me to the Computational Linguistics department at Heinrich-Heine-Universität Düsseldorf. I also must acknowledge the financial support I received from working as a student assistant at Prof. Kallmeyer's research group, this support was crucial to dedicate myself fully to this research work. The teaching staff in the Master's program of AI and Data Science has been of great help in acquiring the required knowledge to complete this thesis. Special thanks to Prof. Dr. Markus Kollmann, Prof. Dr. Stefan Harmeling, Dr. Peter Arndt, and Tim Kaiser for their classes and talks that formed big parts of my knowledge in the fields of Machine learning and Deep Learning

The research and experiments in this thesis could not be completed in this form without making use of the open-source resources available in the computational linguistics and NLP research communities. On top of that, Yikang Shen and his colleagues published open research papers and codebases for advancing the problem of structure induction from language models. The BabyLM challenge's authors also published a complete framework (including several datasets) to assess language models on a broad list of linguistic tasks. The openness and transparency in this field is undeniable.

On a personal note, I want to express my gratitude to my parent, the two persons who are the main reason behind completing this thesis and any other success in my life. All the words of gratitude are not enough to acknowledge your support. My brother, you are my best friend; your words have always encouraged me. My friends, thank you for being there when I needed you.

\cleardoublepage

\pdfbookmark[0]{\contentsname}{content}
\tableofcontents

\listoffigures % entfernen, wenn leer

\listoftables % entfernen, wenn leer

% \listoflistings % entfernen, wenn leer

\mainmatter

\cleardoublepage

%%%%%%%%%%%%%%%%%%%%%%%%%%%%%%%%%%%%%%%%%%%%%%
%%    Beginning of the main document        %%
%%                                          %%
%%    Include your tex-files with \input{}  %%
%%%%%%%%%%%%%%%%%%%%%%%%%%%%%%%%%%%%%%%%%%%%%%

\chapter{Foundations}
\label{ch:intro}

\section{Introduction}

\textit{Language} is a complex system that plays a significant role in human life. It enables us to communicate, shapes our culture, and underpins our cognitive processes. Linguists develop theories and frameworks to systematically analyze the human language, aiming at understanding its structure and function. A foundational theory in linguistics posits the existence of hierarchical structures in human language. Although these structures are not directly observable in the sequences of words and sentences, it has been proven to exist based on substantial evidence~\cite{Chomsky+1957,Tesniere1959,Melcuk1988}. In parallel, computer scientists and computational linguists work on developing computational models capable of performing language tasks comparable to human capabilities~\cite{clark-etal-2021-thats}. Recently, this effort has resulted in computational language models achieving unprecedented performance in many language tasks~\cite{ziyu-etal-2023-lens}, popularly renowned in applications like OpenAI's ChatGPT\footnote{\url{https://chat.openai.com/}}. Despite this significant progress, it remains unclear how these models interpret languages and whether hierarchical linguistic structures akin to those proposed by linguists are present in their underlying representations. Answering this question has the potential to refine existing linguistic theories and develop more interpretable computational language models. 

Hierarchical structures are integral to languages in governing the forming of valid sentences. This hierarchical nature is formalized by the study of \textit{Syntax} in linguistics, where the syntax of a language describes the latent structure of a valid sentence, irrespective of its meaning validity~\cite{ling-fund}. On the other hand, foundation language models\footnote{Foundation models are deep learning models that have conferred a broad shift in AI research and deployment~\cite{foundation_models}.} (e.g. ELMo~\cite{peters-etal-2018-deep}, BERT~\cite{devlin-etal-2019-bert}, GPT-3~\cite{brown2020language}) have proven their capability to generate human-level textual outputs and solve specific linguistic tasks that imply certain degree of acquired knowledge of hierarchical syntactic structures~\cite{foundation_models}. This implication inspired several studies to attempt extracting these hypothesized underlying structures from language models~\cite{yogatama2017learning, maillard2018jointly, choi2018, onlstm, kim-etal-2019-unsupervised, drozdov-etal-2019-unsupervised-latent, wang-etal-2019-tree, shen-etal-2021-structformer}, extending an older research direction, \textit{Grammar Induction}, and intersecting with a recent promising objective of developing \textit{Explainability} in AI research.

However, despite their undeniable values, these studies face limitations that keep the problem far from being solved. The structures induced from language models in these studies are typically compared against reference structures annotated by human experts based on linguistically defined theories and frameworks. However, the evaluation scores reported in these studies are often too weak to prove that these language models adhere to the referenced linguistic theories and frameworks~\cite{han-etal-2020-survey,li-risteski-2021-limitations}. Furthermore, the methodologies employed in many of these studies lack clarity regarding their operational mechanisms (how do these methods induce structures?) and their rationale (what makes these methods work?)~\cite{li-risteski-2021-limitations,shrodinger}. Additionally, the experimental procedures in this domain still rely on traditional tools and concepts, imposing unnecessary constraints on recent studies that incorporate new methodologies (e.g., neural-based models) that differ from those traditionally used. These traditional tools include relying on size-limited annotated datasets for pretraining and evaluation, applying basic tokenization methods, and using evaluation metrics whose suitability in this problem needs reevaluation~\cite{plank-etal-2015-dependency}.

This thesis attempts to understand the problem of inducing syntactic structures from language models on different levels. The scope of research in the study is narrowed down to two main syntactic frameworks, \textit{Constituency} and \textit{Dependency} structures, and further limited to methodologies primarily based on neural networks. \textit{StructFormer}~\cite{shen-etal-2021-structformer}, one of the recent promising methods that follow this narrowed scope is selected as a study subject to aid the process of analyzing the mechanisms of methodologies addressing the problem, exploring the evaluation protocols applied there, investigating the limitations that similar methods face, and potentially propose solutions to these limitations. As a practical application, the public challenge \textit{BabyLM}~\cite{warstadt-etal-2023-findings} is chosen to act as a platform for an extensive linguistic evaluation and a benchmark for the studied model and the proposed modifications in this thesis. Participation in this challenge is documented and published in the proceedings of the Conference on Computational Natural Language Learning \textit{CoNLL 2023}~\cite{omar-2023}.

\textbf{The contributions of this thesis are as follows:}

\begin{enumerate}
    \item Elucidate the architecture and concepts of the StructFormer model, illustrate an input/output case of a pretrained variant, verify the model's performance, and measure its reliability in inducing consistent structures.
    \item Introduce potential modifications on StructFormer by repositioning its parser network within the attention blocks in the transformer architecture and adapting its pretraining and parsing evaluation to subword-based tokenization.
    \item Pretrain StructFormer on a larger dataset and different in domain than the one implemented in the original publication, and evaluate its induced structures against structures generated from an automatic parser. 
    \item Analyse the linguistic performance of two variants of StructFormer and a baseline transformer to measure the effects of StructFormer's architecture and the introduced modifications using the BabyLM evaluation pipeline and benchmark these variants against 118 other models that are pretrained on the same dataset.
\end{enumerate}

In this thesis, the terms Computational Linguistics (CL) and Natural Language Processing (NLP) are used interchangeably to refer to the fields and tasks that utilize computational methods on language-related tasks.

\section{Thesis Structure}

The thesis is organized into six chapters, including this foundational chapter. In the latter portion of this chapter at Section~\ref{sec:prelim}, a brief review of some essential concepts relevant to the thesis topic is provided. Chapter~\ref{ch:related_work} presents an overview of related work in the field. Following this, Chapter~\ref{ch:sf} offers a comprehensive introduction to the StructFormer model, detailing its foundational concepts and showcasing the results of its reproduction experiments, in addition to a self-consistency experiment. Chapter~\ref{ch:sf_dev} discusses three further experiments to develop the StructFormer model and similar methods. Chapter~\ref{ch:ev} details the evaluation pipeline of the BabyLM challenge, including the results and a discussion section that summarizes the findings from the experiments. Finally, Chapter~\ref{ch:conc} concludes the study, highlighting the main findings and suggesting directions for future research.

%%%%%%%%%%%%%%%%%%%%%%%%%%%%%%%%%%%%%%%%%%%%%%%%%%%%%%%%%%%%%%%%%%%%%%%%%%%%

\section{Preliminaries}
\label{sec:prelim}

This section covers key concepts related to the thesis content to prepare the readers for the following chapters. In Subsection~\ref{sec:linguistic}, I briefly introduce the necessary concepts from the linguistics perspective, introducing the subfields of linguistics, Syntax, and Constituency and Dependency structures. In Subsection~\ref{sec:comput}, I briefly present the paradigm of deep learning and its core concepts in NLP. Eventually, I discuss the connections between the linguistic and computational perspectives with respect to Syntax and deep learning in Subsection~\ref{sec:deep_syntax}.

\subsection{Linguistics}
\label{sec:linguistic}

The field of linguistics is the scientific study of human natural language. Traditionally, linguistics adopts a structural perspective, where linguists propose various formalizations concerning the structures inherent in language. Linguistics encompasses several subfields, each focusing on distinct aspects of language structure. As summarized in Table~\ref{tab:subfields}, these subfields explore the systematic patterns observed across languages, covering units such as phonemes, morphemes, words, phrases, clauses, and sentences. These units and their associated patterns frequently demonstrate similarities and differences across languages~\cite{AuthorYearLinguistics,ling-fund}.

\begin{table}[!htp]
    \centering
    \caption[Linguistics Subfields]{A non-exhaustive sample of structural subfields of linguistics as listed by~\cite{ling-fund}.}
    \label{tab:subfields}
    \begin{tabular}{lp{12cm}}
        \toprule
        Subfield   & Description\\
        \midrule
        Phonetics  & The study of the sounds of human language\\
        Phonology & The study of sound systems in human languages\\
        Morphology  & The study of the formation and internal structure of words\\
        Syntax  & The study of the formation and internal structure of sentences\\
        Semantics & The study of the meaning of sentences\\
        Pragmatics  & The study of the way sentences with their semantic meanings are used for particular communicative goals\\
        \bottomrule
    \end{tabular}
\end{table}

\begin{figure}[!htp]
    \centering
    \caption[Structural Ambiguity Example]{Two constituency trees corresponding to Interpretation \#1 (left) and Interpretation \#2 (right). Constituency Parsing is one of the most common structural syntactic frameworks. These annotations are inspired by popular examples in the field as in~\cite{Charniak1997StatisticalPW}.}
    \label{fig:str_ambg}
    \vspace*{10mm}
    \includegraphics[width=1\linewidth]{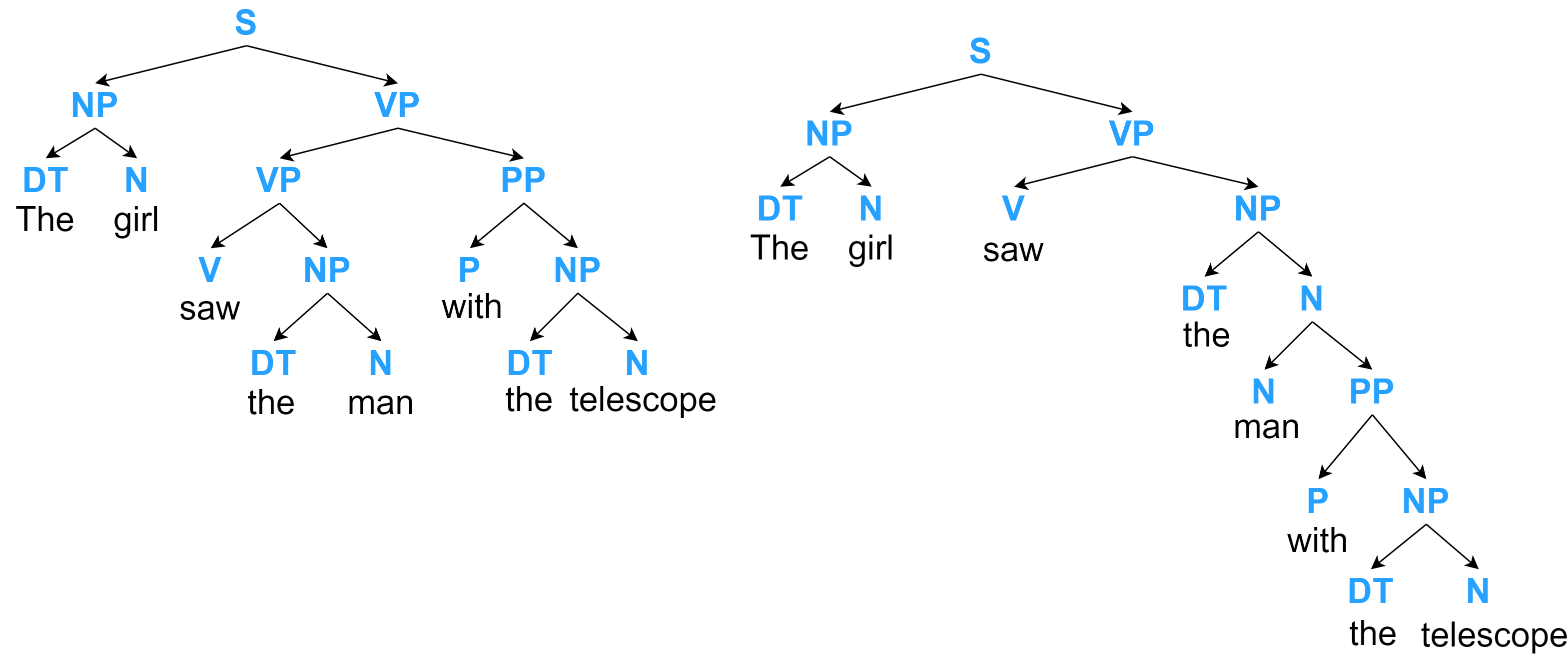}
\end{figure}

\textbf{Syntax} is the set of rules that organizes words into valid sentences and phrases. Although we can not see syntactic structures directly, linguists agree that languages are built hierarchically. When we communicate via language, we combine sounds or letters to make words and then arrange these words into sentences. However, even though these units seem to be combined sequentially, our brains implicitly understand and process these units based on structural representations. The most compelling evidence for this structural nature lies in structural ambiguity, where a single sequence of words can yield multiple interpretations. Consider the sentence: "The girl saw the man with the telescope." Upon encountering this sentence, two interpretations emerge: Interpretation \#1 suggests the girl used the telescope to see the man, implying the man is far from the girl. Interpretation \#2 implies that the girl observed a man holding a telescope. These divergent interpretations are believed to reflect two distinct syntactic structures processed at the brain level, as depicted in Figure~\ref{fig:str_ambg}.

\textbf{Syntactic Parsing} is a crucial task in computational linguistics (CL) and natural language processing (NLP), focusing on the analysis of language's syntactic structure. This process involves analyzing the words in a sentence according to a specific grammar framework and assigning a structure to the sentence~\cite{JurafskyMartin2024}. There are various theories of grammar, each proposing different frameworks for describing language syntactic structures. This thesis considers the two primary grammar frameworks: constituency grammars and dependency grammars. \textit{Part-of-speech} (POS) tagging is also related to syntactic parsing and often serves as either a prerequisite or a component of it. POS tagging helps resolve some semantic ambiguities by labeling each word in a text as corresponding to a particular part of speech based on its definition and context. Syntactic parsing can be performed manually by linguists who apply a specific grammar framework to a text and construct the corresponding syntactic structure, a process known in linguistics as \textit{Annotation}. Annotations of large amounts of texts according to a specific grammar framework are often compiled in a \textit{Treebank}. However, manual parsing is labor-intensive and time-consuming, often requiring months or even years to construct the Treebank by teams of linguistics experts and students. \textit{Automatic Parsing} is also possible using computer programs following efficient parsing algorithms and methods that are in continuous development to yield more accurate results in less running time.

\textbf{Constituency Grammar (CG)} suggests that in various languages, groups of words work together as a single unit, known as a constituent. Linguists suggest different frameworks to model constituency. This thesis adopts only the Context-Free Grammar (CFG) framework for simplicity. A CFG comprises rules that apply to non-terminal symbols (symbols that can be broken down into further symbols) and terminal symbols (symbols that cannot be broken down further). The language defined by a particular CFG includes all possible sequences that can be created according to the rules of this CFG~\cite{JurafskyMartin2024}. A CFG rule processes a sentence's structure breakdown into its constituent parts. An example of a CFG rule within this framework can be: \(S \rightarrow NP \; VP\). In this rule, Sentence (\(S\)) is comprised of a Noun Phrase (\(NP\)) followed by a Verb Phrase (\(VP\)). Further decomposition of (\(NP\)) and (\(VP\)) into their constituent elements is guided by additional rules, such as: \(NP \rightarrow DT \; N\) and \(VP \rightarrow V \; NP\). Those rules specify that a (\(NP\)) consists of a Determiner (\(DT\)) and a Noun (\(N\)), and a (\(VP\)) is formed by a Verb (\(V\)) followed by a Noun Phrase (\(NP\)). Terminal symbols such as \(DT\) are then mapped to real words to enable the parsing of word sequences. The aggregation of these rules results in constituency parses for sentences. Constituency parses are primarily represented in the form of \textit{brackets} as shown below; this notation uses parentheses to encapsulate syntactic units within a sentence, maintaining the hierarchical structure in a linear form. This notation places syntactic category labels at the beginning of each unit, followed by its constituents, down to the terminal actual words.

\begin{center}
    % \caption{A constituency parse represented in the brackets notation for the sentence: \textit{The cat sat on the mat.}}
    % \label{eq:brackets}
    % \centering
    \footnotesize{\texttt{(S (NP (DT \textbf{The}) (N \textbf{cat})) (VP (V \textbf{sat}) (PP (P \textbf{on}) (NP (DT \textbf{the}) (N \textbf{mat})))))}}    
\end{center}

% \begin{listing}[!htp]
% \caption{A constituency parse represented in the brackets notation for the sentence: \textit{The cat sat on the mat.}}
% \label{eq:brackets}
% \centering
% \footnotesize{\texttt{(S (NP (DT \textbf{The}) (N \textbf{cat})) (VP (V \textbf{sat}) (PP (P \textbf{on}) (NP (DT \textbf{the}) (N \textbf{mat})))))}}
% \end{listing}

\begin{figure}[!htp]
    \centering
    \caption[An Example of a Constituency Tree]{An Example of a Constituency Tree for the sentence: \textit{The cat sat on the mat.}}
    \label{fig:const_tree}
    \vspace*{3mm}
    \includegraphics[width=0.4\linewidth]{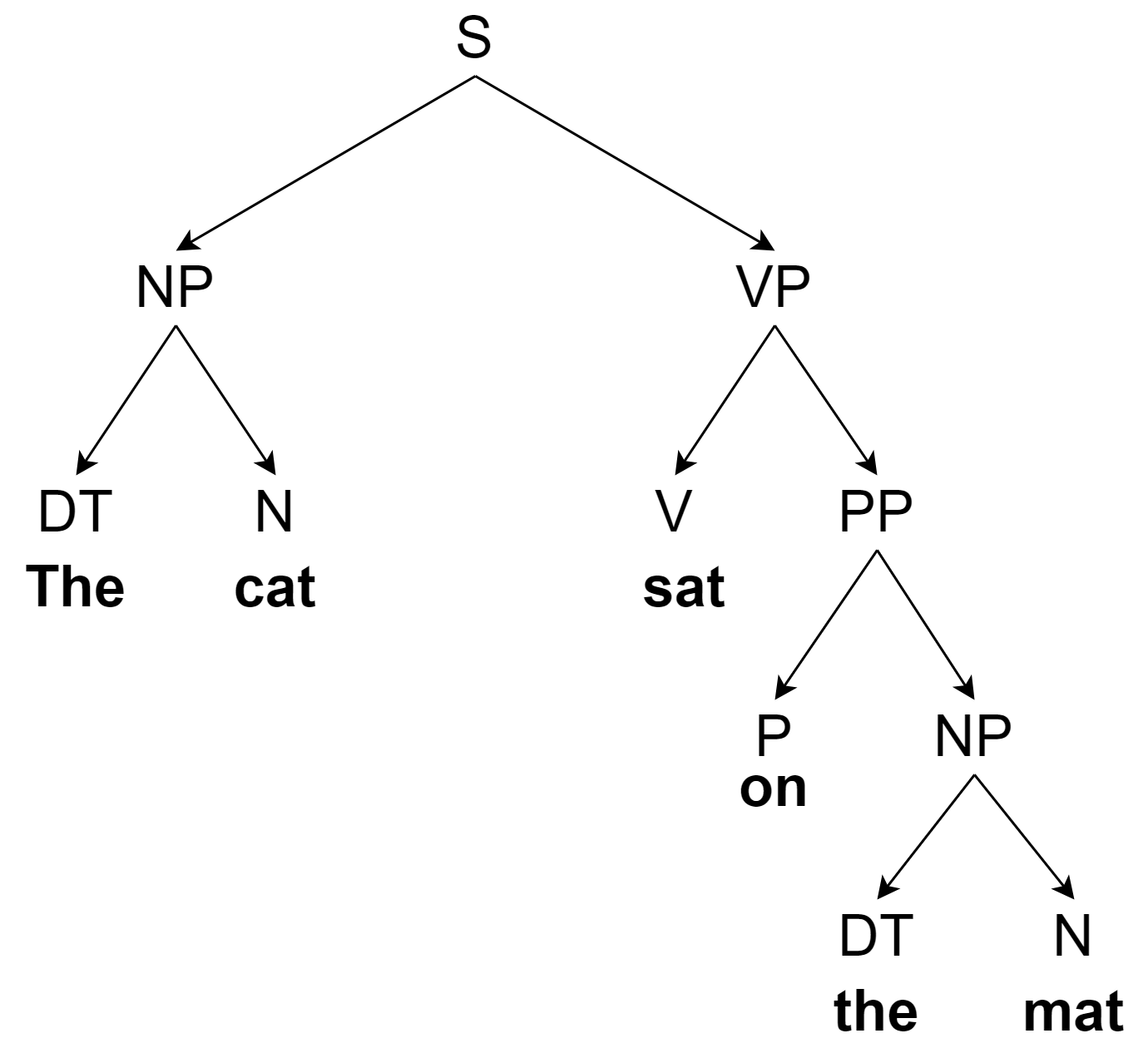}
\end{figure}

Constituency parses can also be represented through \textit{Constituency Trees}. These trees visually outline sentences' hierarchical syntactic structure as constituency grammar prescribes. The tree starts with a root, symbolizing the entire sentence. It branches out into internal nodes for syntactic categories like noun phrases (NP) and verb phrases (VP), down to the leaves, representing the terminal elements or the actual words of the sentence. An example of a constituency tree is depicted in Figure~\ref{fig:const_tree}. I constructed this constituency tree based on outputs from the Berkeley Neural Parser~\cite{kitaev-klein-2018-constituency}. Some constituency frameworks enforce the trees to be binary branching only. In contrast, others allow for non-binary branching trees, where nodes can split into two or more subnodes to accommodate various syntactic relationships. Apart from the leaves, each node carries only a label indicating its syntactic role (e.g., NP for noun phrase). The leaves carry, in addition to a label, an actual word from the sentence. Also, many constituency frameworks enforce a restriction that prevents branches from crossing each other. This restriction ensures that the syntactic structure of sentences is represented without any overlap in branches, maintaining the linear, sequential order of words. Several key treebanks have been constructed for constituency parsing, offering extensive annotated corpora across various languages and domains. The Penn Treebank PTB~\cite{marcus-etal-1993-building} is perhaps the most renowned, providing a large corpus of English text annotated for syntactic structure. OntoNotes~\cite{onto_notes} stands out for its multi-language and multi-domain annotations, including English, Chinese, and Arabic. The TIGER Treebank~\cite{tiger} serves the German language with comprehensive syntactic annotations.

\begin{figure}[!htp]
    \centering
    \caption[An Example of a Dependency Parse in CoNLL format]{An Example of a Dependency Parse in CoNLL format for the sentence: The cat sat on the mat. \textit{ID} is the word index in the sentence, \textit{Word} is the actual word in the sentence, \textit{Lemma} is the base or dictionary form of the word, \textit{POS} is the part-of-speech tag of the word, \textit{Head} is the ID of the head word for each word, and \textit{DepRel} is the dependency relation label. The root of the sentence (the main verb) points to "0" because it doesn't have a head.}
    \label{fig:conll}
    \vspace*{3mm}
    \includegraphics[width=0.5\linewidth]{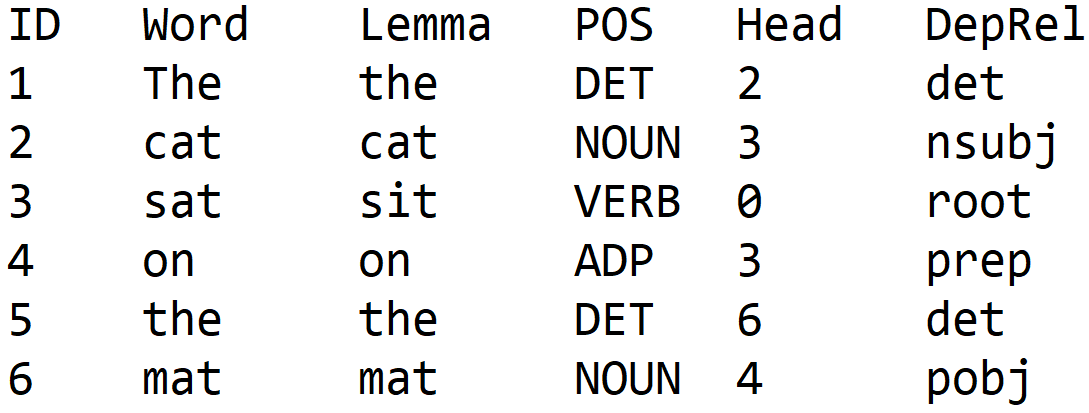}
\end{figure}

\begin{figure}[!htp]
    \centering
    \caption[An Example of a Dependency Tree]{An Example of a Dependency Tree for the sentence: The cat sat on the mat.}
    \label{fig:dep_treee}
    \vspace*{3mm}
    \includegraphics[width=0.8\linewidth]{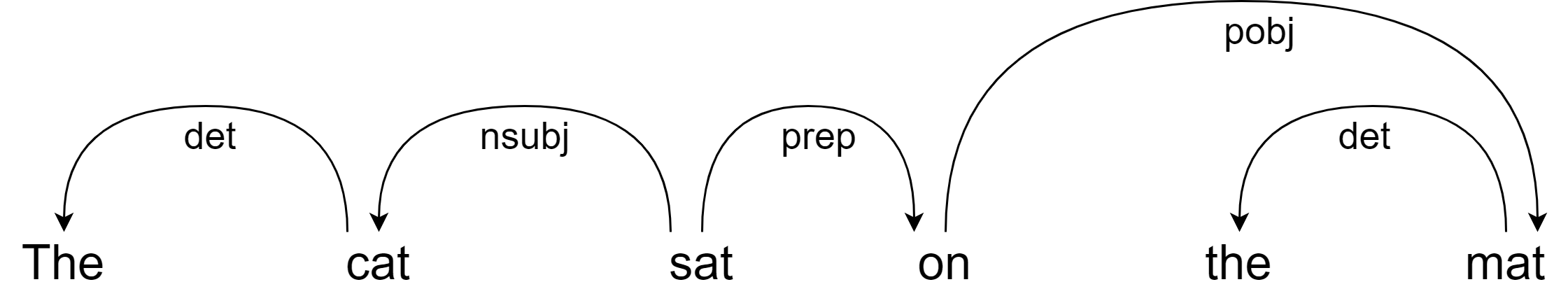}
\end{figure}

\textbf{Dependency Grammar (DG)} is another common framework that analyzes the grammatical structure of a sentence by establishing relationships between "head" words and "dependant" words that modify those heads. Unlike constituency grammar, which focuses on the hierarchical structure of phrases, dependency grammar identifies direct binary relations between words in a sentence, illustrating how words depend on each other. Each word in a sentence is linked to its head through a label indicating the type of syntactic relationship between the words. Dependency parses can be represented through CoNLL format tables, a textual representation that captures the dependency structure in a tabular form as shown in Figure~\ref{fig:conll}. A dependency parse can also be represented graphically via a dependency tree where nodes represent the words in the sentence, directed edges (arcs) connect head words to their dependents, indicating the dependency direction and labels on the edges describe the type of dependency relationship. The dependency tree's root is typically the sentence's main verb or predicate, from which all other elements are directly or indirectly connected, reflecting the sentence's predicate-argument structure. An example of a dependency tree is illustrated in Figure~\ref{fig:dep_treee}; I constructed this dependency tree based on outputs from an automatic parser\footnote{\url{https://spacy.io/api/dependencyparser}}. In this example, \(det\) refers to the Determiner relation, \(nsubj\) refers to the nominal subject relation, \(prep\) refers to the prepositional modifier, and \(pobj\) refers to the prepositional object relation. The Universal Dependencies (UD)~\cite{nivre-etal-2016-universal} treebanks are at the forefront of dependency treebanks, providing cross-linguistically consistent annotations for a wide range of languages, which makes them indispensable for multilingual NLP tasks.

The theoretical contrast between constituency and dependency lies in constituency parsing being rooted in generative grammar theories asserting that sentences can be generated from a finite set of rules applied to a finite set of elements. In contrast, Dependency Parsing is based on dependency grammar theories, which argue that syntactic structure primarily consists of relations between lexical items (words), with one word governing its dependents. Despite these differences, linguists have developed algorithms that convert between constituency and dependency parses to a certain degree of accuracy as in~\cite{de-marneffe-etal-2006-generating}.

\subsection{Deep Learning}
\label{sec:comput}

The pursuit of computational language modeling has seen a significant transition through the introduction of deep learning methods, most importantly, as deep learning methods have abandoned explicit rule-based modeling in favor of a universal function approximator~\cite{HORNIK1989359}, which learns implicit features, patterns, and rules from large datasets. In this subsection, I touch on some deep learning methods developed over time for processing languages.

\textbf{Word Embeddings} revolutionized CL and NLP by representing words as dense vectors within a continuous, low-dimensional\footnote{The number of dimensions usually ranges between 100 and 1024 dimensions. This range is considered "low" given the complexity and vastness of human language.} vector space. Unlike prior methods that relied on high-dimensional one-hot\footnote{One-hot encoding is a vector representation where each word is represented as a sparse vector with a '1' in the position corresponding to the word's index in the vocabulary and '0's in all other positions.} vectors, leading to sparse and computationally inefficient representations, word embeddings capture the semantic and syntactic essence of words. Techniques such as \textit{Word2Vec}~\cite{mikolov2013efficient}, \textit{GloVe}~\cite{pennington-etal-2014-glove}, and \textit{fastText}~\cite{bojanowski-etal-2017-enriching} facilitated this leap by learning word representations from textual contexts using variants of neural networks, thereby mapping semantically similar words to nearby points in the vector space. This approach addressed the curse of dimensionality associated with one-hot encoding and introduced models with a deeper understanding of linguistic knowledge. However, word embeddings like Word2Vec and GloVe provide static, context-independent representations, limiting their ability to capture the nuances of words with multiple meanings. The emergence of transformer models, such as BERT~\cite{devlin-etal-2019-bert} and GPT~\cite{brown2020language}, revolutionized learning word embeddings by introducing dynamic, context-sensitive embeddings. This transition from static to dynamic embeddings has significantly enhanced NLP applications, allowing models to grasp complex linguistic structures and polysemy\footnote{Polysemy refers to the phenomenon where a single word has multiple related but distinct meanings, depending on the context in which it is used.} with unprecedented accuracy.

\textbf{Recurrent Neural Networks (RNNs)}~\cite{rnn87} and \textbf{Long Short-Term Memory (LSTM)} \cite{lstm97} are crucial architectures in NLP designed to handle sequential data, such as text. Before the advent of RNNs, traditional neural networks faced challenges with sequences due to their inability to maintain contextual information over variable-length inputs, limiting their effectiveness in tasks that required an understanding of order and context, like language modeling and text generation. RNNs addressed this limitation by introducing loops within their architecture, allowing them to pass information across sequence steps and maintain a form of memory, thereby capturing temporal dynamics and dependencies within the text. However, RNNs encountered difficulties with long-term dependencies due to the vanishing gradient problem, where the contribution of information decays over distance, making it hard to learn correlations between distant events in a sequence. LSTMs, a special kind of RNN, were introduced to solve this problem. LSTMs incorporate mechanisms called gates that regulate the flow of information, enabling them to remember and forget information selectively over long sequences.

\textbf{Convolutional Neural Networks (CNNs)}~\cite{cnn98}, while initially developed for image processing and computer vision tasks, have been effectively adapted for use in NLP. The introduction of CNNs to NLP~\cite{kim-2014-convolutional} aimed to improve the capture of local and hierarchical patterns in text data, such as identifying key phrases or syntactic structures within sentences. CNNs address these challenges by applying convolutional filters (kernels) over the text data, allowing the model to detect patterns like n-grams (sequences of n words) across different parts of a sentence. This capability enables CNNs to capture semantic and syntactic relationships in text, improving performance in tasks such as text classification, sentiment analysis, and named entity recognition. CNNs in NLP are primarily used as 1D (one-dimensional) CNNs. While 2D CNNs are prevalent in image processing, where convolutional filters scan two-dimensional spatial structures (height and width of images), 1D CNNs are suited for sequential data like text, where the convolution operates over a single spatial dimension (the sequence of words or characters). The 1D convolutional layer slides over the text sequence represented as sequences of words embedding vectors, capturing patterns within fixed-size windows across the sequence. This sliding allows the model to detect meaningful n-gram features or patterns in the text, which are crucial for understanding linguistic structure and meaning.

\textbf{Attention Mechanism} is another core concept in recent state-of-the-art language models. In 2017, Vaswani et al. published a paper titled "Attention was all you need"~\cite{vaswani17}, proposing the Transformer model for processing textual data. At the heart of the Transformer model is the attention mechanism. The attention mechanism enables models to focus selectively on specific parts of the input data, similar to how human attention works when concentrating on particular aspects of a scene or conversation. Before the attention mechanism was introduced, RNNs and LSTMs dominated the NLP scene by processing input sequences in a fixed order, making it difficult to prioritize more relevant pieces of information, especially in longer sequences. This limitation often led to suboptimal performance on tasks like machine translation, where the relevance of input words can vary significantly throughout a sentence. The introduction of attention mechanisms solved this by allowing models to learn to assign weights of importance to different parts of the input data, dynamically adjusting what to focus on at each processing step. The key equation for self-attention is defined in~\cite{vaswani17} as shown in Equation~\ref{eq:attention}, where $Q$ represents the \textit{query} matrix, corresponding to the set of vectors to which the attention is being directed. $K$ denotes the \textit{key} matrix, matched against the query to compute attention scores. After computing the attention scores, the \textit{value} matrix $V$ contains the information to focus on. Moreover, the dimensionality of the key vectors $d_k$ is used for scaling the dot product to prevent it from growing too large when the dimensions are high. The dot product $QK^T$ measures the alignment between queries and keys, indicating similarity, scaled by the square root of $d_k$ to stabilize the gradients during training. The $\text{softmax}(\cdot)$ function is applied to the scaled dot products to obtain attention scores as positive values that sum to one. Then, $V$ is multiplied by the softmax output, resulting in a weighted sum of the value vectors, with weights indicating attention scores. This dynamic calculation by the Attention function allows the Transformer to focus more on the pertinent parts of the input data, making it contextually sensitive.

\begin{equation}
\label{eq:attention}
    \text{Attention}(Q, K, V) = \text{softmax}\left(\frac{QK^T}{\sqrt{d_k}}\right)V
\end{equation}

\begin{figure}[!htp]
    \centering
    \caption[Transformer Architecture]{The Transformer model architecture. The figure is copied from~\cite{vaswani17}}
    \label{fig:transf}
    \includegraphics[width=0.46\linewidth]{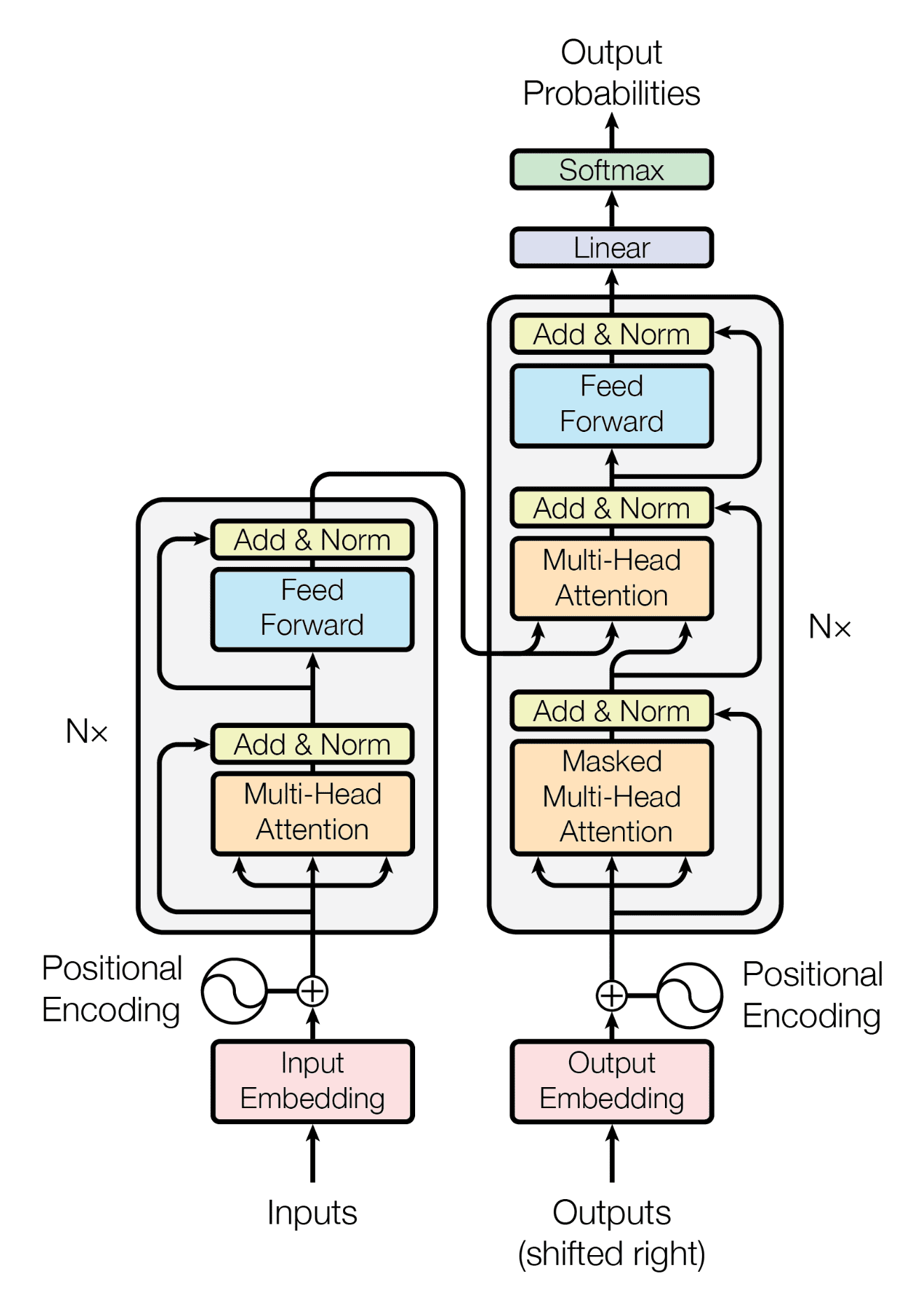}
\end{figure}

The Transformer architecture (illustrated in Figure~\ref{fig:transf}) also introduces the concept of multi-head attention, which involves running several attention mechanisms in parallel. This design allows the model to focus on different positions and represent different subspaces at different positions. The output of each head is then concatenated to yield the expected dimensions. Another crucial aspect of the Transformer model is its use of positional encodings, added to the input embeddings at the bottoms of the encoder and decoder stacks, to make use of the order of the sequence. With its reliance on self-attention and abandonment of recurrence, the Transformer model has been a revolutionary development in NLP. The model's ability to parallelize processing and learn long-range dependencies within sequences has made it a foundation for subsequent developments in the field, including the creation of foundation models like BERT~\cite{devlin-etal-2019-bert} and GPT~\cite{brown2020language}.

\textbf{Language Modeling (LM):} After introducing these sophisticated architectures such as RNN, LSTM, and Transformer, A key question arises: How do these architectures learn language? The answer lies in the pretraining techniques applied to these models when pretrained on extensive textual datasets. Masked Language Modeling (MLM) and Autoregressive Modeling (AR) are the two prevalent techniques. MLM has become essential to training modern language models like BERT~\cite{devlin-etal-2019-bert}. In MLM, random words in a sentence are masked out during training. The model is then tasked with predicting these masked words, relying solely on the surrounding context, an objective formalized by the negative log-likelihood loss function, expressed in Equation~\ref{eq:mlm}. Where $\textsl{L}_{\text{MLM}}$ denotes the MLM loss function, $M$ is the set of indices for masked words, $w_i$ is the actual word at position $i$, and $P(w_i | w_{\backslash i})$ is the probability of predicting the masked word given the remaining context of words. This strategy enables the model to learn the intricacies of language and the interrelationships of words. In contrast, AR modeling predicts the next word in a sequence by considering the preceding words only. The core idea is that the current observation in a sequence is regressed upon its previous observations. It is a cornerstone of models designed for generating coherent and contextually appropriate text. The AR objective is mathematically formalized in Equation~\ref{eq:ar}, demonstrating the sequential dependency from the start of the sentence to the current word. Where $\textsl{L}_{\text{AR}}$ is the loss function for AR modeling, $N$ is the length of the sequence, and $P(w_i | w_1, w_2, ..., w_{i-1})$ represents the conditional probability of the word $w_i$ given all preceding words in the sequence.

\begin{equation}
    \textsl{L}_{\text{MLM}} = - \sum_{i \in M} \log P(w_i | w_{\backslash i})
    \label{eq:mlm}
\end{equation}

\begin{equation}
\textsl{L}_{\text{AR}} = - \sum_{i=1}^{N} \log P(w_i | w_1, w_2, ..., w_{i-1})
\label{eq:ar}
\end{equation}

%%%%%%%%%%%%%%%%%%%%%%%%%%%%%%%%%%%%%%%%%%%%%%%%%%%%%

\subsection{Deep Learning and Syntax} 
\label{sec:deep_syntax}

Before the widespread adoption of deep learning for NLP tasks, computational LMs heavily relied on hand-crafted feature templates. These templates represented linguistic concepts such as POS or syntactic structures~\cite{shrodinger}. Compiling these features was time-consuming and often led to an incomplete understanding of language nuances. Models developed with this approach, while limited in their ability to generalize beyond their hand-picked features, offered the advantage of interpretability. In contrast, deep learning models autonomously learn the necessary features for language representation by processing vast quantities of data and optimizing towards a predefined objective. This shift has enabled deep learning models to achieve unprecedented results across various NLP tasks. However, this improvement introduced a notable trade-off: increased accuracy at the cost of interpretability. The absence of explicit feature engineering made it challenging to determine which linguistic aspects were utilized by these models. This ambiguity spurred a fresh inquiry within the NLP community, focusing on precisely understanding what these models learn. Among various aspects under investigation, syntax emerged as a critical area of interest, highlighting the need to unravel how deep learning-based language models comprehend and utilize syntactic information. Several methodologies have been developed to explore syntax within deep learning-based LMs. One method involves evaluating the models' direct outputs to determine whether they generalize based on syntactic features~\cite{marvin-linzen-2018-targeted,warstadt-etal-2020-blimp-benchmark}. Another popular method is \textit{Probing}~\cite{47786, hewitt-manning-2019-structural, arps-etal-2022-probing}, which delves into the models' inner workings, analyzing their hidden computations and representations to determine whether specific syntactic features are learned. The findings from these studies have motivated researchers to attempt extracting syntactic knowledge directly from LMs. One notable endeavor is inducing constituency and dependency trees from such models. The subsequent chapters will introduce these efforts in greater detail, analyzing specific methods that contribute to advancing this research direction.

\chapter{Related Work}
\label{ch:related_work}

In this chapter, I present the significant steps the research community takes to address the problem of inducing syntactic structures. In Section~\ref{sec:problem_def}, I define the problem and discuss some of its essential aspects. Then, Section~\ref{sec:2.1} discusses the evolution of approaches and methodologies addressing the problem. In Section~\ref{sec:2.2}, I focus on the line of work involving syntactic distance, which is the core of our study subject architecture. In Section~\ref{sec:2.3}, important survey sources for the problem are listed. Lastly, in Section~\ref{sec:2.4}, I reflect on the related work study by pointing out some important observations.

\section{Problem Definition}
\label{sec:problem_def}

The problem under focus in this study is the problem of inducing syntactic hierarchical structures from language models. Induction, here, implies that we seek methods that extract syntactic structures from language models without direct supervision, i.e., we do not seek methods that utilize annotated treebanks to train a model on producing target annotated structures from input sentences. However, classifying methods as Supervised or Unsupervised in the context of this problem is not a clear binary decision, as various levels of supervision are observed. Some methods use complete target structures during training, which are undoubtedly classified as supervised methods, whereas other methods leverage linguistic features (e.g., POS tags) that help the model learn syntactic structures without seeing any structures during training. Also, some supervised methods have inspired succeeding methods to be adapted for unsupervised missions. Due to these reasons, throughout this related work study, I mention some methods that are not fully unsupervised due to their vital association with our problem.

The structures we aim to induce from language models are the constituency and dependency trees presented in Subsection~\ref{sec:linguistic}. Representing these structures within the induction process differs from one method to another. However, constituency trees are often represented as nested spans of words, while dependency trees are often represented as adjacency matrices under some restricting criteria to qualify as valid dependency trees. These representations are then easily translated into graphical tree structures. Also, the induced trees often lack syntactic labels (i.e., phrasal categories in constituency trees and dependency relations in dependency trees); this behavior is expected, especially without supervision during training.

The induced trees are often evaluated against reference structures from human-annotated treebanks or automatic parsers. The evaluation measures how much the unlabeled induced trees align with the reference trees. Induced constituency trees are evaluated using unlabeled Precision (P), Recall (R), and (F1) scores, where the set of constituents in an induced constituency tree \(T_{ind}\) is defined as \(C_{T_{ind}}  \coloneqq \Set{c_i}{c_i \in \text{ internal nodes of } T_{ind}}\), and \(C_{T_{ref}}\) the set of constituents in a reference constituency tree \(T_{ref}\) is defined similarly. Unlabeled P, R, and F1 are computed based on the cardinalities\footnote{Cardinality of a set \(A\) is \( |A| \) which is the number of elements in set \(A\).} of \(C_{T_{ind}}\), \(C_{T_{ref}}\), and their intersection as shown in the Equation set~\ref{eq:f12}. 

Induced dependency trees are typically evaluated using the Unlabeled Attachment Score (UAS), where the set of dependency relations in an induced dependency tree \(T_{ind}\) is defined as a set of ordered pairs \(D_{T_{ind}}  \coloneqq \Set{d_i}{d_i=(a, b); \kern0.5em a,b \in S; \kern0.5em d_i \in \text{ dependency edges in } T_{ind}}\), and \(D_{T_{ind}}\) is defined similarly for the reference dependency tree \(T_{ind}\). UAS measures the ratio of words assigned to the same head in the induced tree as in the reference tree, as shown in Equation~\ref{eq:uasss}.

\begin{equation}
\label{eq:f12}
    \begin{split}
         P = \frac{| C_{T_{ind}} \cap C_{T_{ref}} |}{| C_{T_{ind}} |}, \quad R = \frac{| C_{T_{ind}} \cap C_{T_{ref}} |}{| C_{T_{ref}} |},\quad UF1 = \frac{2 P  R}{P + R}
    \end{split}
\end{equation}

\begin{equation}
\label{eq:uasss}
UAS = \frac{| D_{T_{ind}} \cap D_{T_{ref}} |}{| D_{T_{ref}} |}
\end{equation}

\vspace{2em}

\section{Overview of Approaches}
\label{sec:2.1}

Producing syntactic structures from raw texts has long been a challenging and evolving problem in CL. Over the past several decades, this topic has witnessed varying degrees of interest, and the methodologies addressing this challenge have changed in their paradigm. In this section, I present this evolution by discussing the most popular methods in each period.

\textbf{Pre-Neural Networks Era:} In the initial stages, researchers primarily employed heuristic approaches for structure induction, gradually refining basic grammatical rules with additional ones~\cite{olivier1983stochastic, Wolff1988LearningSA}. This approach begins with an empty grammar framework and POS-tagged corpora. By analyzing these corpora, patterns emerge that gradually become rules, which are added to the initially empty grammar. As the field progressed into the 1990s and 2000s, the advent of probabilistic methods in computational linguistics led to the approach of probabilistic context-free grammars (PCFGs), highlighted in studies as~\cite{magerman-1995-statistical, Charniak1997StatisticalPW, collins-2003-head}. Dan Klein and Christopher D. Manning also made substantial contributions, driving a robust body of work in this domain at that time. Their methodologies included techniques like EM-based parameter optimization~\cite{klein-manning-2002-generative}, distributional clustering strategies~\cite{KLEIN20051407}, and concurrent learning of dependency and constituency structures using the Dependency Model with Valence (DMV)~\cite{klein-manning-2004-corpus}. Additionally, the period saw innovative approaches being proposed, often grounded in statistical, probabilistic, or algorithmic breakthroughs, as demonstrated in the works of~\cite{bod-2006-subtrees, seginer-2007-fast}.

\textbf{RNNs and LSTMs:} Inducing syntactic structures from LMs that are based on neural networks is a complex task. Typically, neural models analyze linear sequences of text without the explicit presence of any hierarchical structures. However, the hope is that these models would implicitly represent hierarchical structures like humans implicitly do through their cognitive processes. One of the first attempts that inspired many following architectures is the Tree-LSTM~\cite{tai-etal-2015-improved}. Tree-LSTM marks a pivotal point, sparking numerous attempts to derive syntactic structures using neural models. Its primary goal is leveraging language's inherent hierarchical nature to enhance performance on downstream tasks like semantic relatedness assessment and sentiment analysis rather than directly inferring syntactic structures. \textit{"While the standard LSTM composes its hidden state from the input at the current time step and the hidden state of the LSTM unit in the previous time step, Tree-LSTM composes its state from an input vector and the hidden states of arbitrarily many child units that are known from a reference tree."}~\cite{tai-etal-2015-improved}. Tree-LSTM introduces a novel concept of "hierarchical" language modeling. Yet, it does not fully address learning syntax from raw text without supervision, as it requires the presence of trees during training. Several innovative approaches adapt Tree-LSTM for an unsupervised setting. \cite{bowman-etal-2016-fast} employs a shift-reduce parser to provide the required trees for Tree-LSTM on the fly. \cite{yogatama2017learning} proposes using reinforcement learning to learn hierarchical syntax in an unsupervised setting using Tree-LSTM through a downstream task. \cite{maillard2018jointly} proposes a differentiable natural language chart parser to aid Tree-LSTMs with syntactic trees during training. Other adaptations to standard RNNs and LSTMs are also proposed, \cite{onlstm} introduces Ordered Neurons LSTMs (ON-LSTMs), an LSTM variant augmented with a constituency inductive bias towards performing tree-like composition operations. \cite{kim-etal-2019-unsupervised} presents Unsupervised Recurrent Neural Network Grammar (URNNG), an unsupervised model based on the supervised Recurrent Neural Network Grammar RNNG~\cite{dyer-etal-2016-recurrent}, which was initially a method for learning grammar in a supervised fashion. URNNG tackles the challenges of unsupervised learning by employing amortized variational inference with a structured inference network.

\textbf{PCFGs:} The exploration of PCFG-based methods did not stop with the rise of neural networks; several studies work on integrating modern probabilistic and neural techniques into classical PCFGs. \cite{jin-etal-2019-unsupervised} shows that the use of ELMo embeddings~\cite{peters-etal-2018-deep} with PCFGs enhances the understanding of language morphology and allows for more nuanced grammar representation through a Normalizing Flow\footnote{A deep learning method that enhances the modeling of complex distributions~\cite{9089305}.}. Compound PCFG~\cite{kim-etal-2019-compound} induces grammar by maximizing the marginal likelihood of the sentences that a PCFG generates. Neural Lexicalized PCFGs~\cite{zhu-etal-2020-return} demonstrates that a PCFG can benefit from modeling lexical dependencies. It also explores inducing both constituents and dependencies within a single model. Neural Bi-Lexicalized PCFG~\cite{yang-etal-2021-neural} develops over~\cite{zhu-etal-2020-return} and parameterized Lexicalized PCFGs by directly modeling bi-lexical dependencies, which are dependencies between two adjacent words in a sentence.

\textbf{Deep Inside-Outside Recursive Autoencoders (DIORA):}~\cite{drozdov-etal-2019-unsupervised-latent} introduces a novel approach, grounded in the belief that the most effective sentence compression is achieved by adhering to its syntactic structure. The authors posit that their model would excel in reconstructing inputs by identifying and leveraging the syntactic patterns inherent in the text. DIORA is an unsupervised model that operates much like a masked language model or a denoising autoencoder. First, it encodes all but one of the words from the input sentence as a vector representation, then decodes the missing word from this vector. DIORA encodes the sentence in the shape of a constituency tree, yet the model is trained using raw text only without access to tree annotations. The target tree is unknown during training, so all valid trees are considered simultaneously using an efficient dynamic program with soft vector weighting. A follow-up update to DIORA is the Single Tree Encoding for Deep Inside-Outside Recursive Autoencoders (S-DIORA)~\cite{drozdov-etal-2020-unsupervised}. S-DIORA authors show that \textit{"encoding a single tree rather than a softly weighted mixture of trees could improve the accuracy of predicted trees"}.

\textbf{Induction of Dependency Trees:} Unsupervised dependency parsing is often seen as less challenging than unsupervised constituency parsing because constituency parsing requires the induction of the internal nodes (phrases) within the constituency tree structure. In contrast, dependency parsing primarily focuses only on establishing relationships or dependencies between words~\cite{han-etal-2020-survey}. Many of the mentioned approaches address both constituency and dependency structure induction as in~\cite{klein-manning-2004-corpus,dyer-etal-2016-recurrent,kim-etal-2019-unsupervised,han-etal-2020-survey,zhu-etal-2020-return}. However, some of the works that were not mentioned, aim only at dependency trees induction as in~\cite{kiperwasser-goldberg-2016-easy,Kim2017StructuredAN,wang-etal-2019-tree,tang-etal-2020-dependency,Ahmad2020GATEGA}

\section{Syntax Distance Approach}
\label{sec:2.2}

Fully unsupervised syntactic tree induction from neural LMs remained seemingly an out-of-reach problem until the proposal of the Parsing-Reading-Predict Networks (PRPN)~\cite{prpn} and the introduction of the concept of \textit{Syntactic Distances}. PRPN performance is thoroughly verified by extensive experiments in~\cite{htut-etal-2018-grammar-induction}. The follow-up paper~\cite{onlstm} introduces the ON-LSTM architecture, which simplified PRPN by still attempting to fit a distance metric but with the help of carefully designed master forget gates. A long line of work has invested in this syntactic distance concept and explored different ideas to improve the originally proposed model. Some reuse the same concept but with a different interpretation as in~\cite{wang-etal-2019-tree}. \cite{luo-etal-2019-improving,gu-etal-2022-phrase,ZENG2022103665} use syntactic distances (or similarly inspired measures) in their architectures. StructFormer~\cite{shen-etal-2021-structformer} introduces the first model that uses syntactic distances to induce both dependency and constituency trees, highlighting this exciting line of work and showing promising results in both the quality of its induced trees and its language modeling capabilities. In this thesis, we select StructFormer to represent all the recent methods addressing our problem for several reasons. StructFormer stands out as one of the few approaches that leverage the transformer architecture, the most widely adopted framework in current state-of-the-art applications in NLP. It also tackles the challenge of generating both constituency and dependency trees simultaneously from the same model by refining the inductive bias of the attention mechanism in transformer models to incorporate a generalized syntactic hierarchical bias.

\section{Survey Works}
\label{sec:2.3}

Several publications have focused on surveying and analyzing the methodologies that address syntactic structure induction. These survey works are deemed valuable sources for readers seeking detailed and quantitative comparisons of the discussed methodologies. I list here the key survey works to the extent of my knowledge:

\begin{itemize}
    \item \cite{li-etal-2020-empirical} surveys unsupervised constituency parsing, aiming to standardize experimental settings across various methods, including neural and non-neural approaches. An empirical comparison of old and new methods %applied to English and Japanese, 
    reveals no significant advantage of recent models (published in 2018-2019) over older models (published in 2002-2007).
    
    \item \cite{li-risteski-2021-limitations} explores the capabilities of models integrating syntax into neural LMs, exemplified by architectures like PRPN and ON-LSTM, particularly through the lens of PCFGs, to examine how the size and direction of context influence syntax modeling.
    
    \item \cite{syst_rev} conducts a systematic review of unsupervised grammar induction, analyzing 43 studies to understand the relationship between grammar theories and computational models. It Highlights a gap between theoretical insights and computational practices.

    \item \cite{han-etal-2020-survey} surveys unsupervised dependency parsing by investigating existing methodologies and categorizing them into two main approaches: generative and discriminative. In a similar direction, \cite{12years} also categorizes the spectrum of unsupervised dependency parsing, but with respect to the supervision level.

    \item \cite{williams-etal-2018-latent} investigates latent tree learning models, revealing that they surpass baselines in sentence classification despite the finding that their parsing strategies lack consistency and produce shallower parses that do not align with established semantic or syntactic grammars.
    
\end{itemize}

\section{Observations on Related Work}
\label{sec:2.4}
By analyzing the literature on syntactic structure induction from neural language models, several key observations emerge:

\begin{enumerate}

    \item The objective often varies among the published works. Some approaches focus on inducing syntactic trees that match specific reference trees~\cite{prpn,wang-etal-2019-tree,shen-etal-2021-structformer}. However, others aim at turning sequential language models into hierarchical models, hypothesizing that understanding language through hierarchical structures rather than linear sequences would enhance language modeling~\cite{tai-etal-2015-improved,luo-etal-2019-improving}. Some methods aim to improve the performance on downstream tasks such as sentiment analysis or sentence classification~\cite{bowman-etal-2016-fast,Kim2017StructuredAN}.
    
    \item Some discussed methods need to provide evaluations for their generated trees~\cite{tai-etal-2015-improved,choi2018}, complicating comparisons regarding structure induction across the field. However, there have been some notable empirical surveys~\cite{williams-etal-2018-latent,li-etal-2020-empirical} that have collectively evaluated the induced trees of various models.
    
    \item While this thesis focuses on the induction of syntactic structures from LMs, not all the discussed models in this section fully align with this criterion. Some utilize linguistic elements such as word classes~\cite{Mareek2015MultilingualUD}, and some are pretrained towards supervised tasks like sentiment analysis or textual entailment~\cite{yogatama2017learning,maillard2018jointly}. These models are included due to their significant contribution to the field's development and because they belong more to the unsupervised side rather than the supervised side, which relies on fully annotated training data.
    
    \item Two hidden subproblems are implicitly addressed but not often explicitly presented in the publications related to syntactic structure induction from neural LMs. The problem of representing a tree-like structure (i.e., a dependency tree or a constituency tree) in a compatible representation that can be processed within neural network architectures. And the problem of how to induce a bias in the model architecture or the learning objective, directing the model's focus toward the hierarchical and syntactic aspects of the language without any direct supervision.
\end{enumerate}

\chapter{StructFormer: An Illustrative Study}
\label{ch:sf}

The model under examination in this thesis is StructFormer~\cite{shen-etal-2021-structformer}. This promising model leverages the concept of syntactic distances, a theory introduced by the model's author a few years before introducing the model~\cite{prpn}. StructFormer is one of the models capable of inducing constituency and dependency trees from transformer-based language models, pretrained on raw text within a fully unsupervised framework. This chapter delves into the details of the architecture of StructFormer in Section~\ref{sec:model_arch}. The concepts of syntactic distances and heights are explained in Section~\ref{sec:dist_heigh}. Section~\ref{sec:comps} breaks down and elucidates the function of each component in StructFormer. In Section~\ref{sec:implem}, details of the original implementation are reported. Furthermore, a detailed example of an input sentence processed through the model and an illustration of the step-by-step procedure of inducing the respective syntactic trees are detailed in Section~\ref{sec:example}. The chapter also presents the results documented in the original paper~\cite{shen-etal-2021-structformer} alongside those replicated by our reproduction experiments in Section~\ref{sec:eval}. A self-consisteny experiment is conducted in Section~\ref{sec:self_consist}, in addition to a brief discussion in Section~\ref{sec:sf_disc}.\textbf{ The concepts (algorithms, definitions, and equations) presented in this chapter are based on the original publication~\cite{shen-etal-2021-structformer} with extensive elaboration and exemplification.}

\section{Model Architecture}
\label{sec:model_arch}

\begin{figure}[!htp]
    \centering
    \caption[StructFormer Architecture]{The model architecture of StructFormer~\cite{shen-etal-2021-structformer}. Copied from the original paper with a few edits for a better interpretation.}
    \label{fig:structformer-arch}
    \vspace{1em}
    \includegraphics[width=0.5\linewidth]{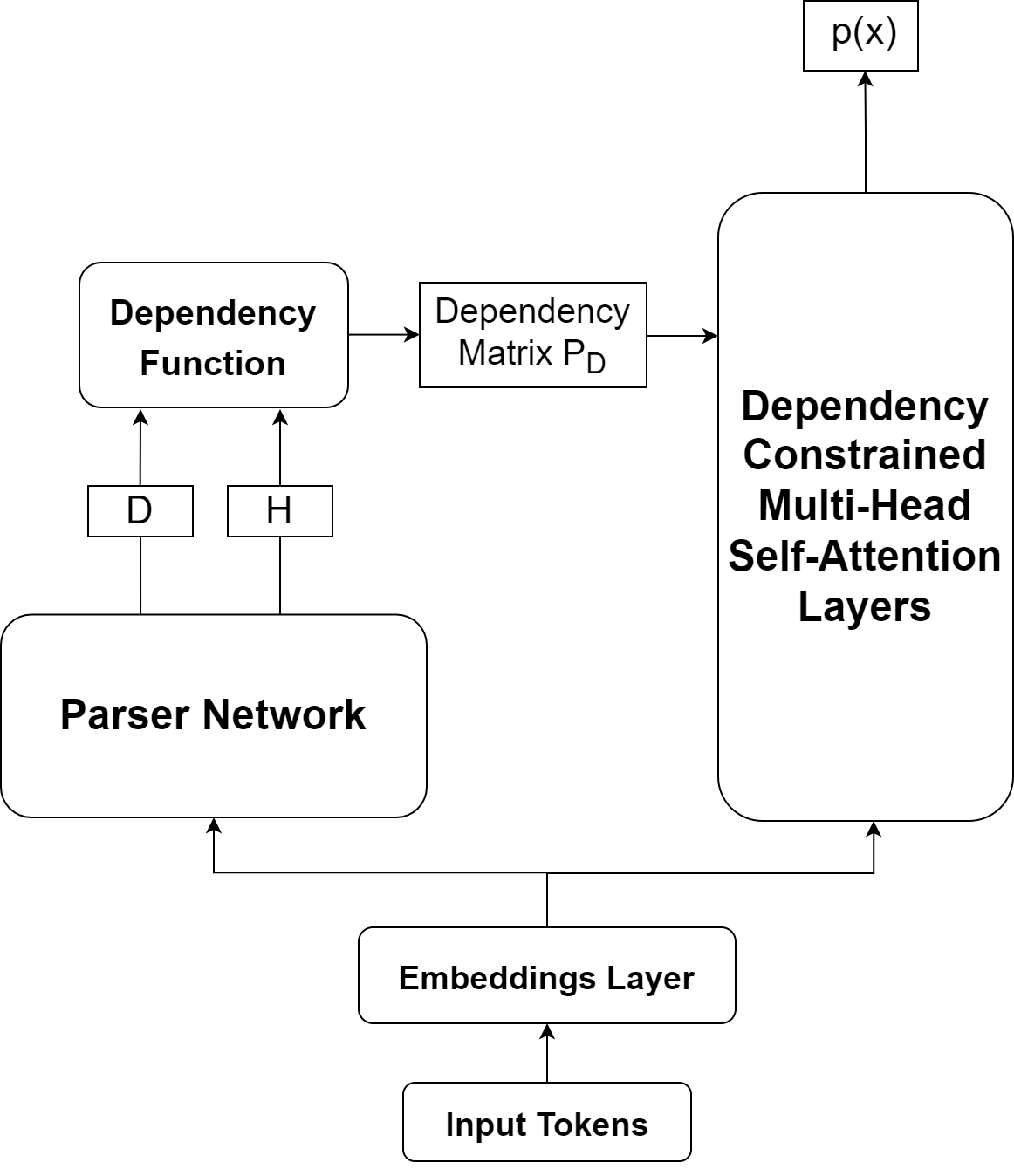}
\end{figure}

The StructFormer architecture (as shown in Figure~\ref{fig:structformer-arch}) represents a nuanced variation of the original vanilla Transformer model, augmented with additional components. These additions aim to embed a hierarchical syntactic inductive bias within the model and to induce syntactic trees using quantifiable measures extracted directly from the model despite the absence of annotated trees. This integration of \textit{Structure} and the \textit{Transformer} framework gives rise to the term \textit{StructFormer}.

The model takes a tokenized sentence \( S = (x_1, x_2, \ldots, x_n) \) as an input. The sequence is first transformed into embeddings through an \textit{Embedding Layer}, resulting in embeddings \( Z = (z_1, z_2, \ldots, z_n) \), where \(z_i \in \textit{\texttt{R}}^{d_{model}} \), \(d_{model}\) is a hyperparameter representing the model's hidden size. Embeddings \(Z\) are then processed by two pathways: a series of \textit{Dependency-Constrained Multi-Head Self-Attention} layers, and a \textit{Parser Network}. The parser network maps the word embeddings \( Z \) to two lists: Syntactic Distances \( D \) and Syntactic Heights \( H \), each comprising \( n \) real-valued scalars. \( D \) and \( H \) are then processed through the \textit{Dependency Function}, a differentiable function which interprets \( D \) and \( H \) as syntactic information. This function is employed to compute a dependency matrix \( P_D \), aiming to encapsulate dependency relationships among words in sentence \( S \). Subsequently, \( P_D \) is fed into the dependency-constrained multi-head self-attention layers, enabling structured attention operations on the embeddings \( Z \). Following a sequence of layers, these embeddings are transformed into a probability distribution \( p(x) \) over the model's vocabulary for each input token. The model is trained on a raw text corpus, guided by the masking language modeling objective. Here, the goal is for the output probability distribution \( p(x) \) to accurately predict masked tokens in the input \( S \). The learning process involves backpropagation of the loss from the output \( p(x) \) relative to the input embeddings \( Z \), optimizing all parameters within the attention layers, the differentiable dependency function, and the parser network. It is postulated that training iterations over a sufficiently large dataset will enable the parser network to accurately predict the values of \( D \) and \( H \), which can then be used to construct meaningful dependency and constituency trees during test time. Notably, no trees are explicitly constructed during the training phase; the model primarily learns to refine its predictions of \( D \) and \( H \) in alignment with the masking language modeling objective. At test time, the parser network can be employed independently to predict \( D \) and \( H \) for any given sentence, thereby facilitating the construction of corresponding trees, as will be elucidated in subsequent sections.

\section{Syntactic Distance and Height}
\label{sec:dist_heigh}

\textbf{Syntactic Distance} was first proposed in~\cite{prpn} to quantify the process of splitting sentences into smaller constituents. According to Definition 3.1 in~\cite{shen-etal-2021-structformer}, syntactic distance is defined as follows:

\begin{quote}
    Let \( T_c \) be a constituency tree for sentence \( S = (x_1, \ldots, x_n)\), and the height of the lowest common ancestor in \( T_c \) for two consecutive words \( x_i \) and \( x_{i+1} \) is \( \tau_i \). Syntactic distances \( D = (d_1, \ldots, d_{n-1}) \) are defined as any sequence of \( n-1 \) real scalars that share the same rank as \( (\tau_1, \ldots, \tau_{n-1}) \).
\end{quote}

Where the height of a node \(a\) in a tree is the number of edges on the longest path from \(a\) to a leaf node, the lowest common ancestor of two nodes \(a\) and \(b\) is the lowest node (the one with the smallest height) in the tree that has both \(a\) and \(b\) as descendants.

\begin{figure}[!htp]
    \centering
    \caption[Syntactic Distance Example]{An example of syntactic distances with the corresponding constituency tree. Lengths of gray bars between words indicate the ranking order of distances (i.e., larger bar = larger distance). Following Algorithm~\ref{alg:dtotree}, at each constituent in the tree, the split point is decided as the point of largest syntax distance (symbolized by the sign '\textbf{+}'). The blue tags here refer to the human-annotated syntactic categories, but they have nothing to do with the concept of syntactic distance.}
    \label{fig:distances}
    \includegraphics[width=0.75\linewidth]{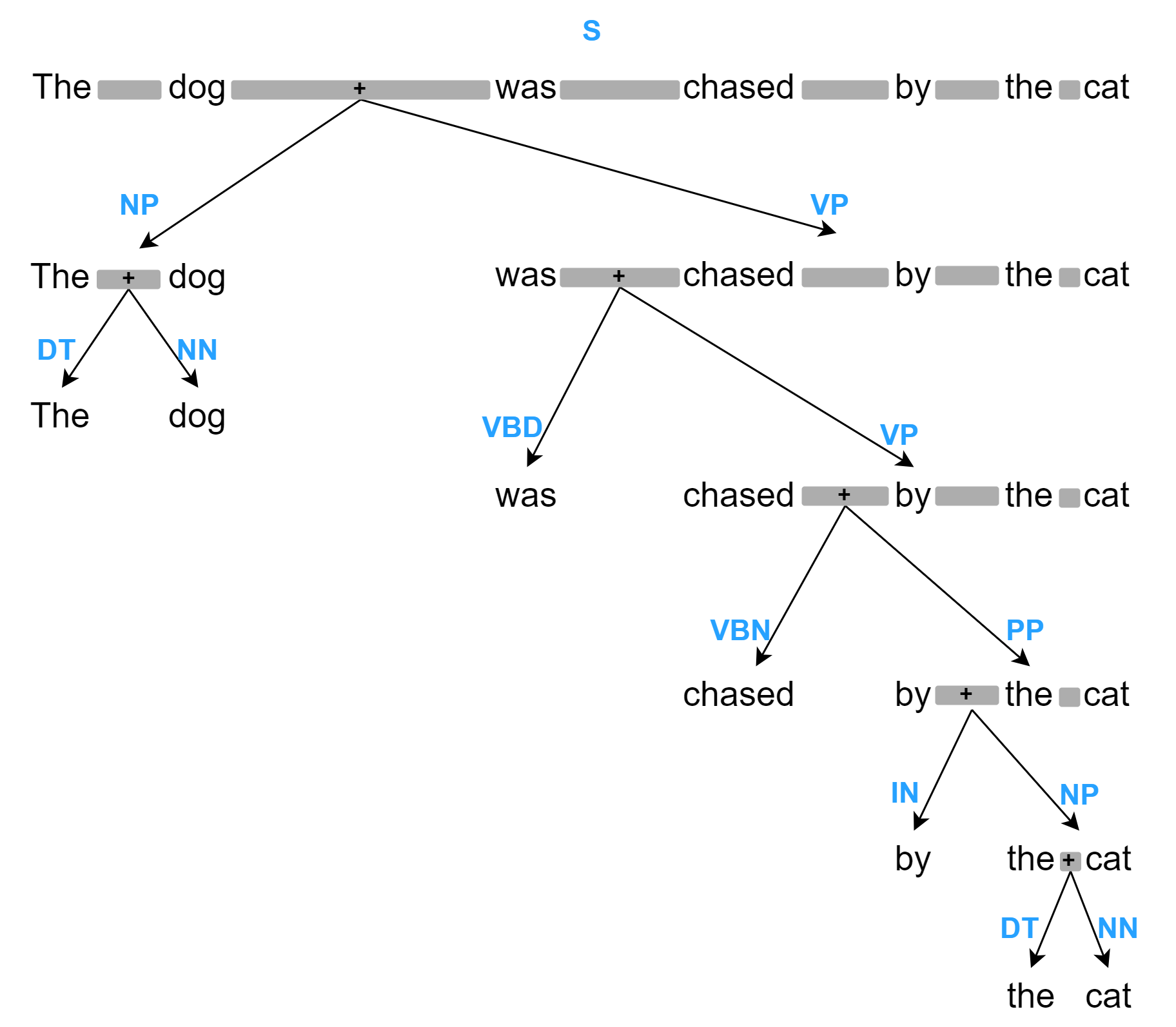}
\end{figure}

In essence, each syntactic distance \( d_i \) correlates with a specific split point \( (i, i+1) \) in a sentence, guiding the hierarchical segmentation into smaller constituents. This approach suggests that any sequence of \( n-1 \) real values can be uniquely mapped to an unlabeled binary constituency tree with \( n \) leaves, following a deterministic recursive procedure as delineated in Algorithm~\ref{alg:dtotree}. Building on the observations of~\cite{prpn, wang-etal-2019-tree}, the syntactic distance metric encapsulates the degree of information exchange between constituents. Specifically, a larger syntactic distance \( d_i \) implies a diminished exchange of short-term or local information between segments \( x \leq i \) and \( x > i \). Therefore, the precision in predicting syntactic distances \( D \) is crucial for accurately constructing binary constituency trees. To exemplify, consider a case where arbitrary\footnote{The values are set in a way that builds the same tree as the target tree.} values of syntactic distances \( D \) are assigned to a sentence \( S \). The resulting configuration, capable of yielding an accurate constituency tree, is depicted in Figure~\ref{fig:distances}.

\begin{algorithm}[!htp]
\caption{(Minimally edited version of Algorithm 1 in~\cite{shen-etal-2021-structformer} for a better interpretation.). Build Binary Constituency Tree \( T_c \) for Tokenized Sentence \(S\) from Syntactic Distances \(D\). The function \texttt{CONSTITUENT} is a recursive function that builds a binary tree with left and right branches containing spans of tokens until reaching single tokens that turn into leaf nodes and terminate the algorithm.}
\label{alg:dtotree}
\begin{algorithmic}[1]
\Function{Constituent}{$S$, $D$}
    \If{$D = []$}
        \State $T_c \gets \Call{Leaf}{S}$
    \Else
        \State $i \gets \arg\max_i(D)$
        \State $child_l \gets \Call{Constituent}{x_{\leq i}, D_{\leq i}}$
        \State $child_r \gets \Call{Constituent}{x_{>i}, D_{>i}}$
        \State $T_c \gets \Call{Node}{child_l, child_r}$
    \EndIf
    \State \Return $T_c$
\EndFunction
\end{algorithmic}
\end{algorithm}

\textbf{Syntactic Height} is proposed in~\cite{luo-etal-2019-improving}. The definition of Syntactic height is more straightforward than the Syntactic distance definition. Syntactic height is used to directly capture the ranking order of node heights in a dependency tree \( T_d \). Where the height of a node \(a\) in a tree is the number of edges on the longest path from \(a\) to a leaf node, and a larger height rank indicates a larger height of the node in the tree. According to Definition 3.2 in~\cite{shen-etal-2021-structformer}, syntactic height is defined as follows:

\begin{quote}
    Let \( T_d \) be a dependency tree for sentence \( S = (x_1, \ldots, x_n) \), and the height of the node (token) \( x_i \) in \( T_d \) is \( \delta_{i} \). Syntactic heights \( H = (h_1, \ldots, h_{n}) \) are defined as any sequence of \( n \) real scalars that share the same rank as \( (\delta_1, \ldots, \delta_{n}) \).
\end{quote}

\begin{figure}[!htp]
    \centering
    \caption[Syntactic Height Example]{An example of syntactic heights with the corresponding dependency tree. Bluish bars' length indicates the ranking order of heights (i.e., larger bar = larger height). Following Algorithm~\ref{alg:binary_to_dependency}, in each constituent produced from Algorithm~\ref{alg:dtotree}, the ranking order of syntactic heights of the words in this constituent is considered to specify the parent of each node.}
    \label{fig:heights}
    \vspace{1em}
    \includegraphics[width=0.55\linewidth]{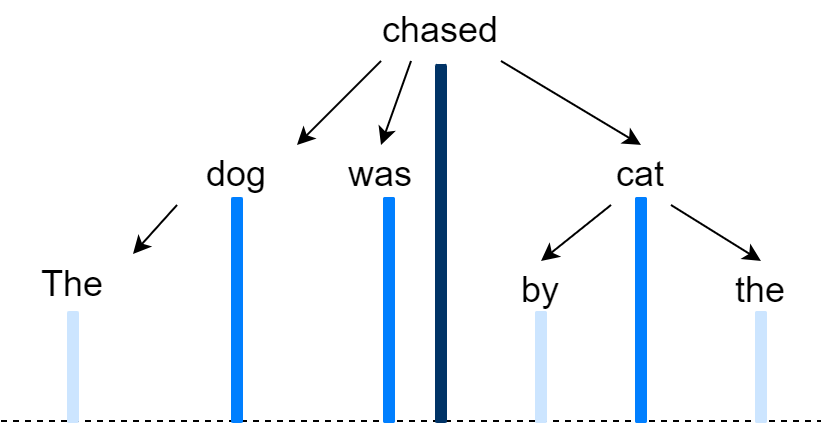}
\end{figure}

While syntactic height is derived through dependency tree structures, reconstructing a dependency tree solely from syntactic heights is infeasible due to the lack of directional information, specifically regarding whether a child node should attach to the left or right side of its parent node. Nevertheless, when provided with syntactic heights in conjunction with an unlabeled constituency tree for a specific sentence, it becomes possible to build a dependency tree for this sentence. This transformation mirrors the established methodology for converting constituency treebanks into dependency treebanks, as delineated by~\cite{10045_1309}. In this approach, instead of relying on constituent labels and part-of-speech (POS) tags to determine the parent node of each constituent, the word exhibiting the greater syntactic height is assigned as the parent. The specifics of this conversion are outlined in Algorithm~\ref{alg:binary_to_dependency}. To illustrate, consider an example where specific arbitrary values of syntactic heights \( H \) are assigned to a sentence \( S \). The resultant configuration, which accurately represents the target dependency tree, is depicted in Figure~\ref{fig:heights}.

\begin{algorithm}[!htp]
\caption{(Minimally edited version of Algorithm 2 in~\cite{shen-etal-2021-structformer} for a better interpretation.) Build Dependency Tree \(T_d\) from Syntactic Heights H and Constituency Tree \(T_c\). \texttt{DEPENDENT} is a recursive function that adds directed edges between tokens (leaves in \(T_c\)) to an empty dependency tree structure \(T_d\), where a token with a larger syntactic height points towards a token with a smaller syntactic height in the produced \(T_d\). \texttt{UNION} merges two trees by adding an edge between the roots of both trees while also following that the root with a larger syntactic height points towards the other root.}
\label{alg:binary_to_dependency}
\begin{algorithmic}[1]
\Function{Dependent}{$H$, $T_c$}
    \If{$T_c = x$}
        \State $T_d \gets []$, $parent \gets x$
    \Else
        \State $child_l, child_r \gets T_c$
        \State $H_l, H_r \gets H$
        \State $T_{d_l}, parent_l \gets \Call{Dependent}{H_l, child_l}$
        \State $T_{d_r}, parent_r \gets \Call{Dependent}{H_r, child_r}$
        \State $T_d \gets \Call{Union}{T_{d_l}, T_{d_r}}$
        \If{$H(parent_l) > H(parent_r)$}
            \State $T_d.\Call{Add}{^{parent_l} \nwarrow _{parent_r}}$
            \State $parent \gets parent_l$
        \Else
            \State $T_d.\Call{Add}{^{parent_r} \nwarrow _{parent_l}}$
            \State $parent \gets parent_r$
        \EndIf
    \EndIf
    \State \Return $T_d, parent$
\EndFunction
\end{algorithmic}
\end{algorithm}

% %%%%%%%%%%%%%%%%%%%%%%%%%%%%%%%%%%%%%%%%%%%%%%%%%%%%%%%%%%%%%%%%%%%%%%

\section{Architecture Components}
\label{sec:comps}

\subsection{Parser Network}

The parser network processes the embeddings of input tokens, \( Z \), yielding real-valued outputs for syntactic distances (\( D \)) and syntactic heights (\( H \)) corresponding to each input token embedding. This transformation from the large dimensionality of the embeddings \(d_{model}\) (e.g., 256 or 512) to only two scalar real values is facilitated through a series of \(L_p\) \texttt{1D convolution} layers acting on the input embedding \( Z \). These convolutions generate embeddings with dimensions identical to the input, but crucially, they convolve sequentially across each word. This convolution considers only a limited contextual window (defined by the kernel size) around each word, producing a modified embedding that encapsulates local information within this restricted neighboring context. The mapping of the embedding dimension to a single scalar value for each token is executed by a linear layer. There are two such linear layers: one for generating a distance value and another for a height value. The transformation of the input embedding \( z_i \) is represented as:

\begin{equation}
z_{l,i} = \tanh\left(\text{Conv}\left(z_{l-1,i-W}, \ldots, z_{l-1,i+W}\right)\right)
\end{equation}

Here, \(z_{l,i}\) denotes the output embedding of token \(i\) from layer \(l\), with \(l\) ranging from 0 to \(L_p\), and \(z_{0,i}\) being the output from the embedding layer. The convolution at token \(i\) considers only the embeddings of tokens within a window of \(W\) positions to the left and right of position \(i\). A linear feed-forward layer is then applied to \(z_{L_p,i}\) and \(z_{L_p,i+1}\) to compute the syntactic distance at position \(i\), as follows:

\begin{equation}
d_i = W_{1_D} \tanh\left( W_{2_D} 
\begin{bmatrix}
z_{L_p,i} \\
z_{L_p,i+1}
\end{bmatrix}
+b_{2_D}
\right) 
+b_{1_D}
\end{equation}

Similarly, a separate linear feed-forward layer is applied to \(z_{L_p,i}\) to determine the syntactic height of token \(i\), as described here:

\begin{equation}
h_i = W_{1_H} \tanh\left( W_{2_H} z_{L_p,i} + b_{2_H} \right) + b_{1_H}
\end{equation}

\subsection{Dependency Function}
\label{dep_funct}

A primary challenge in syntax tree induction from neural-based solutions lies in the inherent nature of tree construction as a hard-decision operation. The natural procedure would involve binary decisions, such as determining whether to split at position \(i\) or not and whether token \(i\) depends on token \(j\) or not. These operations are problematic in neural-based solutions as they disrupt the backpropagation process during training with a zero gradient term. The core objective of the dependency function is to implement parametric, differentiable operations that propagate the unsupervised syntactic knowledge (represented as syntactic distances and heights) from the parser to the multi-head self-attention layers and consequently to the output \(p(x)\). These operations allow for the backpropagation of loss throughout all the architecture's components, eventually optimizing all model parameters to yield the most accurate syntactic distances and heights. The dependency function takes syntactic distances and heights as inputs, and it outputs a dependency matrix \(P_{D}(j|i)\) that indicates the probability of the \(j\)-th token depending on the \(i\)-th token. \(P_{D}(j|i)\) is computed through a series of formulas that are proposed by~\cite{shen-etal-2021-structformer} reflecting hypothesis and assumptions from the parsing research community, and also reflecting the assumptions behind the concepts of syntactic distance and heights, \(P_{D}(j|i)\) is formalized as:

\begin{equation}
\label{eq:main_eq}
    P_{D}(j|i) = 
    \begin{cases} 
        \displaystyle\sum_{[l,r]} \kern0.5em \textbf{p}_C([l,r]|i) \kern0.3em \textbf{p}_{PN}(j|[l,r])   & \text{if } i \neq j \\
            0 & \text{if } i = j 
    \end{cases}
\end{equation}

We look at each term separately to understand the calculation of \(P_D(j|i)\). The first term in Equation~\ref{eq:main_eq}, \(\textbf{p}_C([l,r]|i)\), represents the probability that \([l,r]\) is the smallest legal constituent \(C(x_i)\) encompassing \(x_i\) and \(x_i\) is not the parent of \(C(x_i)\) in a dependency tree. \(\textbf{p}_C([l,r]|i)\) is computed according to Equation~\ref{eq:abc1}, where \(p(l|i)\) is the distribution of \(l\) for token \(i\) over positions \(<i\). \(p(l|i)\) is modelled as shown in Equation~\ref{eq:l1}. This approach of modeling \(p(l|i)\) measures the likelihood of a breaking point within indices \(<i\), where the constituent is presumed to start. The likelihood of a token belonging to \(C(x_i)\) is assessed sequentially backward according to Equations~\ref{eq:l2}, and \ref{eq:l3}. In Equation~\ref{eq:l2}, the probability that the \(l\)-th token (\(l < i\)) falls within \(C(x_i)\) correlates with the likelihood that \(h_i\) exceeds the maximal distance \(d\) between \(l\) and \(i\). To simplify and enhance computational efficiency, the cumulative distribution in Equation~\ref{eq:l2} is parameterized in Equation~\ref{eq:l3} using the sigmoid function \(\sigma\), with \(\mu_1\) being a learnable temperature parameter. Equations~\ref{eq:l1}, \ref{eq:l2}, and \ref{eq:l3} are similarly applied for defining \(p(r|i)\) in the opposite direction. 

\begin{equation}
\label{eq:abc1}
    \textbf{p}_C([l,r]|i) = 
    \begin{cases} 
        \textbf{p}(l|i) \kern0.5em \textbf{p}(r|i) & l \leq i \leq r \\
        0 & \text{otherwise}
    \end{cases}
\end{equation}

\begin{equation}
\label{eq:l1}
    p(l|i) = p(l \in C(x_i)) \kern0.5em -  \kern0.5em p(l - 1 \in C(x_i))
\end{equation}

\begin{equation}
\label{eq:l2}
    \textbf{p}(l \in C(x_i)) = \textbf{p}(h_i > \max(d_l, \ldots, d_{i-1}))
\end{equation}
\begin{equation}
\label{eq:l3}
    \textbf{p}(l \in C(x_i)) = \sigma\left(\frac{h_i - \max(d_l, \ldots, d_{i-1})}{\mu_1}\right)
\end{equation}

While the second term in Equation~\ref{eq:main_eq} \(\textbf{p}_{PN}(j|[l,r])\) denotes the probability of \(j\) being a Parent Node (PN) in a dependency tree concerning the interval \([l,r]\), identified as the token with the largest height in \([l,r]\). This probability is parameterized using a softmax function as shown in Equation~\ref{eq:abc3}, reflecting the main idea behind syntactic height and Algorithm~\ref{alg:binary_to_dependency}.

\begin{equation}
\label{eq:abc3}
    \textbf{p}_{PN}(j|[l,r]) = 
    \begin{cases} 
    \displaystyle{\frac{\exp(h_j/\mu_2)}{\sum_{l \leq k \leq r} \exp(h_k/\mu_2)}} & \text{if } l \leq t \leq r \\
    0 & \text{otherwise}
    \end{cases}
\end{equation}

Eventually, as demonstrated, the dependency function~\ref{eq:main_eq} is differentiable and maintains the integrity of the backpropagation operation. The output of this function, \(P_{D}(j|i)\), represents the dependency probability between all tokens in the input sentence \(S\).

\subsection{Dependency-Constrained Multi-Head Self-Attention Layers}

To effectively propagate syntactic information from the parser network and the dependency function towards the language modeling objective, the multi-head self-attention mechanism in the vanilla Transformer~\cite{vaswani17} is utilized, albeit with a modification to incorporate the dependency matrix \(P_{D}(j|i)\). In StructFormer, the standard scaled dot-product attention operation of Transformer, as represented in Equations~\ref{eq:att111} and \ref{eq:att112}, is supplanted by a dependency-constrained self-attention mechanism, as delineated in Equations~\ref{eq:att121} and \ref{eq:att122}. In this modification, the softmax function is replaced by a sigmoid function, converting it into an independent probability that signifies whether \(x_i\) should attend to \(x_j\) in the context of the current attention head. The dependency matrix \(P_{D}\) is integrated into the original attention mechanism, reinforcing the tokens' syntactic dependency relations within the attention process.

\begin{equation}
\label{eq:att111}
\text{Attention}_{\text{tf}}(Q_i, K_j, V_j) = q_{i,j}V_j
\end{equation}

\begin{equation}
\label{eq:att112}
where \kern0.4em q_{i,j} = \textit{softmax}\left(\frac{QK^T}{\sqrt{d_k}}\right)
\end{equation}

\begin{equation}
\label{eq:att121}
\text{Attention}_{\text{sf}}(Q_i, K_j, V_j, P_D) = p_{i,j} \kern0.3em q_{i,j} \kern0.3em V_j 
\end{equation}

\begin{equation}
\label{eq:att122}
where \kern0.4em q_{i,j} = \textit{sigmoid}\left(\frac{QK^T}{\sqrt{d_k}}\right)
\end{equation}

Given that the dependency matrix \(P_{D}\) lacks directional specificity—i.e. it does not indicate whether the relation is parent-to-child or child-to-parent—both directions are considered: \(P_{D}(j|i)\) represents the scenario where token \(i\) is the parent of token \(j\), and \(P_{D}(i|j)\) (its transpose) for token \(j\) being the parent of token \(i\). Furthermore, \(P_{D}\) is multiplied with different weights in each attention head, indicating the strength of each relation (parent or child) in that specific head. This is represented in the term \(p_{i,j}\) in Equation~\ref{eq:att121}. Fomrally, \(p_{i,j}\) is computed according to Equation~\ref{eq:pij}. Here, \(p_{\text{parent}}\) and \(p_{\text{child}}\) signify the probabilities that a particular attention head will propagate information in the parent-to-child and child-to-parent directions, respectively. They are parameterized as shown in Equations~\ref{eq:pppc1} and \ref{eq:pppc2}. Where \(w_{\text{parent}}\) and \(w_{\text{child}}\) are learnable weights associated with each attention head. The model learns to assign these weights based on the masking language modeling objective.

\begin{equation}
\label{eq:pij}
    p_{i,j} = p_{\text{parent}}P_D(j|i) + p_{\text{child}}P_D(i|j)
\end{equation}

\begin{equation}
\label{eq:pppc1}
P_{\text{parent}} = \frac{\exp(w_{\text{parent}})}{\exp(w_{\text{parent}}) + \exp(w_{\text{child}})}
\end{equation}

\begin{equation}
\label{eq:pppc2}
    P_{\text{child}} = \frac{\exp(w_{\text{child}})}{\exp(w_{\text{parent}}) + \exp(w_{\text{child}})}
\end{equation}

%%%%%%%%%%%%%%%%%%%%%%%%%%%%%%%%%%%%%%%%%%%%%%%%%%%%%%%%%%%%%%%%%%%%%%%

% \vspace{2em}

\section{Implementation Details}
\label{sec:implem}

In this section, I detail the experimental setup as reported in the original publication of StructFormer~\cite{shen-etal-2021-structformer}. The authors pretrained the StructFormer model on two datasets: the Penn Tree Bank (PTB)~\cite{marcus-etal-1993-building} and the Brown Laboratory for Linguistic Information Processing (BLLIP)~\cite{bllip}. My reproduction, however, focuses solely on the PTB dataset due to its prevalent use in syntax tree parsing evaluation within the existing literature.

% ~\cite{Uen2012StatisticalLM}

\textbf{Dataset:} The PTB is a benchmark dataset for language modeling and unsupervised constituency parsing~\cite{onlstm, kim-etal-2019-compound}. The PTB project curated 2,499 stories from a three-year collection of 98,732 Wall Street Journal (WSJ) stories for syntactic annotation. This effort yielded approximately 7 million words of POS-tagged text, 3 million words of structured parsed text, and over 2 million words parsed for predicate-argument structure. The standard splits are adopted in this implementation: sections 0 to 21 for training, section 22 for validation, and section 23 for testing. The dataset undergoes common filtering and preprocessing steps, specifically following the procedure outlined in~\cite{Uen2012StatisticalLM}, which includes the removal of all punctuation and the substitution of low-frequency tokens with a unique token (\textit{<unk>)}.

\textbf{Tokenizer:} The tokenization of the PTB dataset employs a word frequency-based approach consistent with the common methodology in the literature for syntax structure induction. This approach results in a vocabulary of 10,001 tokens, inclusive of the unique tokens: \textit{<unk>} token replacing out-of-vocabulary words, \textit{<pad>} token to fill input sequences ensuring the same length across all sequences in an input batch, and \textit{<mask>} token to replace random tokens during training. As a common preprocessing step in this problem, numerical digits are replaced with an \texttt{N} character.

\textbf{Pretraining:} The implementation of StructFormer closely parallels the original Transformer encoder~\cite{vaswani17}, with the modifications specified in section~\ref{sec:comps}. Additionally, the authors introduced a normalization layer at the beginning of each layer, akin to the T5 model~\cite{raffel-etal}. This adjustment was found to accelerate model convergence. The hyperparameters settings are set as follows: the number of layers \( L = 8 \), the embedding dimension size \( d_{\text{model}} = 512 \), the number of self-attention heads \( h = 8 \), the feed-forward size \( d_{\text{ff}} = 2048 \), a dropout rate of 0.1, and the number of convolution layers in the parser network \( L_p = 3 \). The pretraining batch size is set to \( bsz = 4096 \) tokens. The model undergoes training with a language masking objective for 100 epochs while only saving the model with the smallest loss value on the validation dataset. The masking rate per input sentence is set at 0.3.

%%%%%%%%%%%%%%%%%%%%%%%%%%%%%%%%%%%%%%%%%%%%%%%%%%%%%%%%%%%%%%

\section{Visualized Example}
\label{sec:example}

In this section, I provide an illustrative example from our reproduction experiments to break down the process of inducing syntactic trees from an input sentence using StructFormer. This includes the initial preprocessing steps, the flow inside the parser network, and the construction of the corresponding constituency and dependency trees. Eventually, I detail the method for evaluating these constructed trees against the reference trees.

\textbf{Input Sentence:} A sample sentence was chosen from the PTB test set, with a special attention to ensure the sentence length is manageable for visualizations. The sentence selected for analysis is:

\begin{equation}
\label{eq:sen1_pre}
    \textit{Superconcentrates aren't entirely new for P\&G.}
\end{equation}

\begin{figure}[!htp]
    \centering
    \caption[Case Study: Human-annotated Constituency Tree]{Human-annotated Constituency Tree for sentence~\ref{eq:sen1_pre} as found in the PTB dataset.}
    \label{fig:ex_const_pre}
    \includegraphics[width=0.55\linewidth]{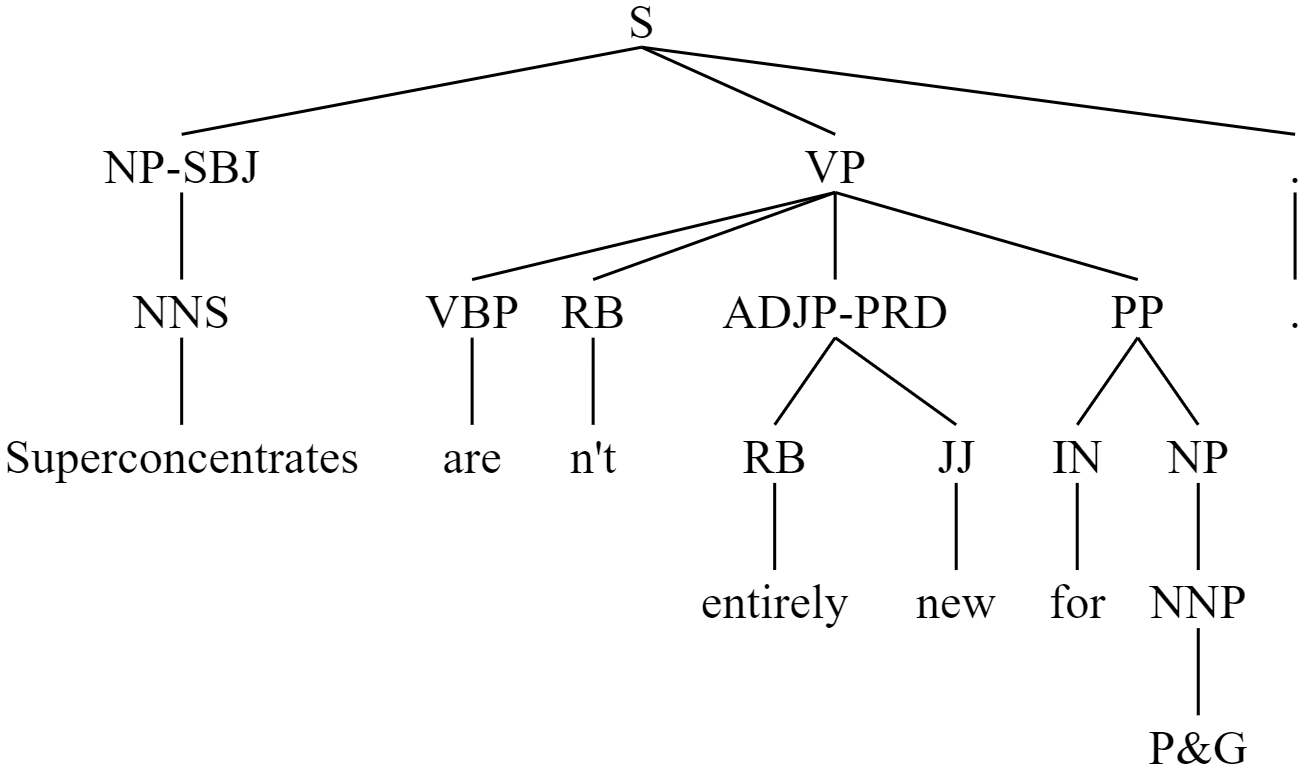}
    
\end{figure}

\begin{figure}[!htp]
    \centering
    \caption[Case Study: Reference Dependency Tree 1]{Automatic-annotated Dependency Tree for sentence~\ref{eq:sen1_pre} as found in the \(Dep_{auto}\) annotations for the PTB sentences.}
    \label{fig:ex_dep_pre}
    \includegraphics[width=0.75\linewidth]{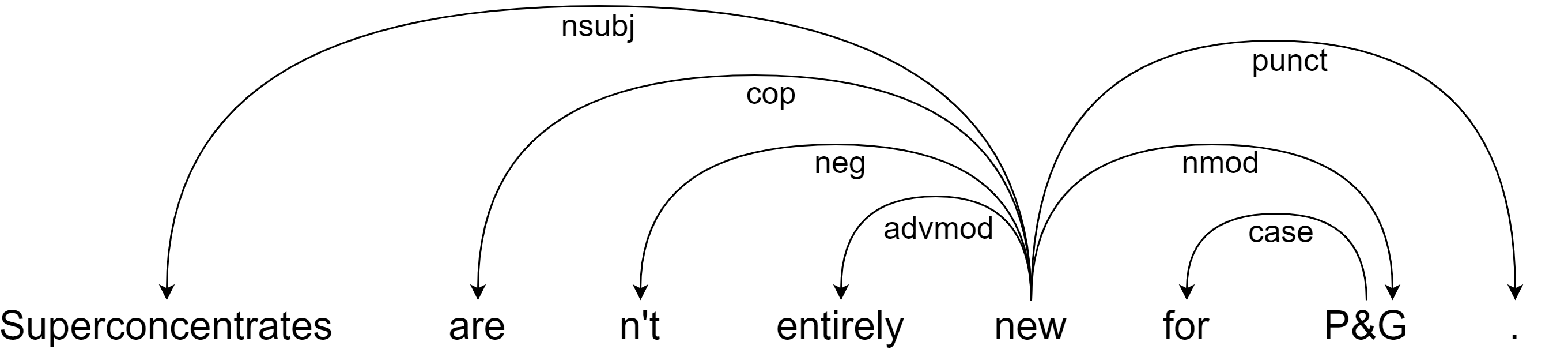}
    
\end{figure}

\begin{figure}[!htp]
    \centering
    \caption[Case Study: Reference Dependency Tree 2]{Automatic-annotated Dependency Tree for sentence~\ref{eq:sen1_pre} generated from spaCy Dependency Parser based on~\cite{honnibal-johnson-2015-improved}.}
    \label{fig:ex_dep_pre_stanza}
    \includegraphics[width=0.75\linewidth]{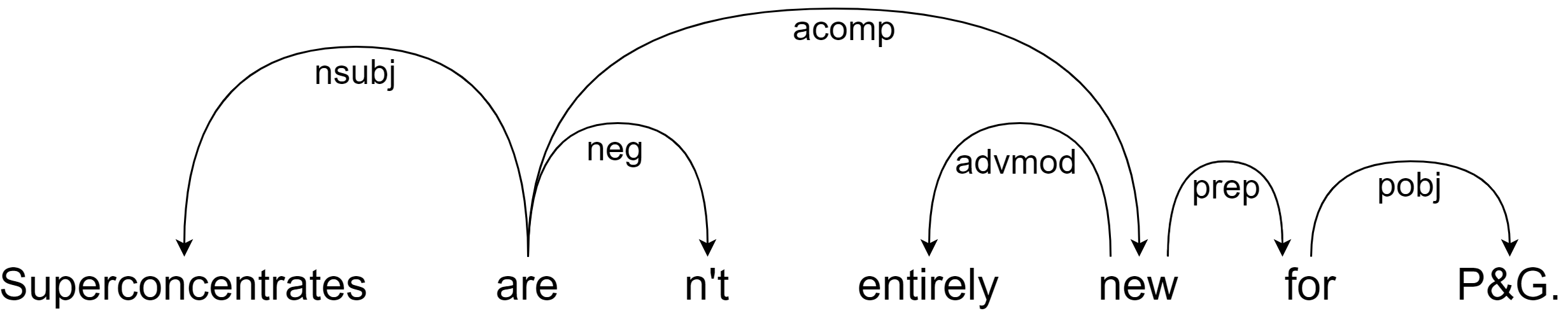}
    
\end{figure}

\textbf{Reference Trees:} The human-annotated constituency tree of the sentence in the PTB dataset is visualized in Figure~\ref{fig:ex_const_pre}. We use automatic-annotated reference trees for annotated dependency trees converted from the PTB human-annotated constituency parses using the specified algorithms in~\cite{de-marneffe-etal-2006-generating}. In our discussion, we refer to these reference trees dataset as \(Dep_{auto}\). The reference dependency tree for the selected sentence is depicted in Figure~\ref{fig:ex_dep_pre}. It was noticed that some annotations in \(Dep_{auto}\) include linguistically incorrect relations. However, we only consider the \(Dep_{auto}\) dataset in our experiments despite the spotted mistakes, as it follows the same tokenization procedure used in almost all the published studies that use the PTB dataset, including the original implementation of StructFormer. To have an example of how reference trees can vary for the same sentence according to the applied annotation procedures, a dependency tree annotated by the spaCy dependency parser\footnote{\url{https://spacy.io/api/dependencyparser}} is illustrated in Figure~\ref{fig:ex_dep_pre_stanza}. 

\begin{figure}[!htp]
    \centering
    \caption[Case Study: Preprocessed Reference Constituency Tree]{Preprocessed Reference Constituency Tree for sentence~\ref{eq:sen1_post} after performing the preprocessing steps on~\ref{fig:ex_const_pre}. Each "X" represents an unlabelled constituent.}
    \label{fig:ex_const_post}
    \vspace{0.4em}
    \includegraphics[width=0.5\linewidth]{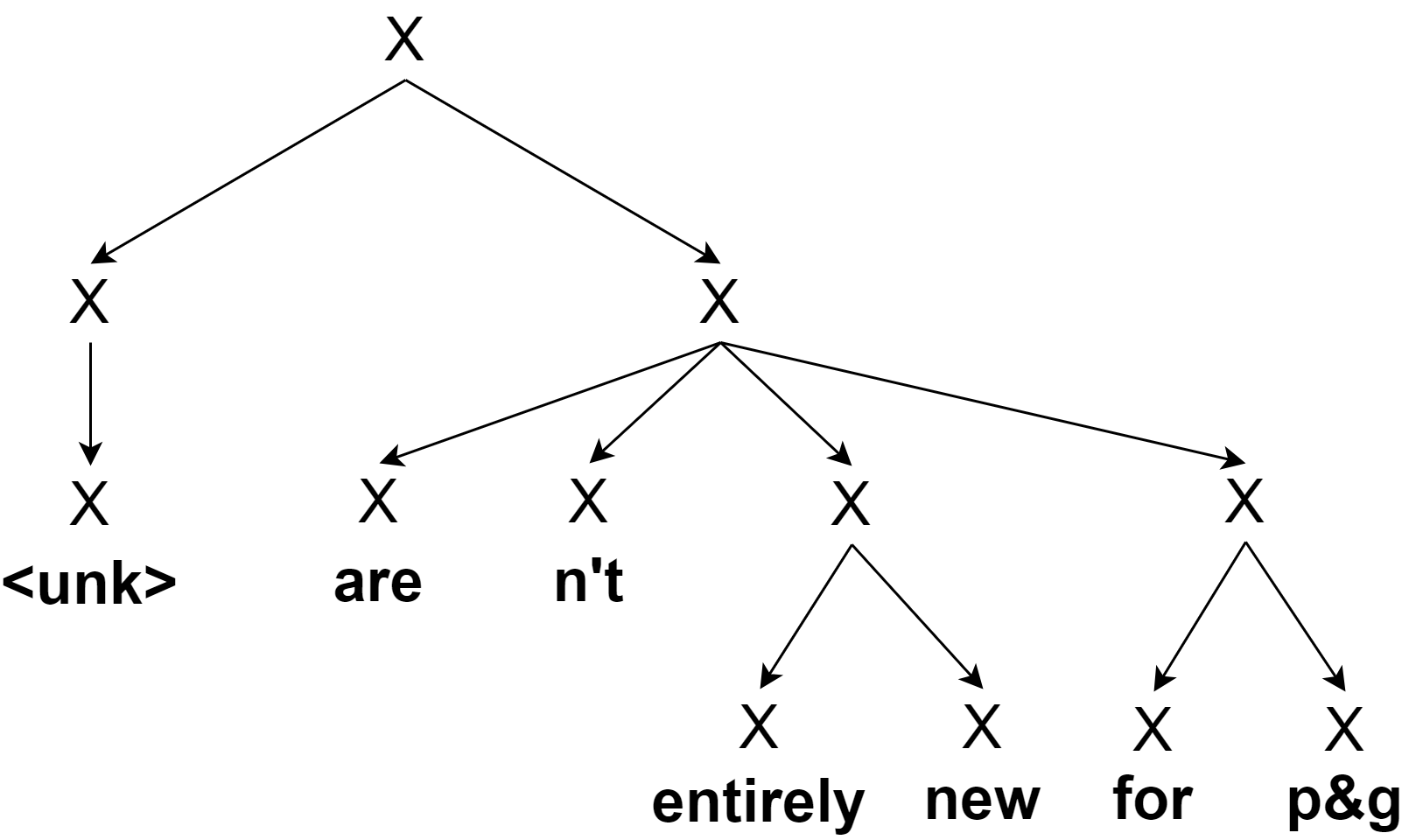}
    
\end{figure}

\begin{figure}[!htp]
    \centering
    \caption[Case Study: Preprocessed Human-annotated Dependency Tree]{Preprocessed Human-annotated Dependency Tree for sentence~\ref{eq:sen1_post} after performing the preprocessing steps on~\ref{fig:ex_dep_pre}. Each "X" represents an unlabelled relation.}
    \label{fig:ex_dep_post}
    \includegraphics[width=0.65\linewidth]{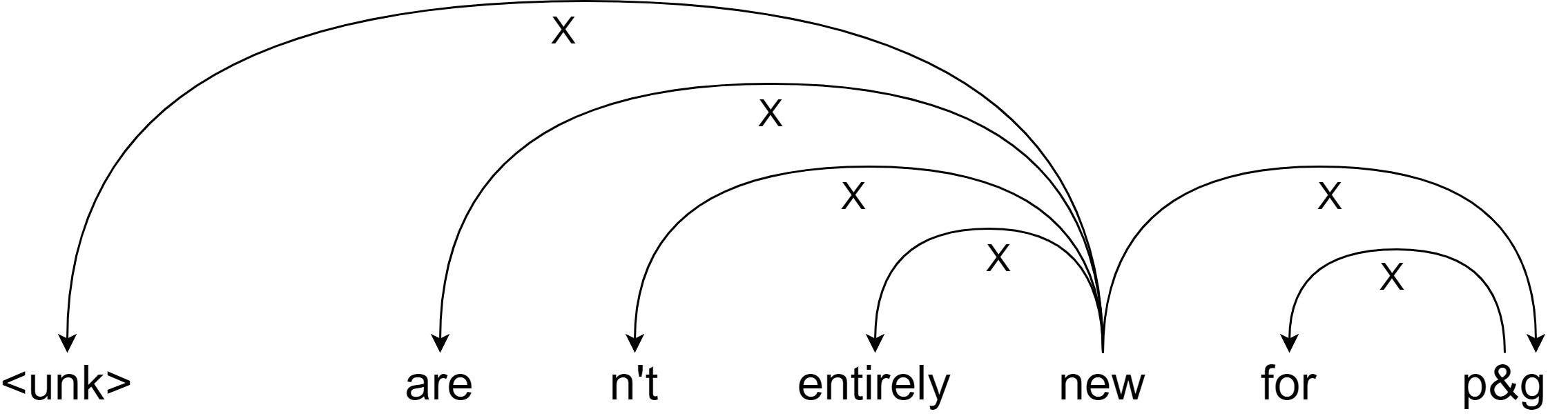}
\end{figure}

\textbf{Preprocessing Procedure:} The preprocessing steps include 1) Tokenization, where the tokenization algorithm~\ref{sec:implem} is applied to the input sentence and reference trees. Tokenizing the reference trees is performed simply by replacing the out-of-vocabulary words with an \textit{<unk>} token and replacing any digit with an \textit{N} character. The split words (as in \textit{aren't --> are n't}) are already present in the reference trees. The word-based tokenizer has the advantage of having such a simple tokenization process that can easily be applied to trees of words. 2) Remove all the punctuation marks from the input sentence and reference trees. 3) Apply the lower-case to the input sentence and reference trees. 4) Replace the tree labels with a placeholder (X). After the preprocessing steps, the sentence is formed as shown in~\ref{eq:sen1_post}, and this form serves as the input to the model. Correspondingly, the target constituency and dependency trees (after performing the preprocessing steps) are visualized in Figures~\ref{fig:ex_const_post} and \ref{fig:ex_dep_post}, showcasing the structures the model aims to replicate.

\begin{equation}
\label{eq:sen1_post}
    \textit{<unk> are n't entirely new for p\&g}
\end{equation}

\begin{figure}[!htp]
    \centering
    \caption[Case Study: Predicted Syntactic Distances and Heights]{Predicted Syntactic Distance and Heights for sentence~\ref{eq:sen1_post}}
    \label{fig:ex_pred_d_h}
    \includegraphics[width=0.7\linewidth]{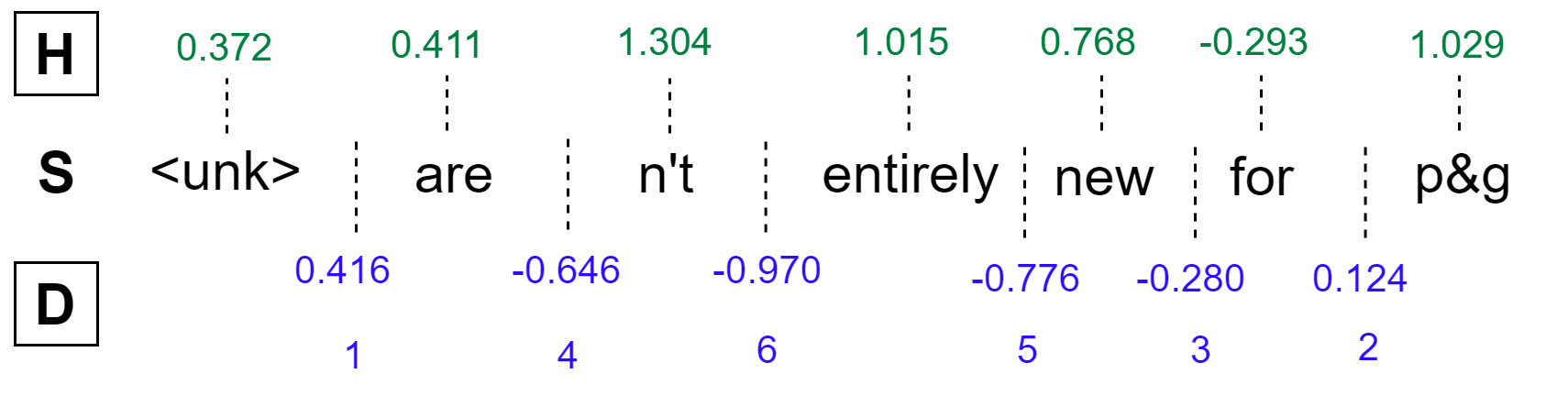}
    
\end{figure}

\begin{figure}[!htp]
    \centering
    \caption[Case Study: Predicted Dependency Matrix]{Predicted Dependency Matrix \(P_D(j|i)\) for sentence~\ref{eq:sen1_post}. Values in each row indicate the likelihood of each token serving as a parent to the token corresponding to that row. E.g. In the first row, \texttt{p\&g} is the most likely parent to the \texttt{<unk>} token.}
    \label{fig:ex_pred_pd}
    \includegraphics[width=0.75\linewidth]{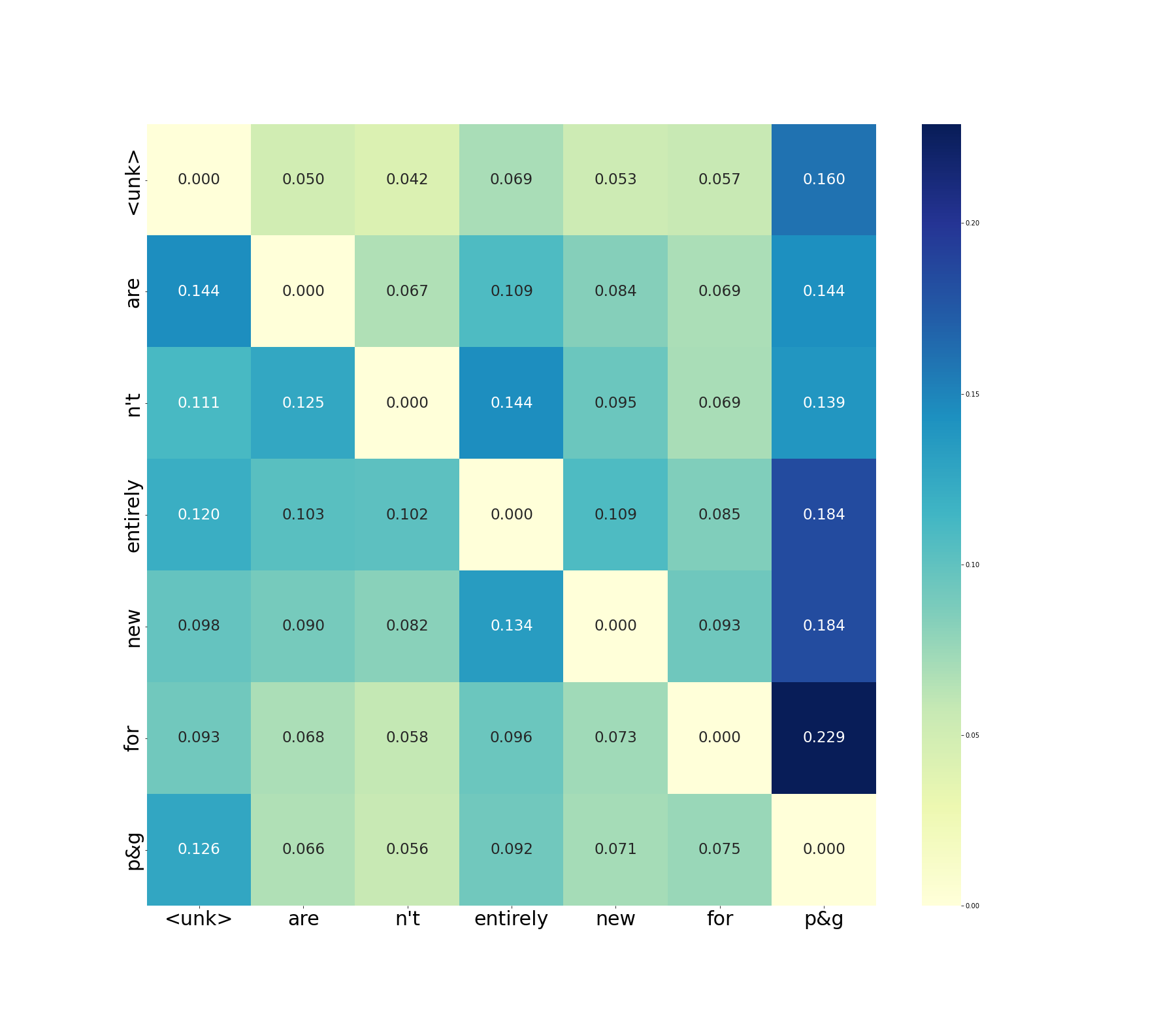}
    
\end{figure}

\textbf{Predicted Distance and Height:} In Figure~\ref{fig:ex_pred_d_h}, the distances and heights predicted by the parser module of a pretrained model are visualized. Furthermore, Figure~\ref{fig:ex_pred_pd} depicts the actual \(P_D(j|i)\) values, derived from the dependency function, providing insights into the model's internal values that are the basis for the induced trees eventually.

\begin{figure}[!htp]
    \centering
    \caption[Example Study: Induced Constituency Tree by StructFormer]{Induced Constituency Tree by StructFormer for sentence~\ref{eq:sen1_post}. Each "X" represents a predicted unlabelled constituent}
    \label{fig:ex_pred_const}
    \vspace{1em}
    \includegraphics[width=0.33\linewidth]{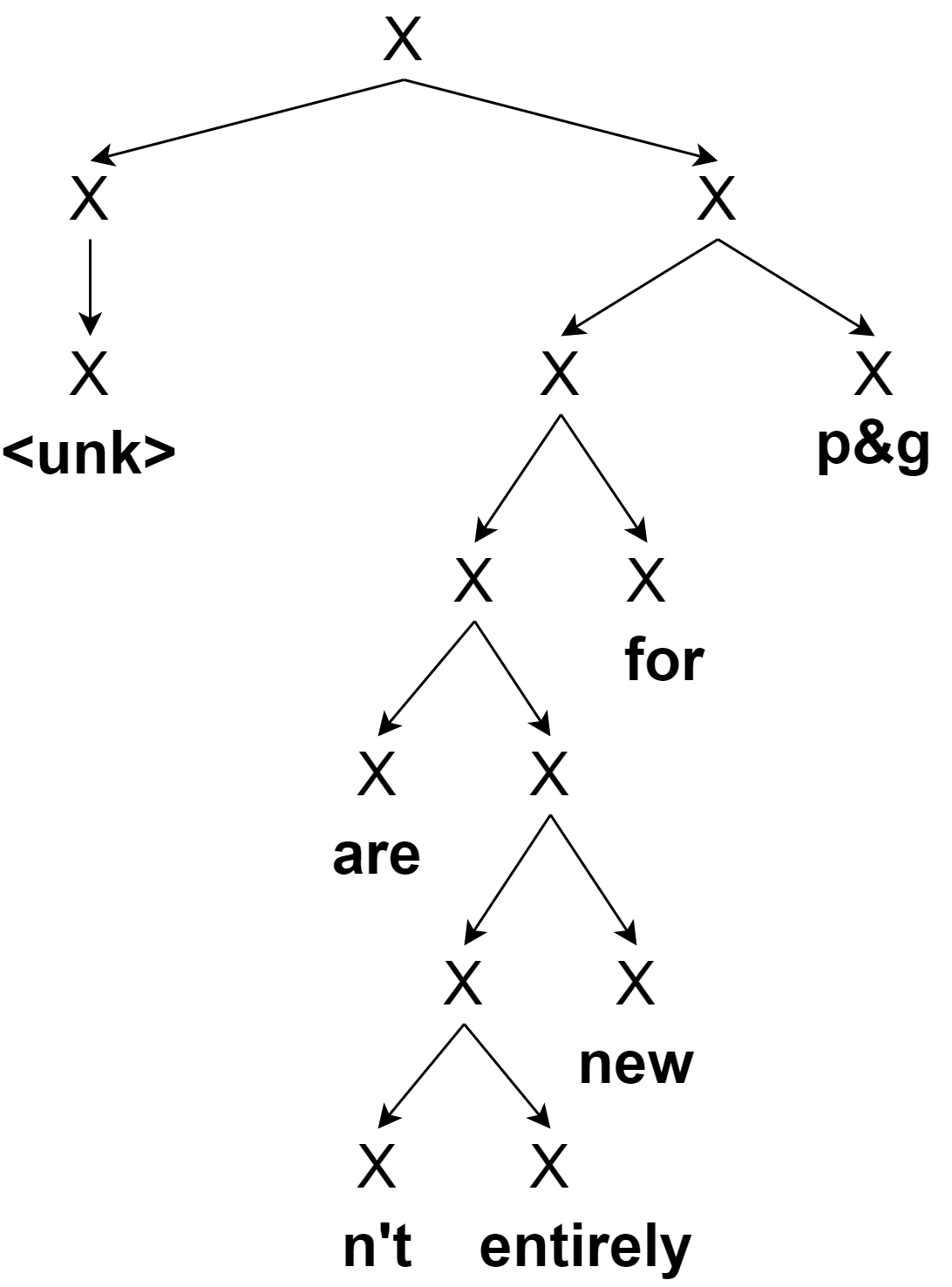}
    
\end{figure}

\begin{figure}[!htp]
    \centering
    \caption[Example Study: Induced Dependency Tree by StructFormer]{Induced Dependency Tree by StructFormer for sentence~\ref{eq:sen1_post}}
    \label{fig:ex_pred_dep}
    \vspace{1em}
    \includegraphics[width=0.65\linewidth]{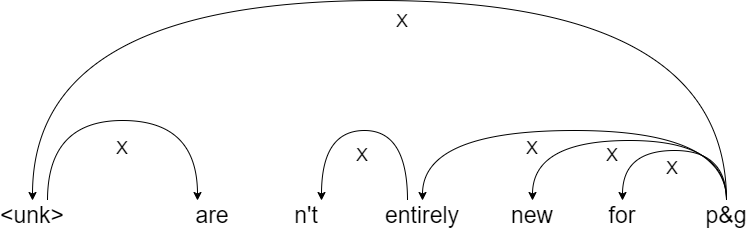}
    
\end{figure}

\textbf{Induced Trees:} Leveraging the predicted distances and heights values in accordance with Algorithms ~\ref{alg:dtotree} and \ref{alg:binary_to_dependency}, we induce an unlabeled constituency tree and a dependency tree. These are presented in Figures~\ref{fig:ex_pred_const} and~\ref{fig:ex_pred_dep}, demonstrating the transformation process of the algorithms in generating syntactic structures from model predictions.

\textbf{Trees Evaluation:} For unsupervised constituency parsing, unlabelled precision, recall, and F1 scores are calculated. In the target constituency tree for our example (Figure~\ref{fig:ex_const_post}), there are four constituents: \texttt{(<unk>)}, \texttt{(are n't entirely new for p\&g)}, \texttt{(entirely new)}, and \texttt{(for p\&g)}. Conversely, in the predicted constituency tree (Figure~\ref{fig:ex_pred_const}), six constituents are identified as follows: \texttt{(<unk>)}, \texttt{(are n't entirely new for p\&g)}, \texttt{(are n't entirely new for)}, \texttt{(are n't entirely new)}, \texttt{(n't entirely new)}, and \texttt{(n't entirely)}. Two target constituents were accurately induced, but four additional constituents were identified but are not part of the target structure. Therefore, for this example, the unlabelled recall \(R = \frac{2}{4} = 0.5\), precision \(P = \frac{2}{6} = 0.33\), and \(UF1 = \frac{2 \times 0.5 \times 0.33}{0.5 + 0.33} = 0.39\). For unsupervised dependency parsing, UAS is calculated. Upon comparing the target dependency tree (Figure~\ref{fig:ex_dep_post}) and the predicted dependency tree by StructFormer (Figure~\ref{fig:ex_pred_dep}), it is observed that only one out of the seven tokens in the sentence correctly traced back to its parent in the predicted tree. Thus, for this example, \(UAS = \frac{1}{7} = 0.14\).

%%%%%%%%%%%%%%%%%%%%%%%%%%%%%%%%%%%%%%%%%%%%%%%%%%%%%%%%%%%%

\section{Evaluation Results}
\label{sec:eval}

The StructFormer model is evaluated on three tasks: masked language modeling, unsupervised constituency parsing, and unsupervised dependency parsing. We have replicated the original authors' experiment with identical settings. This section reports both the results from the original publication and our reproduction.

\textbf{Masked Language Modeling:} Perplexity is one of the most common metrics used to evaluate the fluency of a language model~\cite{JurafskyMartin2024}. It is a way to capture the uncertainty or surprise of a model when predicting new data. The lower the perplexity, the better the model is at making predictions. \textit{Perplexity} is defined as the exponential of the average negative log-likelihood of a sequence. For a tokenized sentence \( S = (x_1, x_2, \ldots, x_n) \), the perplexity of \( S \) is defined as in Equation~\ref{eq:ppl_ar}.
\begin{equation}
\label{eq:ppl_ar}
    PPL(\mathbf{S}) = \exp \left( -\frac{1}{n} \sum_{i}^{n} \log p_{model}(x_i | x_{<i}) \right)
\end{equation}
 
Perplexity can be clearly computed for an Autoregressive model, as the core objective for autoregressive models aligns with the metric definition, which measures the probability of predicting a token based on a sequence of tokens preceding this certain token in a sentence. Masked language models' objective, on the other hand, aims at predicting a certain token based on all the other tokens\footnote{Some of the other tokens in the context during prediction can be masked as well.} in the sentence (either preceding or following this certain token). This MLM objective makes perplexity not well-defined for masked language models. \cite{salazar-etal-2020-masked} proposed an adaptation for measuring perplexity for MLMs, and \cite{kauf-ivanova-2023-better} recently criticized this adaptation and proposed another adaptation. Despite these efforts, measuring perplexity for MLMs is still not standardized, and many important publications report their computed perplexities without reporting the details of their adaptations as in the very popular paper~\cite{devlin-etal-2019-bert}. The recent paper~\cite{wettig-etal-2023-mask} also do not detail their adaptation, and surprisingly, they cite~\cite{devlin-etal-2019-bert} as their reference for calculating MLM perplexity, although~\cite{devlin-etal-2019-bert} did not detail their adaptation in the first place. Our study subject StructFormer~\cite{shen-etal-2021-structformer} also reports perplexity results on their models without detailing the adaptation procedure. However, their original implementation codebase~\footnote{\url{https://github.com/google-research/google-research/tree/master/structformer}} revealed their adaptation for computing MLM perplexity. We use the same adaptation in all our experiments to enable fairly comparable experiments in this thesis. Hence, our perplexity results are only comparable within the scope of this thesis and should not be compared to perplexity results from other studies. Our perplexity results should only be interpreted as an estimate of a masked language model's fluency.

In both the experiments conducted by StructFormer~\cite{shen-etal-2021-structformer} and those presented in this thesis, perplexity is adapted to evaluate masked language models. Specifically, for a sentence \(S_m = (x_1, x_2, \ldots, x_n)\) that has been tokenized and masked, a subset of tokens \(x_j\) is selected based on \(m\) indices. These indices are chosen randomly\footnote{The selection process for these indices follows a Bernoulli distribution with a probability parameter of \(p=0.3\).} from the sequence's range \(1,\ldots,n\), and each selected token is replaced by a \texttt{<mask>} token. Here, \(x_j\) denotes the original token at each selected index \(j\). The adapted perplexity for masked language modeling denoted as \(PPL_{MLM}\), is then calculated using Equation~\ref{eq:ppl_mlm}.

\begin{equation}
\label{eq:ppl_mlm}
PPL_{MLM}(S_m) = \exp\left( -\frac{1}{m} \sum_{j=1}^{m} \log p_{\text{model}}(x_j | S_m) \right)
\end{equation}

To benchmark StructFormer's masked language modeling performance, StructFormer is compared against a vanilla transformer encoder~\cite{vaswani17} under similar pretraining conditions. The results, detailed in Table~\ref{tab:ppl}, demonstrate StructFormer's consistent superiority over the Transformer baseline. This result aligns with prior observations that linguistically informed self-attention enhances Transformer performance, and StructFormer also exhibits a faster convergence rate. The reproduction results are based on five runs of the pretraining process with the same settings each time. Reproduction results confirm the outperformance trend of the StructFormer against the vanilla Transformer. However, it does not produce the exact same values despite using the same setting in all aspects.

\begin{table}[!htp]
\centering
\caption[StructFormer Results: Masked Language Modeling]{Masked tokens perplexity results on PTB test set for a baseline vanilla transformer against the StructFormer architecture (reporting the original results from~\cite{shen-etal-2021-structformer}) and the mean and best values of the reproduction experiment.}
\label{tab:ppl}
\begin{tabular}{l|c|ccc}
\hline
Model  & params \(\times 10^6\)    & \cite{shen-etal-2021-structformer}   & Reproduction\(_{mean}\) & Reproduction\(_{best}\) \\ \hline
Transformer & 30.6 & 64.05 &   67.08     & 65.41         \\
StructFormer & 38.5 & 60.94 & 65.85 & 64.59  \\ \hline
\end{tabular}

\end{table}

\textbf{Unsupervised Constituency Parsing:} This task assesses the model's ability to induce latent constituency tree structures and compares them with human-annotated constituency trees in the PTB dataset~\cite{marcus-etal-1993-building}, using Algorithm~\ref{alg:dtotree} for tree construction from predicted syntactic distances by StructFormer. The evaluation follows the standard settings in~\cite{htut-etal-2018-grammar-induction}, testing on the WSJ test set comprising 2,416 expert-labeled sentences, excluding punctuation. Unlabeled F1 (UF1) scores are reported in Table~\ref{tab:uf1}, comparing StructFormer's performance with other architectures as documented in their respective publications. Reproduction results are also reported based on five model runs with the same settings.

\begin{table}[!htp]
\centering
\caption[StructFormer Results: Unsupervised Constituency Parsing]{Unlabeled F1 (UF1) scores of random baselines (upper part), most popular unsupervised methods (middle part), and StructFormer (both original and reproduction) (bottom part). Precision and Recall are computed in our reproduction only. Baseline and common methods results are from~\cite{kim2020} }
\label{tab:uf1}
\begin{tabular}{lccc}
\toprule
Method & UF1 & Precision & Recall  \\
\midrule
RANDOM & 21.6  \\
LBRANCH & 9.0  \\
RBRANCH & 39.8  \\
\midrule
PPRN~\cite{prpn} & 37.4 (0.3)  \\
ON-LSTM~\cite{onlstm} & 47.7 (1.5)  \\
Tree-T~\cite{wang-etal-2019-tree} & 49.5  \\
URNNG~\cite{kim-etal-2019-unsupervised} & 52.4  \\
C-PCFG~\cite{kim-etal-2019-compound} & 55.2 \\
Neural L-PCFGs~\cite{zhu-etal-2020-return}  & 55.31  \\
\midrule
StructFormer~\cite{shen-etal-2021-structformer} & 54.0 (0.3)  \\
StructFormer\(_{reproduction}\) & 51.9 (1.85) & 45.5 (1.6) & 62.62 (2.2)  \\ 
\bottomrule
\end{tabular}
\end{table}

\textbf{Unsupervised Dependency Parsing:} Similar to constituency parsing, the model's performance in dependency parsing is evaluated on the PTB test set (section 23). Reference dependency trees, derived from human-annotated constituency trees, vary based on conversion rules. Unlike the original paper, which considers Stanford and CoNLL dependencies, our reproduction focuses solely on the \(Dep_{auto}\) reference trees due to their availability. Punctuation is disregarded during evaluation, as per~\cite{jiang-etal-2016-unsupervised}. This evaluation measures the accuracy of induced dependency relations against reference trees using the Unlabeled Attachment Score (UAS). Results are compared with established methods in the field in Table~\ref{tab:uas}. The reproduction results are not comparable to the other results as they use different reference trees during evaluation.

\begin{table}[!htp]
\centering
\caption[StructFormer Results: Unsupervised Dependency Parsing]{Unlabeled Attachment Score (UAS) of most popular unsupervised methods (upper part), and StructFormer (both original and reproduction) (bottom part). Baseline and common methods results are from~\cite{he-etal-2018-unsupervised}}
\label{tab:uas}
\begin{tabular}{lc}
\toprule
Method & UAS \\
\midrule
DMV~\cite{klein-manning-2004-corpus} & 35.8 \\
E-DMV~\cite{headden-iii-etal-2009-improving} & 38.2 \\
UR-A E-DMV~\cite{tu-honavar-2012-unambiguity} & 46.1 \\
Neural E-DMV~\cite{jiang-etal-2016-unsupervised} & 42.7 \\
Gaussian DMV~\cite{he-etal-2018-unsupervised} & 43.1 (1.2) \\
INP~\cite{he-etal-2018-unsupervised} & 47.9 (1.2) \\
Neural L-PCFGs~\cite{zhu-etal-2020-return} & 40.5 (2.9) \\
\midrule
StructFormer~\cite{shen-etal-2021-structformer} - Stanford & 46.2 (0.4) \\
StructFormer\(_{reproduction}\) - UD & 37.38 (9)   \\
\bottomrule
\end{tabular}
\end{table}

%%%%%%%%%%%%%%%%%%%%%%%%%%%%%%%%%%%%%%%%%%%%%%%%%%%%%%%%%%%%%%%%%%%%%%%%%%%%%%%%%%%
\section{Self-Consistency}
\label{sec:self_consist}

In our replication study, we conducted an additional experiment to assess the consistency of the induced trees across five runs. This experiment aims to measure the model's reliability in generating similar structures across different iterations, a criterion crucial for confirming the model's proficiency in discerning distinct linguistic patterns and syntactic knowledge. The outcomes of this investigation are detailed in Table~\ref{tab:self_consist}. Results of a similar self-consistency experiment are reported in~\cite{williams-etal-2018-latent}; their experiment is conducted on a different dataset. However, we can deduce from their discussion that a self-consistency F1 of more than 65\% is considered \textit{relatively high} in this domain.

\begin{table}[!htp]
\centering
\caption[StructFormer Results: Self-Consistency]{Self-Consistency results of the induced trees by our reproduction runs. Such an experiment was not reported in the original paper.}
\label{tab:self_consist}
\begin{tabular}{lcc}
\toprule
Method & UF1 Self-Consistency & UAS (UD) Self-Consistency \\
\midrule
StructFormer\(_{reproduction}\) & 70.32 (4.97) & 53.77 (12.23)   \\
\bottomrule
\end{tabular}

\end{table}

%%%%%%%%%%%%%%%%%%%%%%%%%%%%%%%%%%%%%%%%%%%%%%%%%%%%%%%%%%%%%%%%%%%%%%%%%%%%%%%%%%%

\section{Discussion}
\label{sec:sf_disc}

At the conclusion of this chapter, we present several observations and conclusions derived from our extensive study:

\begin{itemize}
    \item Our replication efforts revealed trends consistent with the original paper, yet the results did not precisely align. Potential factors for these discrepancies include environmental differences and the model's instability. Additionally, it remains unclear whether the original paper's results represent a single run or an aggregate measure (e.g., average, median, or best) from multiple runs.
    \item The unsupervised dependency parsing results cannot be directly compared with those of the original paper due to our use of the \(Dep_{auto}\) dataset as reference trees during evaluation, whereas the original study employed the Stanford Dependency framework.
    % \item The Unlabeled Attachment Score (UAS) does not account for the validity of an induced tree, potentially rating an invalid tree as positively matching the reference tree.
    \item Tree annotations are not fully ground truth targets, as they are based on specific linguistic theories and procedures that reflect linguists' hypotheses. Hence, the unsupervised parsing results should be carefully interpreted; a low alignment score (UF1 or UAS) should not imply the failure of the induction process but rather imply the misalignment of the induced structures against the target annotation framework.

    \item Automatic-annotated trees can be wrong, as was observed in Figure~\ref{fig:ex_dep_pre}. This fact raises the need to evaluate and analyze the induced trees against other annotation frameworks from both automatic parsers and human annotations. Furthermore, the induced trees can be analyzed independently to potentially deduce a consistent new grammar framework represented by the model. 
    
    \item StructFormer's induced constituency trees are limited to only binary trees. This adds a substantial limitation when evaluating the induced trees against non-binary trees.

    \item A significant discrepancy in parameter count between the vanilla Transformer and StructFormer, primarily due to the parser network module in StructFormer, suggests direct comparisons may not be equitable. Future experiments should neutralize the parameter count effect when comparing these models.
    
    \item Notable advantages of StructFormer include:
    \begin{enumerate}
        \item Its fully unsupervised nature, devoid of the "cheating" sometimes observed in other methods, as discussed in Chapter~\ref{ch:related_work}.
        \item The innovative mechanism of representing syntactic trees through numerical scalar values (syntactic distance and height).
        \item Its capacity to induce constituency and dependency trees simultaneously facilitates knowledge transfer between these syntactic tree types.
    \end{enumerate}
    \item Identified limitations of StructFormer encompass:
    \begin{enumerate}
        \item The use of a limited pretraining dataset, both in size and domain and a basic word-based tokenization algorithm. Despite no explicit limitations to utilizing larger and more diverse training corpora or sophisticated tokenization algorithms, the model operates fully unsupervised and does not require annotated data for training.
        \item The parser network's convolution net's limited kernel size is probably restricting the model's ability to perceive long-term dependencies. More experiments are needed to quantify the effect of kernel size on induced trees.
        \item The induced dependency trees by StructFormer are not guaranteed to be valid, a limitation that appears addressable in future iterations.
        % \item The model's inductive bias does not strongly favor any specific annotation framework, yet it is evaluated against such frameworks.
    \end{enumerate}
\end{itemize}
\chapter{Extending StructFormer: Experimental Modifications}
\label{ch:sf_dev}

Following the introduction of the StructFormer model in the preceding chapter, this chapter explores various modifications that address the limitations of the original implementation. Section~\ref{sec:in_parser} investigates an architectural modification inspired by insights gleaned from previous probing studies. Section~\ref{sec:subword} assesses the implementation of subword-level syntactic structure induction within StructFormer. Section~\ref{sec:babylm} explores a practical application of the model through participation in the BabyLM shared task. Eventually, Section~\ref{sec:dev_res} provides a comprehensive discussion of the developments and experimental insights detailed in this chapter.

\section{In-between parser}
\label{sec:in_parser}

In exploring whether transformer models capture linguistic knowledge and the feasibility of transferring this knowledge to solve downstream tasks in NLP, \cite{liu-etal-2019-linguistic} demonstrate that the middle layers of transformer models are particularly adept at transfer learning. This finding indicates the importance of the middle layers in a transformer model in capturing general linguistic knowledge. Furthermore, studies of probing attention layers in transformer models for syntactic dependency relations, such as~\cite{vig-belinkov-2019-analyzing}, find that attention most strongly aligns with dependency relations in the middle layers. This finding is corroborated by~\cite{arps-etal-2022-probing}, indicating that syntactic information is represented in these middle layers.

\begin{figure}[!htp]
    \centering
    \caption[The Architecture of the In-between Parser Modification]{The model architecture of StructFormer with the In-between Parser Modification.}
    \label{fig:structformer_inparser}
    \includegraphics[width=0.52\linewidth]{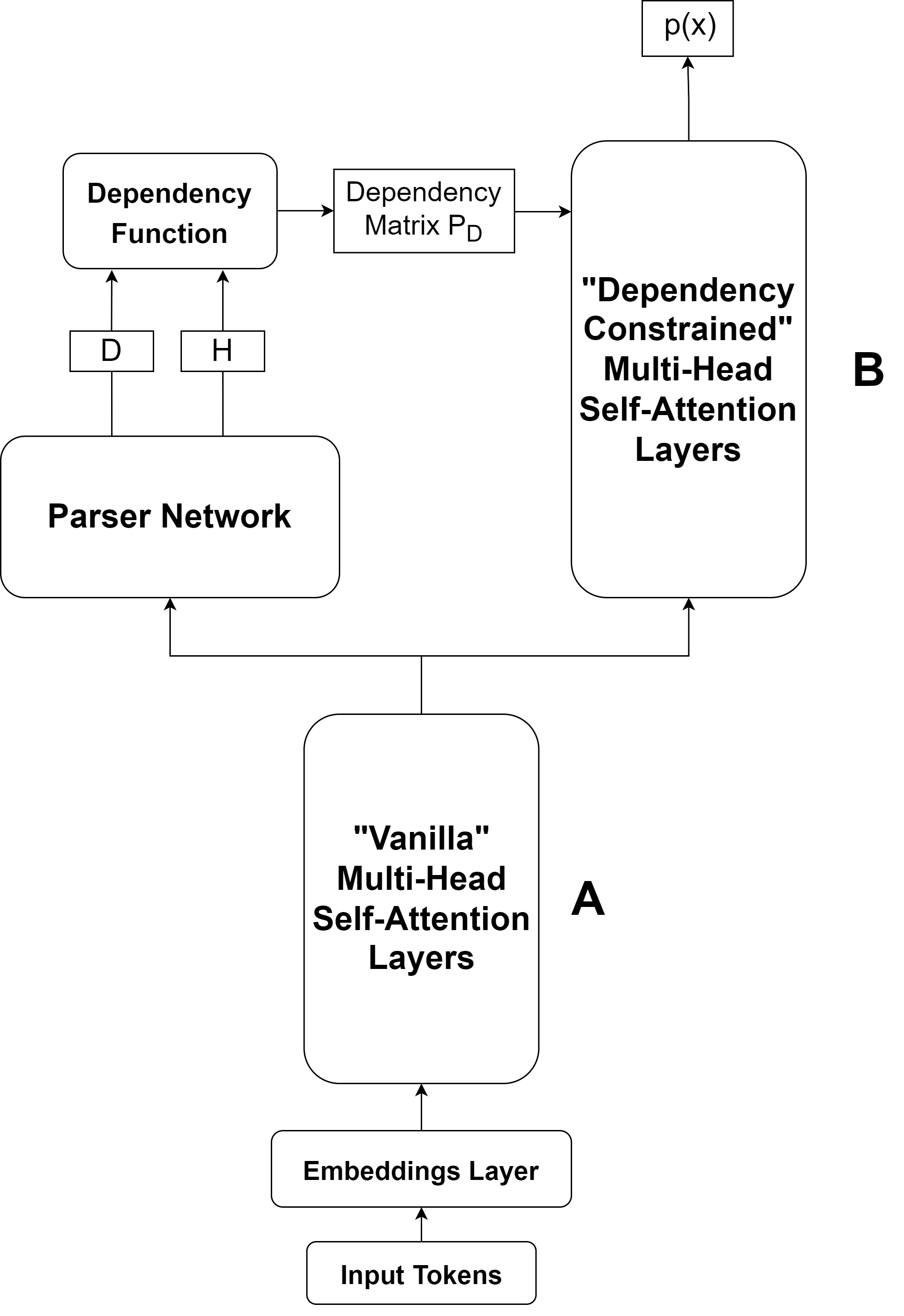}
\end{figure}

The current iteration of StructFormer introduces syntactic information across all layers of a transformer model. This design does not sufficiently take into account the findings of the connections between the middle attention layers and syntactic knowledge. A proposed architectural modification involves integrating the syntactic information after several layers of multi-head self-attention, enabling the parser network to process contextualized rather than static embeddings. As depicted in Figure~\ref{fig:structformer_inparser}, this modification positions the parser network amid the transformer blocks. The initial set of transformer blocks \textit{A} processes embeddings before being fed into the parser network. The parser network then produces the dependency matrix. The dependency matrix, in addition to the output embeddings from \textit{A}, is then fed into a subsequent series of transformer blocks \textit{B} that follow the design of dependency-constrained multi-head self-attention as discussed in Section~\ref{sec:comps}. This architectural change aims to leverage the nuanced representation of syntactic information in the middle layers to examine the applicability of the mentioned findings on StructFormer.

\begin{figure}[!htp]
    \centering
    \caption[Perplexity Results for the In-between Parser Modification]{Masked words perplexity results on the PTB test set for 40 trained models, five runs at each selected position of the in-between parser network. Position 0 means that the parser network is positioned right after the embeddings layer (as in the original StructFormer), and positions 1 to 7 means that the parser network is positioned after \(m\) transformer blocks. The solid line indicates the mean value of 5 runs per configuration, and the highlighted area indicates the 95\% confidence interval}
    \label{fig:inparser_ppl}
    \includegraphics[width=0.72\linewidth]{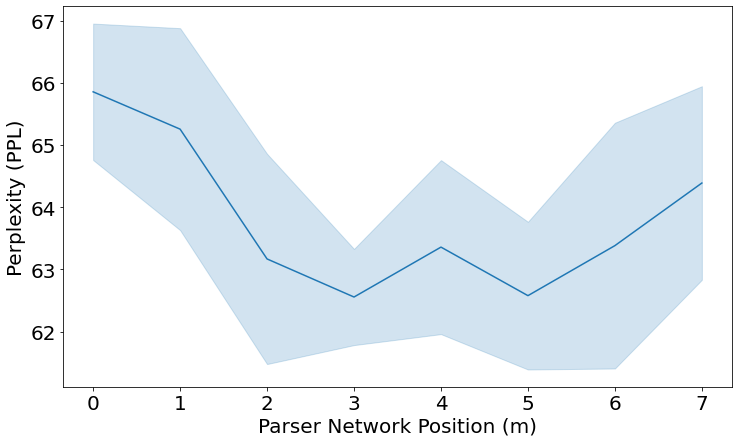}
    
\end{figure}

\begin{figure}[!htp]
    \centering
    \caption[Constituency Parsing Results for the In-between Parser Modification]{Unsupervised Constituency Parsing results on the PTB test set for 40 trained models. The plots from top to bottom show the unlabelled precision, recall, and F1 of constituency parsing.}
    \label{fig:inparser_parsing}
    \includegraphics[width=0.69\linewidth]{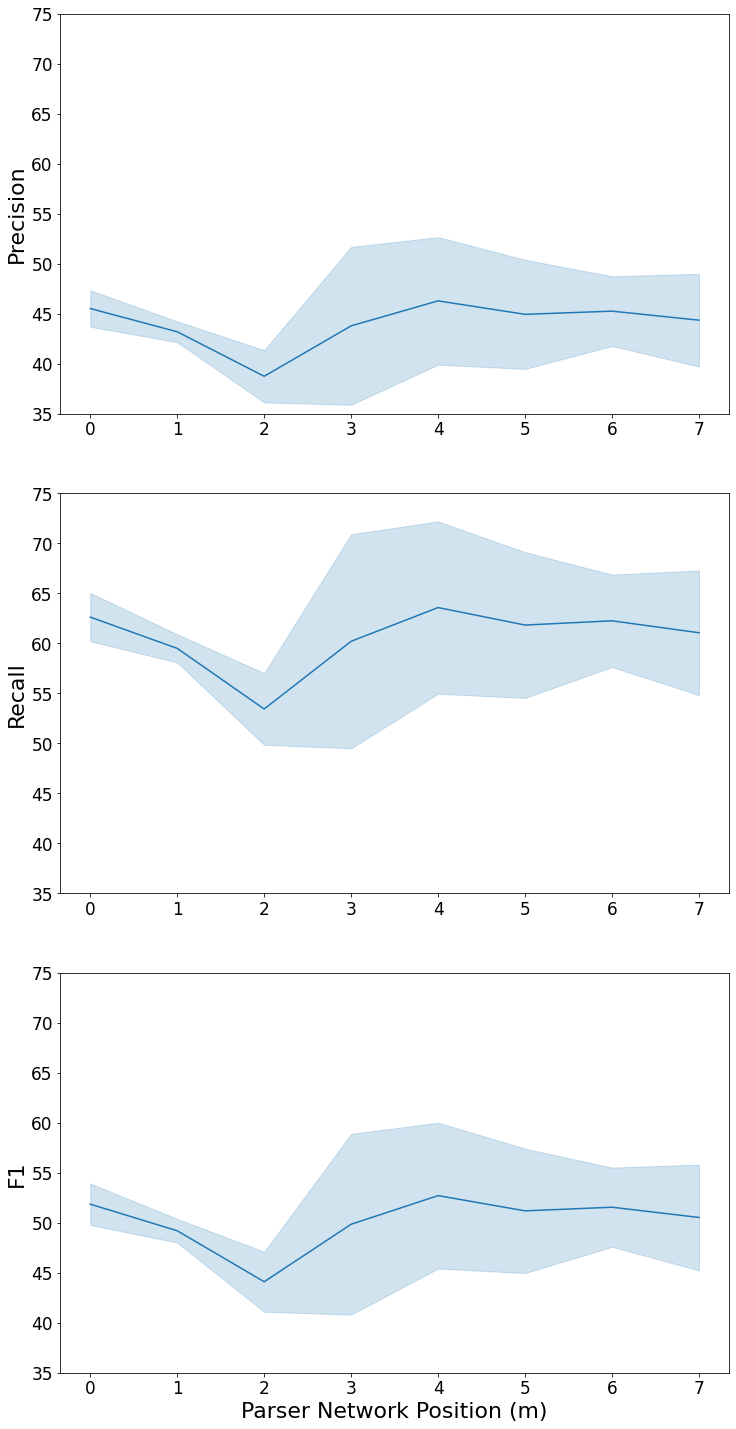}
\end{figure}

\begin{figure}
    \centering
    \caption[Dependency Parsing Results for the In-between Parser Modification]{Unsupervised Dependency Parsing results on the PTB test set for 40 trained models. The plots from top to bottom show the unlabelled precision, recall, and F1 of constituency parsing.}
    \label{fig:inparser_uas}
    \includegraphics[width=0.72\linewidth]{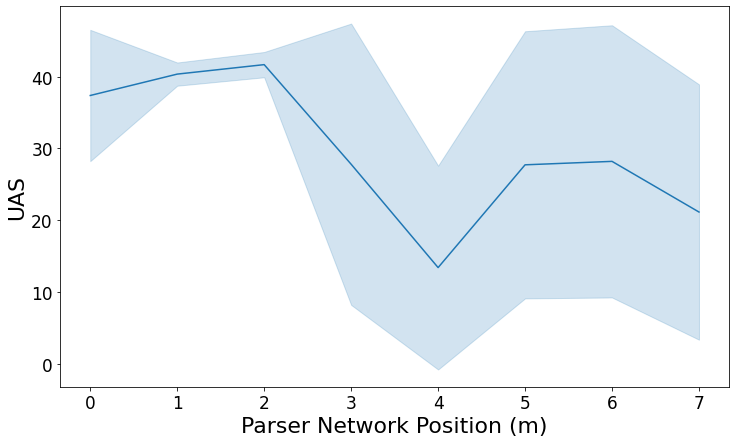}
\end{figure}

To assess the impact of this architectural modification, we maintain the same experimental settings and hyperparameters as outlined in Section~\ref{sec:implem}. The total number of parameters in the modified models is consistent with the original StructFormer, adhering to the same number of attention layers \( L = 8 \). However, these layers are divided into two distinct groups: set \textit{A} and set \textit{B} as shown in Figure~\ref{fig:structformer_inparser}. Our experimentation includes varying the placement position \(m\) of the parser network within the attention layers. This ranges from positioning it immediately after a single attention layer to placing it after seven layers, allowing for an evaluation of the parser network's effectiveness at different stages of attention processing. The outcomes of this experiment are depicted in Figures~\ref{fig:inparser_ppl}, \ref{fig:inparser_parsing}, and \ref{fig:inparser_uas}. To ensure robustness and mitigate the influence of randomness, each configuration was tested across five runs, each initiated with a different random seed. Adapted MLM perplexity~\ref{eq:ppl_mlm} is employed as the evaluation metric for language modeling. Constituency parsing is assessed using unlabelled precision, recall, and F1 scores. While dependency parsing is evaluated based on UAS.

The perplexity improves for models incorporating the parser network within the middle layers (specifically layers 2 to 5), achieving optimal results at \(m=3\). A marginal improvement is observed at layer 4 in unlabelled constituency parsing evaluation. While dependency parsing exhibits a slight enhancement at the \(m=2\) setting. However, models with the parser network placed at \(m>1\) display inconsistent results across the five pretraining iterations, highlighted by the large variance in the confidence intervals in the plots. The decrease in perplexity for models with the parser network in the middle layers aligns with the findings that suggest that the mid-layer embeddings in transformer models capture more nuanced linguistic features. The slight improvements observed in unlabelled constituency parsing at layer 4 and in dependency parsing at layer 2 suggest that different aspects of syntactic parsing may benefit from different layer placements. These inconsistent results could be due to the varying nature of syntactic information encoded at each attention layer in the transformer model. The inconsistent results for models with the parser network positioned at \(m>1\) across iterations suggest a possible trade-off between incorporating syntactic information and maintaining structure induction stability. This inconsistency might be due to the complexity added by the parser network affecting the transformer's ability to generalize across different runs.

%%%%%%%%%%%%%%%%%%%%%%%%%%%%%%%%%%%%%%%%%%%%%%%%%%%%%%%%%%%%%%%%%%%%%%%%%%%%%%%%%%%
\section{Subword Tokenization}
\label{sec:subword}

The StructFormer model originally uses whole word tokenization, which leads to \textit{<unk>} tokens for rare words that appear in input sentences as shown in~\ref{sec:example}. Subword tokenization could mitigate this by building word embeddings from smaller subword units, improving the handling of out-of-vocabulary words, and allowing for efficient pretraining on large corpora~\cite{subword_history}. However, to the extent of our knowledge, there has been limited exploration for applying subword tokenization in unsupervised linguistic structure induction. Existing benchmarks, baseline methods, and evaluation procedures evolve around word-token analysis. Furthermore, the theoretical foundations of syntactic parsing are intrinsically linked to word tokens, justifying the field's alignment with this paradigm. We posit that integrating subword tokenization into unsupervised parsing could yield considerable benefits. Key among these is the ability to pretrain on relatively large datasets without excessively expanding the vocabulary size, enlarging the model size. Also, to enable extending StructFormer on languages that have rich morphologies such as Arabic, German, or Turkish, where applying whole word tokenization is inefficient~\cite{10.1145/3578707}. Moreover, leveraging the advantages of subword over word tokenization in language modeling could potentially enhance a model's capacity for capturing linguistic knowledge.

As an initial step in this direction, we are experimenting with pretraining StructFormer with a subword tokenizer. We also propose a method to compare induced trees containing subword tokens against traditionally annotated trees with only word tokens. Our focus extends to assessing the parser network's ability to group subword units of a word into a single constituent. Ultimately, we aim to develop a feasible pipeline for pretraining and evaluating StructFormer on any textual corpus and potentially languages other than English.

\textbf{Subword Tokenizer:} In our study, we utilize the Byte-Pair Encoding (BPE) algorithm~\cite{sennrich-etal-2016-neural,gage94}, a widely recognized subword tokenization algorithm that strikes a balance between character- and word-level representations. BPE operates on subword units derived from statistical analysis of the training corpus. We train a BPE-based tokenizer\footnote{\url{https://huggingface.co/docs/transformers/model_doc/roberta\#transformers.RobertaTokenizer}} on the PTB training dataset. Resulting in a subword tokenizer with a vocabulary comprising 8k units. This size is modest yet sufficient to encode any input text from the corpus without resorting to the use of \textit{<unk>} tokens.

\textbf{MLM Pretraining:} Equipped with this subword tokenizer, we pretrain the StructFormer and the vanilla traditional Transformer using the default configuration and hyperparameter settings as outlined in Section~\ref{sec:implem}, with the sole modification being the tokenizer employed. The adapted MLM perplexity~\ref{eq:ppl_mlm} is measured on the PTB test set and presented in Table~\ref{tab:subword_eval}. These perplexity values cannot be directly compared with the previous experiments due to the differences in tokenization and vocabulary employed.

\textbf{Parsing Evaluation:} Following the approach outlined in Section~\ref{sec:example}, we apply the same procedure for inducing constituency and dependency trees from the pretrained StructFormer model, albeit with some modifications due to the use of a subword tokenizer. The induced trees in this experiment, structured over subword units, cannot be directly compared with the reference trees that consist solely of whole word units. To address this, we develop a method to convert the reference trees into subword-based trees, enabling meaningful comparison with the induced trees. For converting reference \textit{Constituency Trees} into subword-based trees, the aim is to have the exact same tokens in each pair of reference and induced trees. Two key steps are undertaken:

\begin{enumerate}
    \item \textbf{Merge The Pre-Split Words in PTB Corpus:} The PTB annotations follow specific tokenization standards that result in split words in the corpus and the annotated trees. To align with our subword tokenizer, it was necessary to revert these splits. After a manual analysis of the PTB corpus, these splits were spotted in the cases of contractions (e.g., \textit{it's}, \textit{isn't}, \textit{I'm}), possessive forms (e.g., \textit{Dow's}, \textit{Friday's}), and symbols attached to numbers (e.g., \textit{5\%}, \textit{15\$}). By merging these elements into whole words in the annotated tree, we restored the integrity of the original text.

    \item \textbf{Split Words into Subword Constituents:} The leaf nodes in the annotated trees are split into subwords as per our pretrained tokenizer. In our subsequent analysis, these splits are then incorporated as additional constituents within the annotated tree, termed \textit{subword constituents}.
\end{enumerate}

In contrast to the preprocessing steps in the original implementation discussed in Section~\ref{sec:example}, we do not apply lowercasing on input sentences and reference trees in this subword tokens implementation. However, we keep the removal of punctuation marks for a concrete parsing evaluation. We also keep replacing the constituent labels with a placeholder (X), except for constituents that emerge from splitting words into subwords according to our pretrained tokenizer. These constituents are labeled as "Subword Constituent" \texttt{SWC} in the processed trees.

\begin{figure}
    \centering
    \caption[Subword Preprocessing Steps Example]{Preprocessed Reference Constituency Tree for sentence~\ref{eq:sen1_pre} performing the subword preprocessing steps on Tree~\ref{fig:ex_const_pre}. Each \textit{X} represents an unlabelled constituent, and \textit{SWC} represents a subword constituent.}
    \label{fig:subword_proc}
    \includegraphics[width=0.75\linewidth]{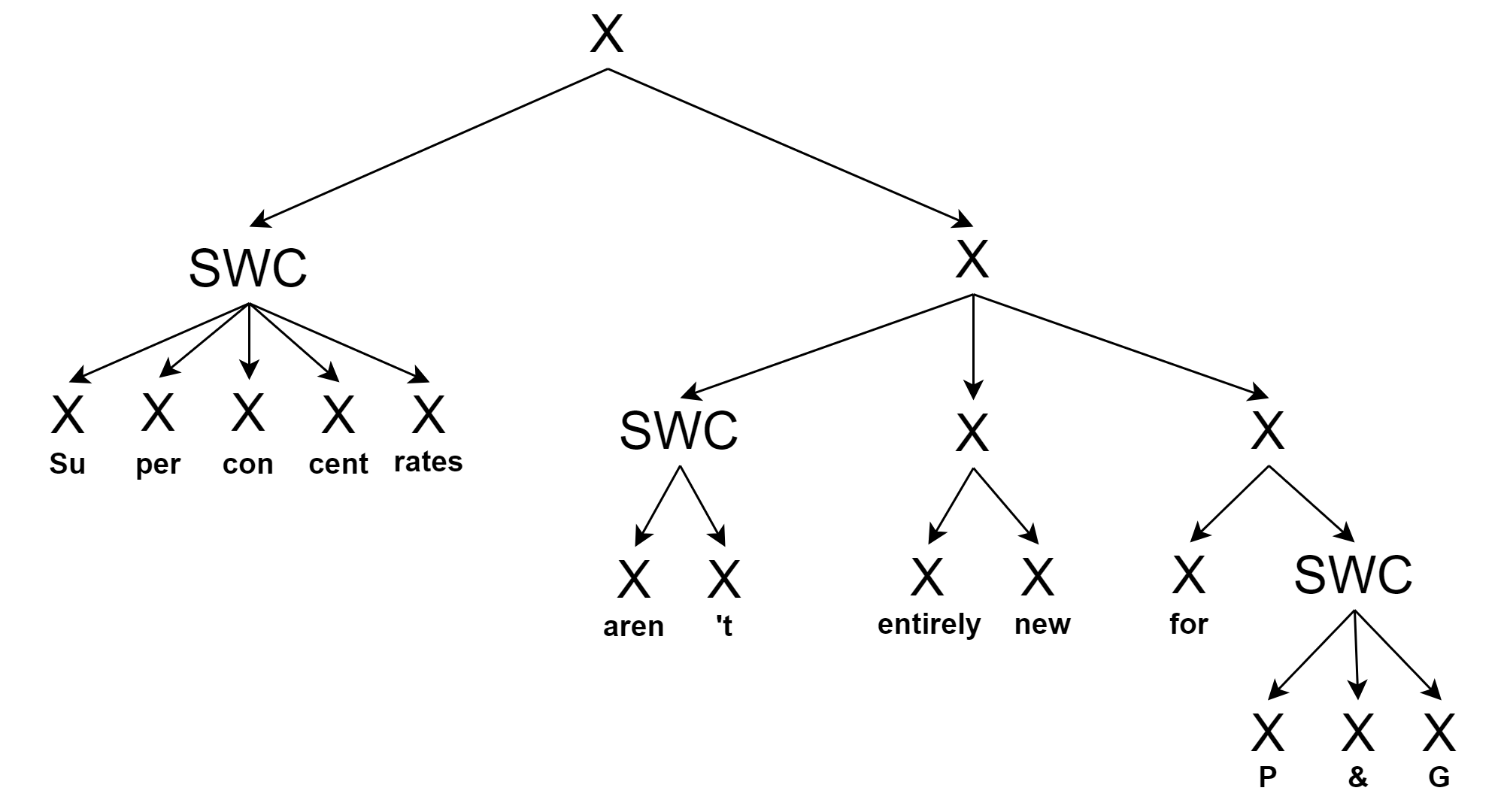}
\end{figure}

\begin{figure}
    \centering
    \caption[Induced Subword Constituency Tree]{Induced Subword Constituency Tree for sentence~\ref{eq:sen1_pre}. Each \textit{X} represents an unlabelled constituent, and \textit{SWC} represents a correctly recognized subword constituent.}
    \label{fig:induced_subword}
    \includegraphics[width=0.65\linewidth]{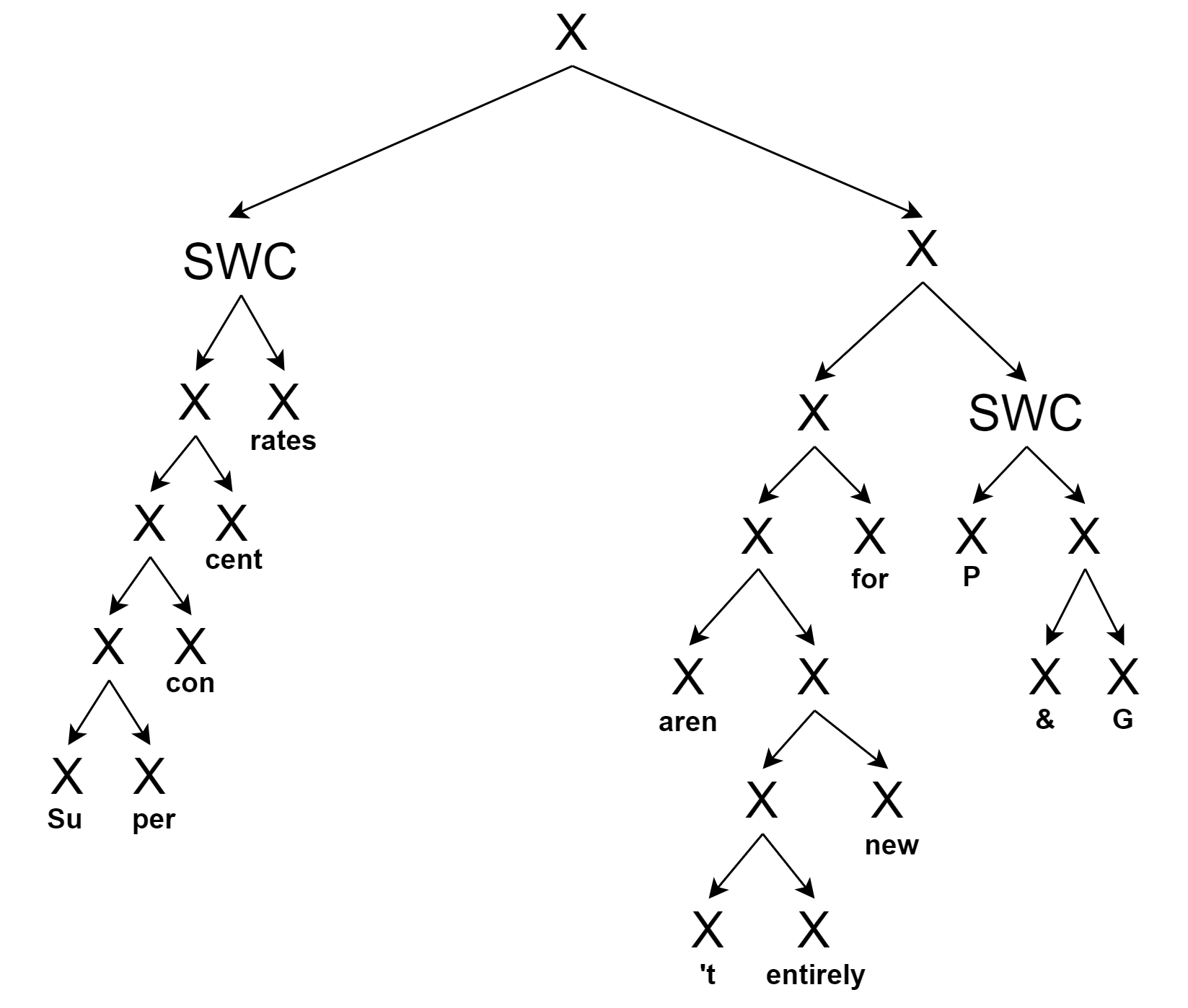}
\end{figure}

An example of the preprocessing steps discussed above is presented in Figure~\ref{fig:subword_proc}, which is the reference tree for the sentence~\ref{eq:sen1_pre} after applying the subword preprocessing steps. Contrasting the original (whole words) reference tree~\ref{fig:ex_const_pre} in the PTB dataset and this preprocessed (subword) reference tree highlights the preprocessing steps. The rare word \textit{Superconcentrates} is no longer replaced with an <unk> token. Alternatively, it is split into the subword units (\textit{Su}, \textit{per}, \textit{con}, \textit{cent}, \textit{rates}) according to our pretrained subword tokenizer. An internal node with the label \texttt{SWC} emerges, replacing the leaf node \textit{Superconcentrates} in the original reference tree. Similarly, the \textit{P\&G} node is also split into subwords constituents. The nodes \textit{are} and \textit{n't} in Figure~\ref{fig:ex_const_pre} are firstly merged into one node, resulting in only three nodes under the parent \textit{VP} node instead of the four nodes in the original tree. Then, this merged node is split according to the subword tokenizer into two subwords and is replaced by a \texttt{SWC} internal node. The induced tree for this sentence from a pretrained model variant that utilizes our subword tokenization procedures is illustrated in Figure~\ref{fig:induced_subword}. In this induced tree, three out of the six constituents in the preprocessed reference tree are correctly predicted, where two of the subword constituents are correctly recognized.

We encountered a significant challenge in converting reference \textit{Dependency Trees} to a subword-based format. Merging pre-split tokens within a dependency tree is a complex problem because these split units do not always have direct parent-child relationships in the reference trees. Attempting to merge such distantly related nodes can result in fragmented subtrees or necessitate linguistic decisions about re-establishing connections to form a valid reference dependency tree. This complexity is further amplified when one of the splits, which requires merging with other nodes, is a descendant of the root node. In such cases, a new node descending from the root is required to reconnect the tree. Given these complexities and the need for decisions grounded in linguistic knowledge that goes beyond the scope of this thesis, we decided not to evaluate induced dependency trees against reference trees in our current work, leaving this as an area for future research.

After preprocessing the reference annotated constituency trees, as explained, we were able to evaluate the induced trees. The results, including precision, recall, and F1 scores, are reported in Table~\ref{tab:subword_eval}. However, these metrics are not directly comparable to those in~\ref{sec:eval} due to the different constituent structures in the reference trees. Additionally, we assess the parser network's proficiency in grouping subword units into a single constituent, as assumed in our preprocessed reference trees. We measure this aspect through the recall of subword constituents \texttt{SWC} in the induced trees. The percentage of subword constituents out of all the constituents in the reference trees is only 20\%.

\begin{table}[!htp]
\centering
\caption[Subword-Level StructFormer Results]{Subword-Level evaluation results of StructFormer and Transformer on PTB test split. The values show the mean and (standard deviation) of 5 different runs with exactly the same configuration but with a different random generator seed each time. The percentage of subword constituents out of all the constituents in the reference trees is 20\%. Adapted MLM perplexity is measured following~\ref{eq:ppl_mlm}.}
\label{tab:subword_eval}
\begin{tabular}{llcc}
\toprule
Aspect & Metric & StructFormer & Transformer  \\
\midrule
Masked Language Modeling & MLM Perplexity & 46.19 (1.3) & 51.16 (1.62)  \\
Constituency Parsing & Precision & 45.18 (1.17) \\
Constituency Parsing & Recall & 63.6 (1.78)  \\
Constituency Parsing & F1 & 52.41 (1.39) \\
Subword Constituents Recovery & Recall & 87.79 (5.37)  \\
\bottomrule
\end{tabular}
\end{table}

Our utilization of a subword-level approach for unsupervised structure induction can not be compared against other studies, as this methodology is uncommon within this domain. Despite the lack of similar experiments for direct comparison, our results verify the StructFormer's outperformance over the traditional Transformer model in masked language modeling, as shown by its lower perplexity scores. Additionally, the F1 constituency score (52.41) is deemed acceptable while not directly compared against other baselines. This outcome underscores the viability of subword tokenization in unsupervised structure induction and validates our proposed method for evaluating subword-based trees against word-based reference trees. Moreover, the relatively high recall of subword constituents highlights StructFormer's effective parser mechanism in capturing contextual relationships among input tokens, further attesting to its efficacy in parsing and structure induction tasks.

%%%%%%%%%%%%%%%%%%%%%%%%%%%%%%%%%%%%%%%%%%%%%%%%%%%%%%%%%%%%%%%%%%%%%%%%%%%%%%%%%%%

\section{BabyLM Challenge}
\label{sec:babylm}

In advancing beyond the limitations of the problem of unsupervised linguistic structure induction, a significant step is to pretrain the models on a more expansive dataset, leveraging the subword tokenizer we previously introduced. Our selection of a larger dataset led us to the BabyLM shared task, a challenge aimed at data-efficient language modeling for English. This task challenges participants to pretrain a language model (LM) from scratch on raw text using an amount of data akin to that available to a child. The BabyLM task offers a dataset for pretraining and a comprehensive evaluation pipeline for benchmarking our pretrained model variants against other innovative architectures proposed by other participants in the challenge, making it an ideal setting for our experiment. The challenge of the shared task is to improve language modeling using a limited amount of raw text, which also aligns with the core benefit of StructFormer. 

The primary goals of our experiment, in the context of the BabyLM shared task, are:

\begin{enumerate}
    \item To assess the scalability of StructFormer with larger datasets and subword tokenization featuring a different level of language complexity in the training corpus.
    
    \item To evaluate StructFormer across a broad spectrum of tasks designed to measure model performance in applied NLP, as well as in cognitive science and linguistics. This aspect is further elaborated in Chapter~\ref{ch:ev}.
    
    \item To investigate the performance of linguistic structure induction against generated reference parsed trees from a state-of-the-art parser.
\end{enumerate}

Our findings and methodology are detailed in a research paper~\cite{omar-2023} published at the proceedings of CoNLL 2023. The third objective above was not part of the original experiment documented in~\cite{omar-2023} but is included in this thesis as an additional exploration. More details about the BabyLM dataset can be found in Appendix~\ref{apx:babylm}.

\textbf{Large-scale Pretraining:} We pretrain two variants of StructFormer that reflect the previously discussed modifications; a subword-based StructFormer with the parser network placed before all the transformer blocks \(SF_{m=0}\) (reflecting the original implementation of StructFormer), a subword-based StructFormer with the parser network placed after \(4\) transformer blocks \(SF_{m=4}\) (reflecting our in-between parser modification). Additionally, we pretrain a subword-based vanilla Transformer \(TF\) to act as a baseline. All model variants are pretrained on the small version of the BabyLM training dataset, which contains approximately 10 million tokens in its training split. For this experiment, a subword BPE-based tokenizer was pretrained on the training data, resulting in a vocabulary size of 16K units. We adhere to the same configuration and hyperparameter settings delineated in Section~\ref{sec:implem}. The adapted MLM perplexity~\ref{eq:ppl_mlm} results on the BabyLM test set are detailed in Table~\ref{tab:babylm_eval}.

\textbf{Structure Induction:} The BabyLM dataset comprises solely raw text sentences devoid of any annotated syntactic trees. To evaluate the induced trees from \(SF_{m=0}\) and \(SF_{m=4}\), we utilize the Berkeley Neural Parser (BNP), a supervised parser proficient in Constituency Parsing with a Self-Attentive Encoder~\cite{kitaev-klein-2018-constituency}, which demonstrates a high F1 score of 95.13 on the PTB dataset. BNP is employed in our experiment to generate reference trees for the test split of the BabyLM dataset. BNP offers two operational modes: it can either build constituency trees based on a provided list of tokens or perform tokenization internally on a raw sentence input, with the latter method recommended by the authors for optimal results. However, we opted for the former approach, ensuring that the reference trees contain the exact same tokens as our induced trees. This choice was crucial because our tokenizer might generate tokens that are different from those of the BNP's built-in tokenizer. Additionally, it is noteworthy that the BNP parser may encounter performance variations due to domain differences between the dataset it was trained on and the BabyLM dataset. It is essential to consider these points when analyzing the performance of StructFormer to induce syntactic structures that match BNP's generated reference trees. We present the unlabeled precision, recall, and F1 scores for the induced trees of the sentences in the BabyLM test set by \(SF_{m=0}\) and \(SF_{m=4}\) in Table~\ref{tab:babylm_eval}.

\begin{table}[!htp]
\centering
\caption[BabyLM Pretraining Results]{Comparison of StructFormer pretraining with the parser network in its original configuration (SF\(_{m=0}\)), an intermediary parser variant (SF\(_{m=4}\)), and a standard transformer (TF) on the BabyLM dataset, showing evaluation results on the test split. 23.35\% of all constituents in the reference trees are subword constituents.}
\label{tab:babylm_eval}
\begin{tabular}{llccc}
\toprule
Aspect & Metric & TF & SF\(_{m=0}\) & SF\(_{m=4}\)  \\
\midrule
Masked Language Modeling & MLM Perplexity & 27.02 & 28.90 & 26.64 \\
Constituency Parsing & Precision &          &  35.85 & 27.15   \\
Constituency Parsing & Recall &             &  70.94 & 55.38 \\
Constituency Parsing & F1 &                 &  42.93 &  32.12 \\
Subword Constituents Recovery & Recall &    &  76.88 &  68.08  \\
\bottomrule
\end{tabular}
\end{table}

The results reported in Table~\ref{tab:babylm_eval} underscore StructFormer's robustness in masked language modeling when pretrained on datasets larger and distinct in domain from the traditional PTB dataset. Moreover, StructFormer achieves acceptable results in Constituency Parsing compared to reference trees from automatic parsers. For the first time in the experiments in this thesis, the vanilla transformer achieves a lower perplexity than StructFormer. Positioning the parser network after 4 attention layers confirms the results from Section~\ref{sec:in_parser} that in-between parser variants reduced the measured perplexity. However, the in-between parser variant fails to improve StructFormer's structure induction capabilities highlighted by the lower Constituency Parsing results in Table~\ref{tab:babylm_eval} for \(SF_{m=4}\) in contrast to \(SF_{m=0}\).

%%%%%%%%%%%%%%%%%%%%%%%%%%%%%%%%%%%%%%%%%%%%%%%%%%%%%%%%%%%%%%%%%%%%%%%%%%%%%%%%%%%

\section{Discussion}
\label{sec:dev_res}

In this chapter, we explore three main ideas to enhance the original StructFormer framework, aiming not only to improve upon its initial implementation but also to broaden the scope of research in linguistic structure induction from language models. Our investigation into optimizing the StructFormer architecture by adjusting the parser network's placement position \(m\) within the transformer blocks reveals that positioning the parser in the middle layers (specifically at \(2 \leq m \leq 5\)) slightly enhances model efficiency, as shown by reduced perplexity and slight improvements in syntactic parsing tasks at some positions \(m\). These findings align with existing literature, suggesting that mid-layer embeddings are pivotal for capturing complex linguistic features. However, the variability in results across different pretraining iterations points to a nuanced trade-off between integrating syntactic information for improved accuracy and maintaining consistent induced trees. This exploration underscores the delicate balance required in transformer model architecture adjustments to optimize linguistic structure induction, highlighting the potential for future research to refine these approaches further.

Our exploration of the subword-level approach for unsupervised structure induction presents unique findings that contribute positively to the field, especially given the novelty of this methodology within the domain. The StructFormer's outperformance in masked language modeling, demonstrated by its lower perplexity scores compared to the traditional Transformer models, underscores the success of implementing subword tokenization in StructFormer. Furthermore, the acceptable UF1 constituency score, despite the absence of direct baseline comparisons, reinforces the potential of our approach. This achievement highlights the feasibility of subword tokenization for unsupervised structure induction and affirms the validity of our method for evaluating subword-based trees in contrast to word-based reference trees. The notably high recall of subword constituents further shows StructFormer's proficiency in discerning contextual relationships between input tokens, reinforcing its parsing and structure induction capabilities. While specific to our study, these insights lay the groundwork for future research to go deeper into the implications and applications of subword tokenization in linguistic structure induction.

The subword-level approach enables large-scale pretraining of StructFormer, demonstrated by pretraining the model on a larger dataset and from different domains than the PTB dataset traditionally used in such tasks. Additionally, StructFormer's performance in Constituency Parsing demonstrates acceptable outcomes when compared with reference trees generated by automatic parsers. This result might suggest that StructFormer's approach fulfills a partial alignment with various frameworks, hinting at the presence of a more general inductive bias rather than one favoring a particular framework. This suggestion hints at potential challenges in achieving high F1 scores due to the lack of a clear inductive bias towards a specific annotation framework. This realization opens up alternative avenues for interpreting the trees StructFormer induced. Rather than focusing solely on the degree to which these trees align with existing reference trees, future research can explore the linguistic patterns and theories these induced trees suggest.
\chapter{Linguistic Evaluation}
\label{ch:ev}

To further our analysis of integrating syntactic inductive bias into transformer-based language models, we assess the StructFormer pretrained model variants discussed in Chapter~\ref{ch:sf_dev} across various language tasks and compare their performance against other model architectures. 

The BabyLM challenge~\cite{warstadt-etal-2023-findings} provides an extensive evaluation pipeline, encompassing 39 diverse tasks designed to gauge a model's proficiency in capturing the English language's latent syntactic and semantic structures. Therefore, we have opted to leverage the challenge's evaluation setup for our language evaluation experiment. We published the experiments and results of our participation in the BabyLM challenge in~\cite{omar-2023}. We assess three variants of the discussed models in this thesis that are pretrained using the default configuration settings detailed in Section~\ref{sec:implem}. The three variants are:
\begin{enumerate}
    \item Vanilla Transformer (\(TF\)); as introduced in~\cite{vaswani17}.
    \item StructFormer (\(SF_{m=0}\)); original StructFormer with the parser positioned before any transformer blocks.
    \item StructFormer (\(SF_{m=4}\)); an in-between parser variant with the parser positioned after 4 transformer blocks.
\end{enumerate}

In the reported results, we also include the scores of 118 submitted model architectures to the \textit{small-strict} track in the BabyLM Challenge 2023.

Section~\ref{sec:tests} describes the tasks and metrics employed in the evaluation process. Subsequently, the results of this experiment are detailed in Section~\ref{evp_results}, followed by a discussion of our observations and insights in Section~\ref{evp_disc}.

\section{Tasks \& Metrics}
\label{sec:tests}

According to the BabyLM challenge rules, 39 tasks and tests are selected to judge a pretrained model's linguistic performance. To assess the grammatical capabilities of a model, the Benchmark of Linguistic Minimal Pairs for English (\textit{BLiMP}) task suite~\cite{warstadt-etal-2020-blimp-benchmark} is utilized, complemented by the \textit{BLiMP-Supplement} tasks~\cite{warstadt-etal-2023-findings} which explore linguistic phenomena not covered by BLiMP. For evaluating the models' performance on more conventional downstream NLP tasks, a selection from the Stickier benchmark for General-purpose Language Understanding Evaluation (\textit{SuperGLUE})~\cite{super_glue} is chosen. Additionally, the Mixed Signals Generalization Set (\textit{MSGS})~\cite{warstadt-etal-2020-learning}, a text classification task designed to probe the inductive biases of LMs, is also applied. This comprehensive approach aims to evaluate models' linguistic capabilities and applicability to various NLP tasks.

\subsection{Tasks}

\textbf{BLiMP} is designed to test the alignment of language models with the structural aspects of the English language through 12 distinct tasks. Each task features a dataset containing a few thousand examples organized into minimal pairs of sentences. One sentence is linguistically correct in these pairs, while the other, differing only minimally, is incorrect. The evaluation criterion for a model hinges on its zero-shot prediction ability to assign a higher probability to the linguistically correct sentence within each pair, thereby demonstrating its comprehension of English morphology, syntax, and semantics. Based on~\cite{warstadt-etal-2020-blimp-benchmark}, we briefly illustrate the objective of each task, with a supporting example to exemplify the task for the reader in Table~\ref{tab:blimp_tasks}, the objectives descriptions are summarized from~\cite{warstadt-etal-2020-blimp-benchmark}, while examples and fields (the tested linguistic field) are extracted from the BLiMP published dataset\footnote{\url{https://huggingface.co/datasets/nyu-mll/blimp}}.

\newpage

\begin{longtable}{|p{2cm}|p{4.5cm}|p{2cm}|p{5cm}|}
\caption[BLiMP Tasks Catalogue]{List of the BLiMP tasks included in the BabyLM evaluation pipeline.}\label{tab:blimp_tasks} \\
\hline
\textbf{Name} & \textbf{Objective} & \textbf{Field} & \textbf{Example} \\ \hline
\endfirsthead
\multicolumn{4}{c}%
{{Table \thetable\ continued from previous page}} \\
\hline
\textbf{Name} & \textbf{Objective} & \textbf{Field} & \textbf{Example} \\ \hline
\endhead
\hline \multicolumn{4}{r}{{Continued on next page}} \\ 
\endfoot
\hline 
\endlastfoot

Anaphor Agreement & The requirement that reflexive pronouns like himself (a.k.a. anaphora) agree with their antecedents in person, number, gender, and animacy. & Morphology & \textbf{Correct}: "Some bank likes \textit{itself}."\footnote{This example was selected randomly, and we noticed that the labeled correct sentence, in this case, is semantically incorrect. A bank cannot like. Despite that, we keep the selected sentence to raise questions about the correctness of some samples in the BLiMP dataset.} 

\textbf{Incorrect}: "Some bank likes \textit{himself}." \\ \hline

Argument Structure & The ability of different verbs to appear with various types of arguments. & Syntax\footnote{The linguistic field label is extracted from the original BLiMP dataset. However, we doubt the correctness of some of these labels, such as Argument Structure, we believe it fits Semantics more than Syntax.} & \textbf{Correct}: "Amanda was respected by some \textit{waitresses}." 

\textbf{Incorrect}: "Amanda was respected by some \textit{picture}." \\ \hline

Binding & The structural relationship between a pronoun and its antecedent. & Syntax \& Semantics & \textbf{Correct}: "It's herself that \textit{Sharon talked about}." 

\textbf{Incorrect}: "It's herself that \textit{talked about Sharon}." \\ \hline

Control 

Raising & Differences between types of predicates that embed an infinitival VP. This includes control, raising, and tough-movement predicates. & Syntax \& Semantics & \textbf{Correct}: "Frank isn't \textit{about} to cry." 

\textbf{Incorrect}: "Frank isn't \textit{enjoyable} to cry." \\ \hline

Determiner Noun 

Agreement & Number agreement between demonstrative determiners (e.g. this/these) and the associated noun. & Morphology & \textbf{Correct}: "Craig explored that grocery \textit{store}." 

\textbf{Incorrect}: "Craig explored that grocery \textit{stores}." \\ \hline

Ellipsis & The possibility of omitting expressions from a sentence. & Syntax & \textbf{Correct}: "Brad passed one big museum and Eva passed \textit{several}." 

\textbf{Incorrect}: "Brad passed one museum and Eva passed \textit{several big}." \\ \hline

Filler Gap & Dependencies from phrasal movement in structures like wh-questions. & Syntax & \textbf{Correct}: "Linda did notice \textit{that couch that Naomi hides}." 

\textbf{Incorrect}: "Linda did notice \textit{what Naomi hides that couch}." \\ \hline

Irregular Forms & Focuses on irregular morphology of English past participles. & Morphology & \textbf{Correct}: "The \textit{known} teacher isn't troubled." 

\textbf{Incorrect}: "The \textit{knew} teacher isn't troubled." \\ \hline

Island Effects & Restrictions on syntactic environments for a gap in a filler-gap dependency. & Syntax & \textbf{Correct}: "Who could Amelia leave \textit{while appreciating Paul}?" 

\textbf{Incorrect}: "Who could Amelia leave \textit{Paul while appreciating}?" \\ \hline

NPI 

Licensing & Constraints on the distribution of negative polarity items. & Semantics & \textbf{Correct}: "\textit{Had Suzanne} ever sat down." 

\textbf{Incorrect}: "\textit{Suzanne had} ever sat down." \\ \hline

Quantifiers & Constraints on the distribution of quantifiers. & Semantics & \textbf{Correct}: "There were \textit{many} sketches bothering Ruth." 

\textbf{Incorrect}: "There were \textit{all} sketches bothering Ruth." \\ \hline

Subject Verb Agreement & Subjects and present tense verbs must agree in number. & Morphology & \textbf{Correct}: "The grandfathers of Diana \textit{drink}." 

\textbf{Incorrect}: "The grandfathers of Diana \textit{drinks}." \\

\end{longtable}

\textbf{BLiMP-Supplement} task set expands the evaluation scope of the language models by introducing five test suites featuring BLiMP-style minimal pairs, which examine areas not explored by the original BLiMP, explicitly focusing on dialogue and questions. Evaluation of models using the BLiMP supplement tasks follows a zero-shot approach, similar to BLiMP, where models predict the more probable sequence in each minimal pair, operating under the premise that linguistically acceptable sequences will inherently be more likely than their unacceptable counterparts. These test suites were introduced in~\cite{warstadt-etal-2023-findings} and listed in Table~\ref{tab:blimp_supp_tasks}. Tasks objectives are summarized from~\cite{warstadt-etal-2023-findings}, while fields (tested linguistic field) and examples are extracted from the original dataset\footnote{\url{https://github.com/babylm/evaluation-pipeline}}.

\begin{longtable}{|p{2.5cm}|p{4.5cm}|p{1.6cm}|p{5.5cm}|}
\caption[BLiMP-Supplement Tasks Catalogue]{List of the BLiMP-Supplement tasks included in the BabyLM evaluation pipeline}
\label{tab:blimp_supp_tasks}
\\
\hline
\textbf{Name} & \textbf{Objective} & \textbf{Field} & \textbf{Example} \\ \hline
\endfirsthead

\multicolumn{4}{c}%
{{Table \thetable\ continued from previous page}} \\
\hline
\textbf{Name} & \textbf{Objective} & \textbf{Field} & \textbf{Example} \\ \hline
\endhead

\hline \multicolumn{4}{r}{{Continued on next page}} \\ 
\endfoot

\hline 
\endlastfoot

Hypernyms & Evaluates LMs' understanding of lexical entailment, focusing on hypernym–hyponym relationships. & Semantics & \textbf{Correct:} "If she has a dog, it must be the case that she has a \textit{mammal}." 

\textbf{Incorrect:} "If she has a dog, it must be the case that she has a \textit{chihuahua}." \\
\hline

Subject 

Auxiliary 

Inversion & Explores the subject auxiliary inversion rule essential for forming questions in English. & Syntax & \textbf{Correct}: "Is the novel he \textit{is putting away} from the library?" 

\textbf{Incorrect}: "Is the novel he \textit{putting away is} from the library?" \\

\hline

Turn-Taking & Examines LMs' ability to predict appropriate pronoun usage in dialogues that involve turn-taking. & Syntax & \textbf{Correct}: "A: Should \textit{you} consider asking for another option? B: No, I shouldn't." 

\textbf{Incorrect}: "A: Should \textit{she} consider asking for another option? B: No, I shouldn't." \\
\hline

Question 

Answer 

Congruence (Easy) & Assesses LMs' ability to recognize syntactic constraints imposed by a question on its responses. & Syntax\footnote{We doubt this labeled field. We believe it tests Semantics rather than Syntax.} & \textbf{Correct}: "David: What did you close? Sarah: \textit{The chest}." 

\textbf{Incorrect}: "David: What did you close? Sarah: \textit{David}." \\
\hline

Question 

Answer 

Congruence (Tricky) & Tests LMs' ability to ignore salient distractors and identify congruent answers reflecting the wh-word's intent. & Syntax\footnote{Again, we doubt this labeled field. We believe it tests Semantics rather than Syntax.}  & \textbf{Correct}: "Who studies? \textit{Sarah} studies." 

\textbf{Incorrect}: "Who studies? \textit{Science} studies." \\
\hline

\end{longtable}

\textbf{SuperGLUE}~\cite{super_glue} benchmark serves as a comprehensive toolkit for assessing the capabilities of models on a wide range of Natural Language Understanding (NLU) tasks. This benchmark includes tasks with limited finetuning training data to favor models capable of leveraging general linguistic knowledge across different tasks. In the BabyLM evaluation pipeline, a selection of tasks from SuperGLUE, primarily consisting of text classification challenges, is utilized to gauge model performance. We use the default finetuning hyperparameters as reported in~\cite{warstadt-etal-2023-findings}. Table~\ref{tab:glue_tasks} lists a brief illustration of each task. The objectives are summarized from~\cite{super_glue}, while the examples are extracted from the original dataset\footnote{\url{https://super.gluebenchmark.com/tasks}}.

\begin{longtable}{|p{1.5cm}|p{3.5cm}|p{2.5cm}|p{6.5cm}|}
\caption[SuperGLUE Tasks Catalogue]{List of the SuperGLUE tasks included in the BabyLM evaluation pipeline}
\label{tab:glue_tasks} \\
\hline
\textbf{Name} & \textbf{Objective} & \textbf{Task} & \textbf{Example} \\ \hline
\endfirsthead

\multicolumn{4}{c}%
{{\bfseries Table \thetable\ continued from previous page}} \\
\hline
\textbf{Name} & \textbf{Objective} & \textbf{Field} & \textbf{Example} \\ \hline
\endhead

\hline \multicolumn{4}{r}{{Continued on next page}} \\ 
\endfoot

\hline 
\endlastfoot

BoolQ & QA task where each example is a short passage with a yes/no question. & Question 

Answering & \textbf{Q}: "Is there going to be a season 4 of the border?" \textbf{Passage}: "The Border (TV series) -- The cancellation of The Border was announced by the CBC after three seasons."

\textbf{ Label: No} \\ \hline

CoLA & Consists of English sentences annotated for grammatical acceptability. & Acceptability Judgment & "They caused him to become angry by making him." 

\textbf{Label: Not grammatical} 

"Bill drank from the hose." 

\textbf{Label: Grammatical} \\ \hline

MNLI & Collection of sentence pairs with textual entailment annotations. & Natural Language Inference & "The other men shuffled." - \textbf{Hypothesis:} "The other men were shuffled around." 

\textbf{Label: Entailment} 
\\ \hline

MRPC & Corpus of sentence pairs with annotations for semantic equivalence. & Paraphrase Detection & \textbf{A} "The Embraer jets are scheduled to be delivered by September 2006." \textbf{B} "The Bombardier and Embraer aircraft will be delivered to U.S. Airways by September 2006." 

\textbf{Label: Equivalent} \\ \hline

MultiRC & QA task with a context paragraph, a question, and multiple answers. & Question 

Answering & \textbf{Paragraph:} "Susan ... She called all of her friends. She has five friends..." \textbf{Question:} "How many people did Susan call?" 

\textbf{Answers:} "5" \textbf{Label: Correct}; "6" \textbf{Label: Incorrect} \\ \hline

QNLI & Dataset of question-paragraph pairs, where one sentence contains the answer. & Natural Language Inference & \textbf{Q}: "Article 34 meant states could be responsible for what?" \textbf{Sentence:} "It also means states can be responsible for private actors." 

\textbf{ Label: Positive} \\ \hline

QQP & Collection of question pairs from Quora to determine semantic equivalence. & Paraphrase Detection & \textbf{A} "Why are African-Americans so beautiful?" \textbf{B} "Why are hispanics so beautiful?" 

\textbf{Label: Not equivalent} \\ \hline

RTE & Examples constructed based on news and Wikipedia text for entailment. & Natural Language Inference & \textbf{A} "Hepburn's family will receive the proceeds from the sale." \textbf{B} "Proceeds go to Hepburn's family." 

\textbf{Label: Positive} \\ \hline

SST-2 & Sentences from movie reviews annotated with their sentiment. & Sentiment Classification & "it's a charming and often affecting journey." 

\textbf{Label: Positive} 

"unflinchingly bleak and desperate." 

\textbf{Label: Negative} \\ \hline

WSC & Coreference resolution task with a sentence, a pronoun, and noun phrases. & Commonsense Reasoning & "The path to the lake was blocked, so we couldn't use it." - \textbf{Noun:} "The lake," \textbf{Pronoun:} "it." 

\textbf{Label: Negative} \\ 
\end{longtable}

\textbf{MSGS}~\cite{warstadt-etal-2020-learning} is a classification task set designed to probe the inductive biases in language models. It seeks to determine whether a pretrained model generalizes based on either a linguistic feature (an aspect tied to the structural or functional components of language) or a surface feature (an observable characteristic that does not necessitate deep linguistic analysis). This investigation unfolds in two phases: Initially, control experiments assess the extent to which the model captures linguistic or surface features. Subsequently, preferential experiments explore which type of feature (linguistic or surface) the model is biased towards, thereby identifying the inductive bias learned by the model. After finetuning on artificially designed datasets, control and preferential experiments are performed as ambiguous binary classification tasks. For the original MSGS experiments details, we refer readers to the original publication~\cite{warstadt-etal-2020-learning}. The BabyLM evaluation framework selects three linguistic features (Syntactic Position, Syntactic Category, and Syntactic Construction) and two surface features (Lexical Content and Relative Position). This approach leads to five control experiments, each focusing on one of these features and six preferential experiments that compare a linguistic feature against a surface feature. These features are represented through template-based labeled datasets for finetuning and testing, briefly described in Table~\ref{tab:msgs_tasks}. Dataset descriptions are summarized from~\cite{warstadt-etal-2020-learning}, while the examples are extracted from the original dataset\footnote{\url{https://github.com/nyu-mll/msgs/tree/master/data}}.

\begin{table}[H]
\caption[MSGS Features]{List of the MSGS features included in the BabyLM evaluation pipeline}
\label{tab:msgs_tasks}
\centering
\begin{tabular}{|p{3.9cm}|p{11.4cm}|}
\hline
\textbf{Feature} & \textbf{Dataset Description} \\
\hline
Syntactic Position & Is the main verb in “ing” form? 

\textbf{Positive example:} Cats who eat mice are purring. 

\textbf{Negative Example:} Cats who are eating mice purr. \\
\hline
Syntactic Category & This feature is marked as 1 if the sentence contains an adjective. 

\textbf{Positive example:} Lincoln was tall. 

\textbf{Negative example:} Lincoln was president.
\\
\hline
Syntactic Construction & Marked as 1 if the sentence includes a control construction.

\textbf{Positive example:} Sue is eager to sleep. 

\textbf{Negative example:} Sue is likely to sleep.
\\
\hline
Lexical Content & This feature is 1 if the sentence contains "the". 

\textbf{Positive example:} That cat chased the mouse.

\textbf{Negative example:} That cat chased a mouse
\\
\hline
Relative Position & Marked as 1 when "the" precedes "a" and 0 when "a" precedes "the". 

\textbf{Positive example:} The cat chased a mouse.

\textbf{Negative example:} A cat chased the mouse.
\\
\hline
\end{tabular}
\end{table}

\subsection{Metrics}

For BLiMP and BLiMP-Supplement tasks, evaluation hinges on the basic \textit{Accuracy} metric. SuperGLUE tasks primarily employ the Accuracy metric for assessment, albeit MRPC and QQP diverge by utilizing the \textit{F1 score} due to their training splits' class distribution imbalances. Conversely, CoLA is evaluated through the \textit{Matthews correlation coefficient (MCC)}~\cite{MATTHEWS1975442} as defined in Equation~\ref{eq:mcc}. \(TP\) stands for true positives, \(TN\) stands for true negatives, \(FP\) stands for false positives, and \(FN\) stands for false negatives. MCC is a robust binary classification quality measure where a score of 1 signifies perfect prediction, 0 indicates an average random prediction, and -1 indicates an inverse prediction. All MSGS tasks are measured using MCC. Within MSGS's control experiments, an MCC of 1 indicates the model's successful representation of the examined feature, 0 implies random behavior, and -1 denotes strong disagreement with the feature. For the six preferential experiments, an MCC of 1 indicates a model's preference for linguistic features over surface features, -1 is the opposite, and 0 denotes no clear preference, suggesting random behavior. 

\begin{equation}
\label{eq:mcc}
\text{MCC} = \frac{TP \times TN - FP \times FN}{\sqrt{(TP + FP)  (TP + FN)  (TN + FP)  (TN + FN)}}
\end{equation}

\textbf{Aggregate Scores:} To facilitate comparison across all models participating in the BabyLM shared task, the challenge's authors employ an aggregation of scores, enabling single-value representations for each submitted model. Our experiment uses the same aggregation procedures for a feasible comparison analysis. Scores of BLiMP in addition to scores BLiMP-Supplement are averaged to yield a BLiMP summarized score \(BLIMP_{aggregate}\). Despite the variety of metrics used in the SuperGLUE suite of tasks (Accuracy, F1, and MCC), all SuperGLUE scores are aggregated, resulting in a SuperGLUE summarized score \(SuperGLUE_{aggregate}\), and likewise for MSGS, \(MSGS_{aggregate}\). Ultimately, a Total Aggregate Score is computed as \[Total_{aggregate} = 0.5 \times BLIMP_{aggregate} + 0.3 \times SuperGLUE_{aggregate} + 0.2 \times MSGS_{aggregate}\]

%%%%%%%%%%%%%%%%%%%%%%%%%%%%%%%%%%%%%%%%%%%%%%%%%%%%%%%%%

\section{Results}
\label{evp_results}

This section presents the results of our linguistic evaluation experiment. We assess three variants of our pretrained models across the tasks outlined in Section~\ref{sec:tests}: the StructFormer in its original setup \(SF_{m=0}\), the StructFormer with its parser network positioned after four attention layers \(SF_{m=4}\), and a standard Transformer model \(TF\). These model variants are pretrained on the small version of the BabyLM dataset, employing the subword tokenizer introduced in Section~\ref{sec:babylm}. The \(SF_{m=0}\) variant serves to analyze the linguistic capabilities of the original implementation of the StructFormer architecture as outlined in Section~\ref{sec:implem}. While \(SF_{m=4}\) investigates the impact of our architectural adjustments, as detailed in Section~\ref{sec:in_parser}. The \(TF\) variant offers a baseline to illustrate the significance of embedding StructFormer's syntactic inductive biases into a traditional transformer architecture. Furthermore, we compare the performance of our model variants against 118 other models submitted to the BabyLM challenge. All participating models are pretrained on the same dataset. This comparative analysis aims to gauge our models' performance relative to the other participating architectures designed to enhance cognitively plausible model pretraining. Results of these competing models are sourced from the BabyLM challenge's official repository\footnote{\url{https://github.com/babylm/submissions2023}}. More details on the architecture of all models are provided in the shared task proceedings~\cite{warstadt-etal-2023-findings}. \textbf{We opt for visual plots to present all the results in this section. However, the exact numerical results are also reported in Appendix~\ref{apx:res}.}

\begin{figure}[!htp]
    \centering
    \caption[Distribution of BabyLM models parameters count]{Distribution of BabyLM models parameters count, the lower, medium, and top black lines in the box represent the distribution's 25\%, 50\%, and 75\% quartiles.}
    \label{fig:n_params}
    \includegraphics[width=0.48\linewidth]{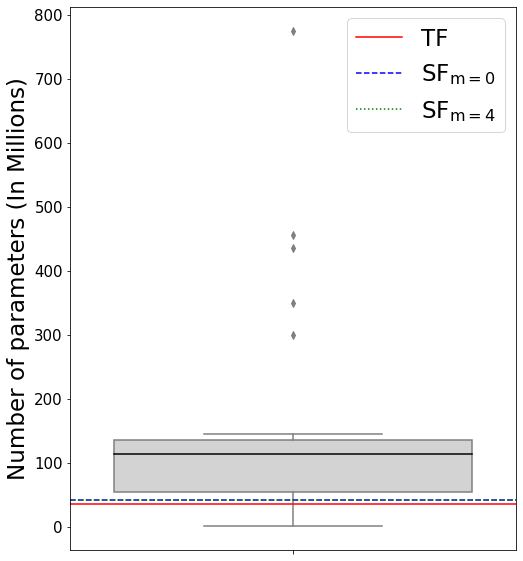} 
\end{figure}

\textbf{Number of Parameters:} The range of the number of learnable parameters across the 118 submitted models is large (ranges from 1.5 million to 774 million) and worth consideration throughout the analysis of the results. Information on the number of parameters for all the models is not publicly available; however, we were able to ascertain these details for 54 of the models. This subset is deemed representative of the broader group of submissions. The distribution of the number of parameters for these models, including our three variant models, is depicted in Figure~\ref{fig:n_params}. The number of parameters of \(SF_{m=0}\) and \(SF_{m=4}\) (41.5M parameters) and \(TF\) (35M parameters) position them in the lower quartile relative to the spectrum of model sizes among the submissions. Thus, our three model variants are relatively smaller than 75\% % of the submitted models in the shared task.

\begin{figure}[!htp]
    \caption[BabyLM models performance on BLiMP tasks]{Distribution of the BabyLM models performance on BLiMP tasks. The lower, medium, and top black lines in the box represent the distribution's 25\%, 50\%, and 75\% quartiles. Tasks in bold are the tasks labeled as "Syntax" in the Field column in Table~\ref{tab:blimp_tasks}. Accuracy is the reported metric.}
    \label{fig:blimp_res}
    \centering
    \includegraphics[width=1\linewidth]{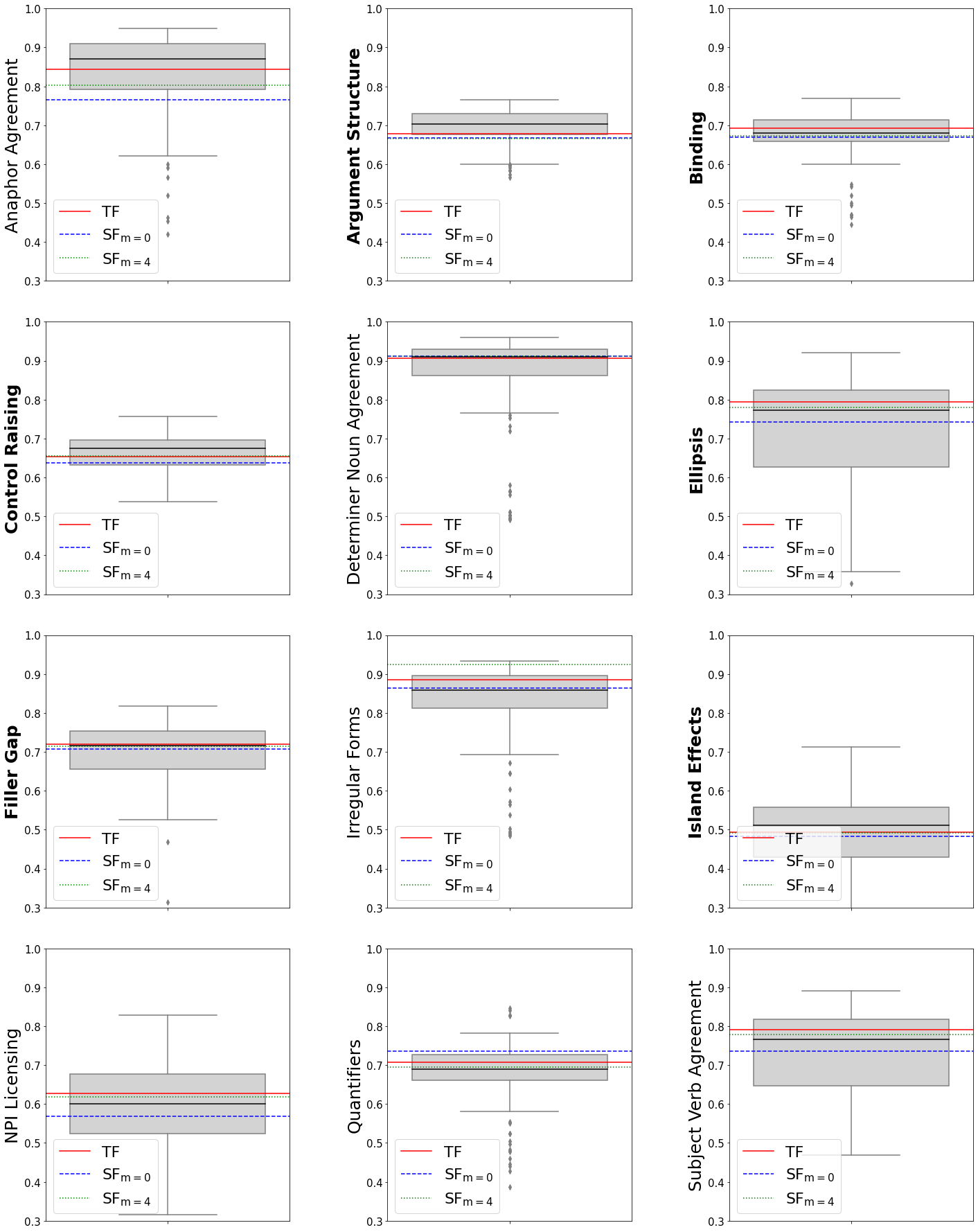}
\end{figure}

\begin{figure}[!htp]
    \caption[BabyLM models performance on BLiMP-Supplement tasks]{Distribution of the BabyLM models performance on BLiMP-Supplement tasks. The lower, medium, and top black lines in the box represent the distribution's 25\%, 50\%, and 75\% quartiles. Tasks in bold are the tasks labeled as "Syntax" in the Field column in Table~\ref{tab:blimp_supp_tasks}. Accuracy is the reported metric.}
    \label{fig:blimp_sup_res}
    \centering
    \includegraphics[width=0.85\linewidth]{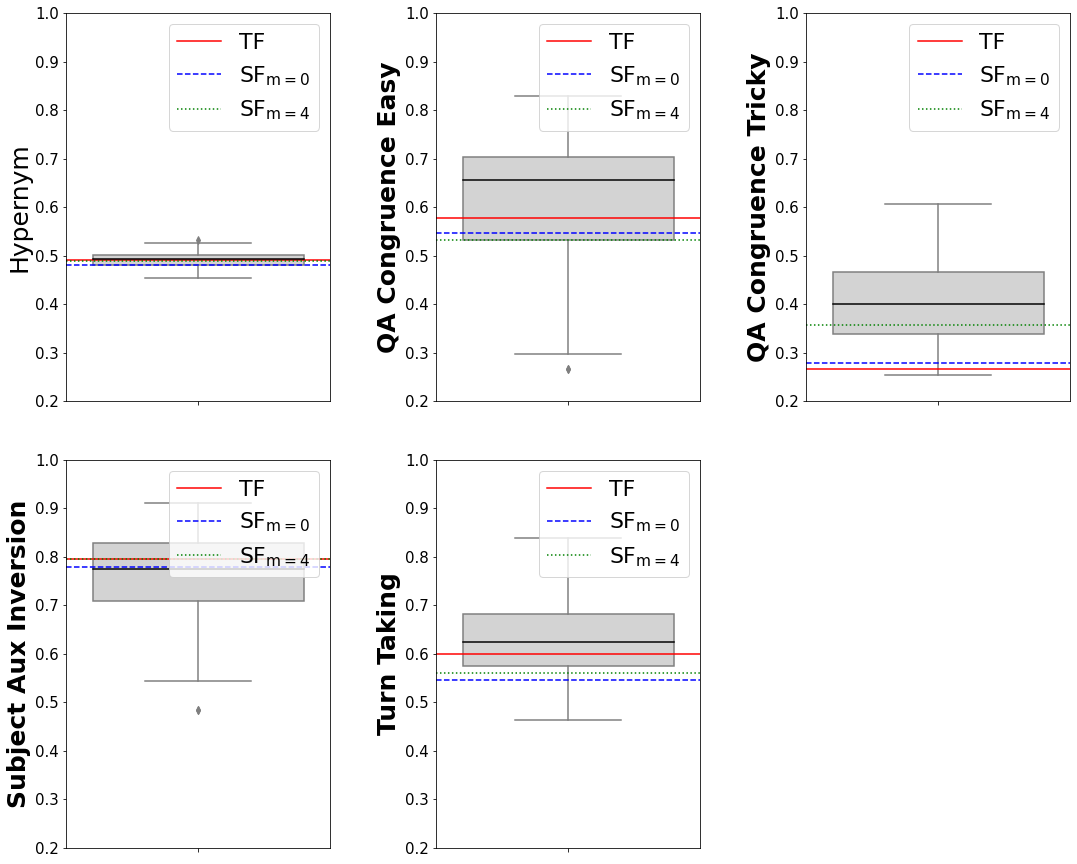}
\end{figure}

\begin{figure}[!htp]
    \caption[BLiMP Aggregate Score for the BabyLM models]{Distribution of the BabyLM models aggregate scores on BLiMP. The lower, medium, and top black lines in the box represent the distribution's 25\%, 50\%, and 75\% quartiles.}
    \label{fig:blimp_ag}
    \centering
    \includegraphics[width=0.4\linewidth]{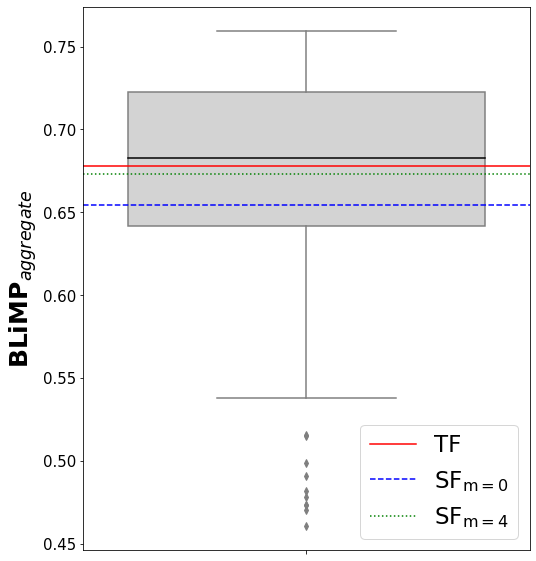}
\end{figure}

\textbf{BLiMP and BLiMP-Supplement:} In Figure~\ref{fig:blimp_res}, we present the zero-shot prediction performance for the BLiMP task set, involving all of the 118 submitted models alongside our three variant models. Similarly, Figure~\ref{fig:blimp_sup_res} displays the outcomes for the BLiMP-Supplement task set, with accuracy as the reported metric in both figures. Contrary to our initial assumptions, the \(SF\) variants do not exhibit a marked advantage over the \(TF\) model within the BLiMP task suite. We anticipated that embedding syntactic inductive bias within the architecture would enhance linguistic capabilities, particularly in tasks probing the syntactic knowledge of the pretrained models (emphasized with bold font type in the figures). However, the performance of our three variants is broadly median when compared against the other participating models across most BLiMP tasks. Interesting results in this context are \(SF_{m=4}\)'s top performance in Irregular Forms compared to the majority of the submitted models and \(SF_{m=0}\) outperformance in Quantifiers. Figure~\ref{fig:blimp_ag} collates the BLiMP aggregate scores for all models. Among our variant models, \(TF\) emerges as the superior, with the parser-intermediate variant \(SF_{m=4}\) showing improved performance relative to \(SF_{m=0}\). Notably, two of our variant models align closely with the median performance across the BLiMP task suite.

\begin{figure}[!htp]
    \caption[BabyLM models performance on SuperGLUE tasks]{Distribution of the BabyLM models performance on SuperGLUE tasks. The lower, medium, and top black lines in the box represent the distribution's 25\%, 50\%, and 75\% quartiles. Tasks in bold are the tasks labeled as "Syntax" in the Field column in Table~\ref{tab:glue_tasks}. Accuracy is the used metric in all plots except otherwise noted. The range for CoLA (MCC) is (-1,1).}
    \label{fig:glue_res}
    \centering
    \includegraphics[width=1\linewidth]{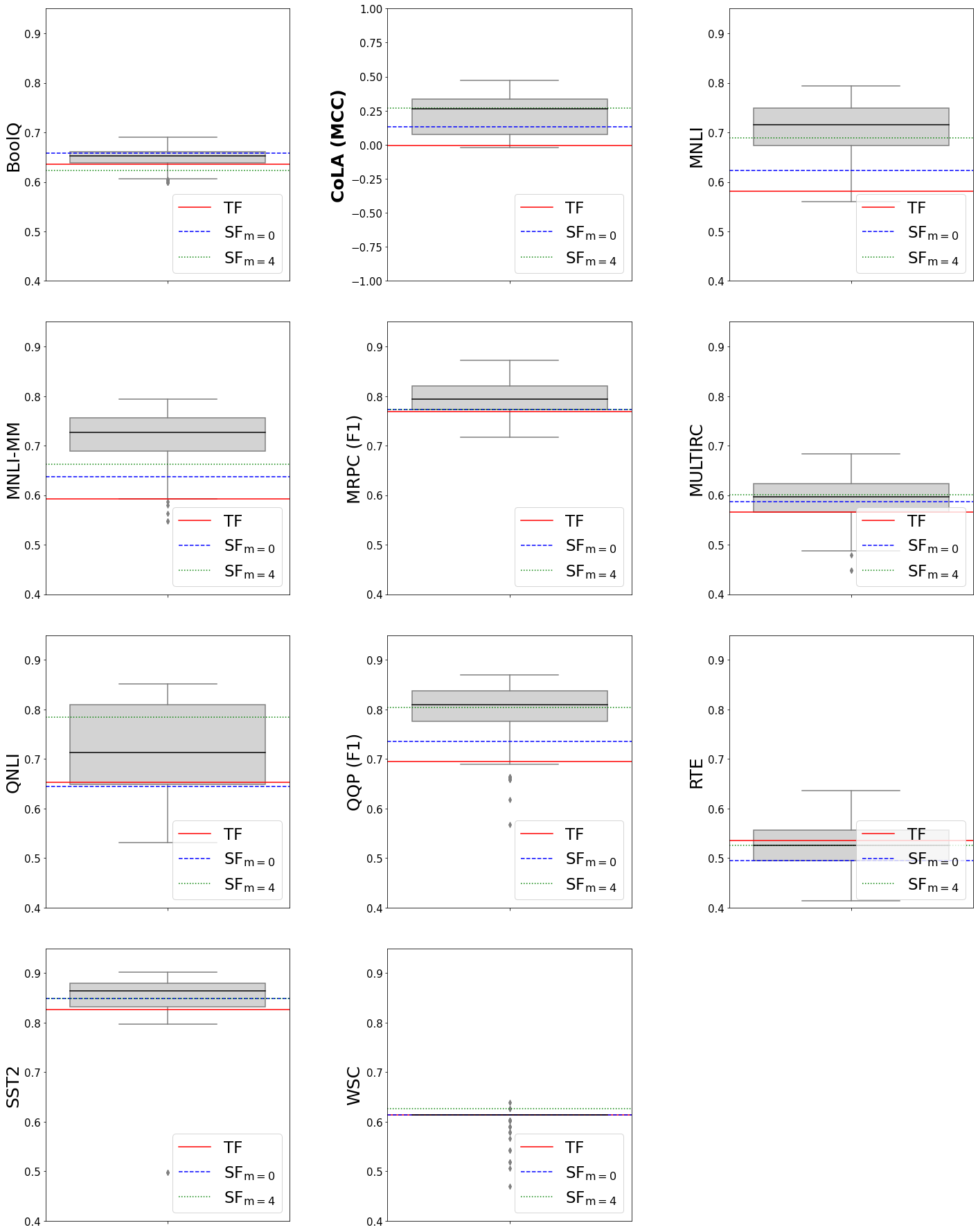}
\end{figure}

\begin{figure}[!htp]
    \caption[SuperGLUE Aggregate Score for the BabyLM models]{Distribution of the BabyLM models aggregate scores on SuperGLUE. The lower, medium, and top black lines in the box represent the distribution's 25\%, 50\%, and 75\% quartiles.}
    \label{fig:glue_ag}
    \centering
    \includegraphics[width=0.4\linewidth]{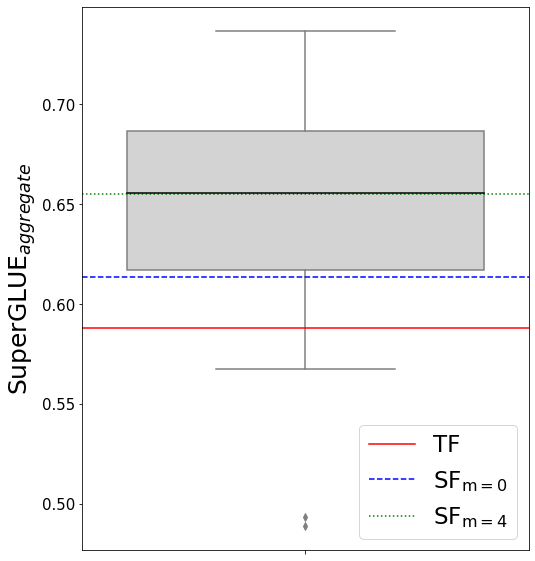}
\end{figure}

\textbf{SuperGLUE:} In Figure~\ref{fig:glue_res}, we present the SuperGLUE task performance of all models, highlighting the superior performance of the \(SF\) variants over the \(TF\) model, especially in tasks like CoLA, with an exception at the RTE dataset. Figure~\ref{fig:glue_ag} shows the SuperGLUE aggregate scores, where \(SF_{m=4}\) notably excels, closely matching the median of all submissions, and \(SF_{m=0}\) outperforms \(TF\), while \(TF\) scores are below the scores of 75\% of the submissions.

\begin{figure}[!htp]
    \caption[BabyLM models results on MSGS Control Experiments]{Distribution of the BabyLM models results on MSGS Control Experiments. The lower, medium, and top black lines in the box represent the distribution's 25\%, 50\%, and 75\% quartiles. Features in bold are the linguistic features. The measured metric in these distributions is MCC.}
    \label{fig:msgs_cn_res}
    \centering
    \includegraphics[width=0.9\linewidth]{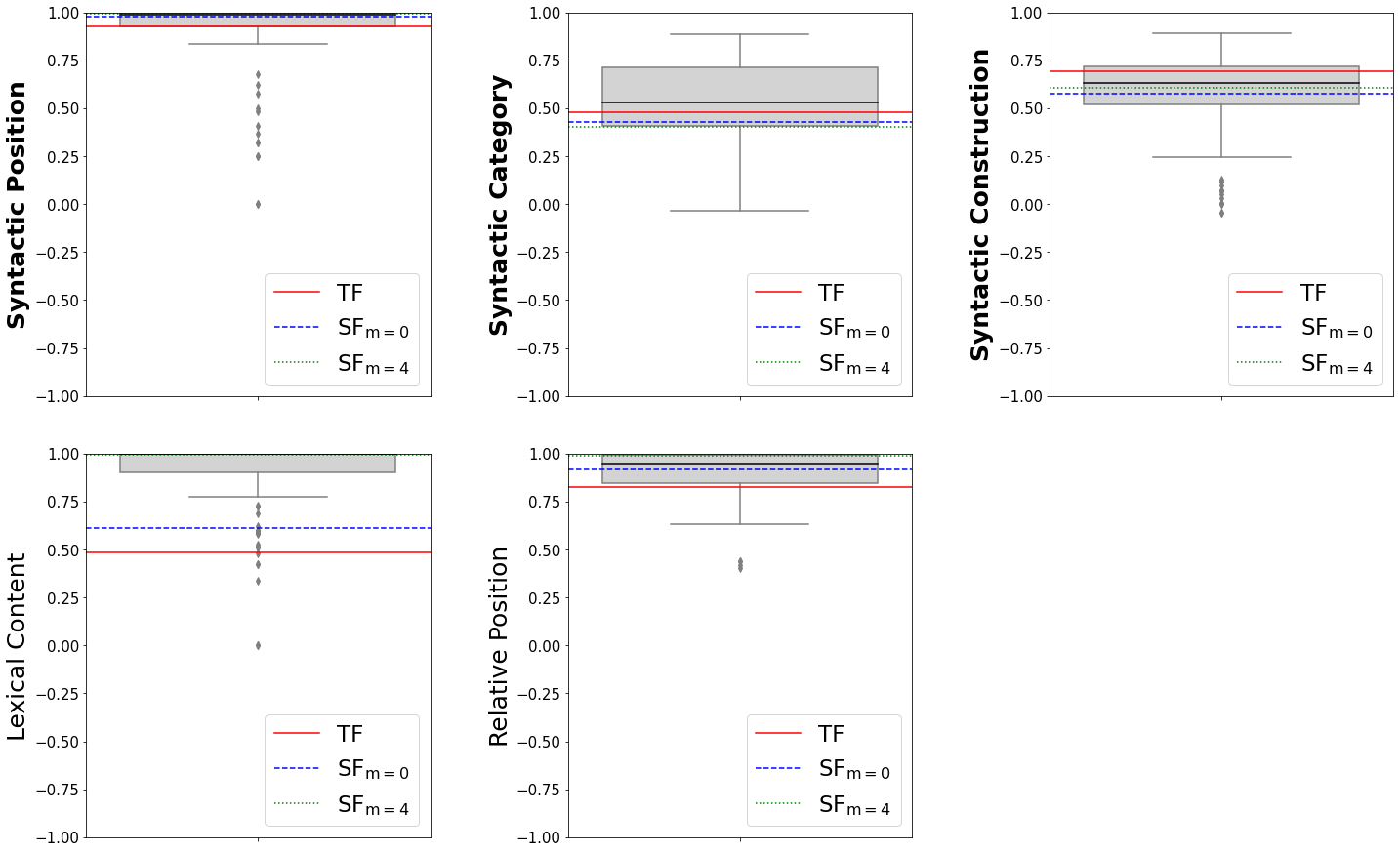}
\end{figure}

\begin{figure}[!htp]
    \caption[BabyLM models results on MSGS Preferential Experiments]{Distribution of the BabyLM models results on MSGS Preferential Experiments. The lower, medium, and top black lines in the box represent the distribution's 25\%, 50\%, and 75\% quartiles. The measured metric in these distributions is MCC.}
    \label{fig:msgs_pref_res}
    \centering
    \includegraphics[width=0.9\linewidth]{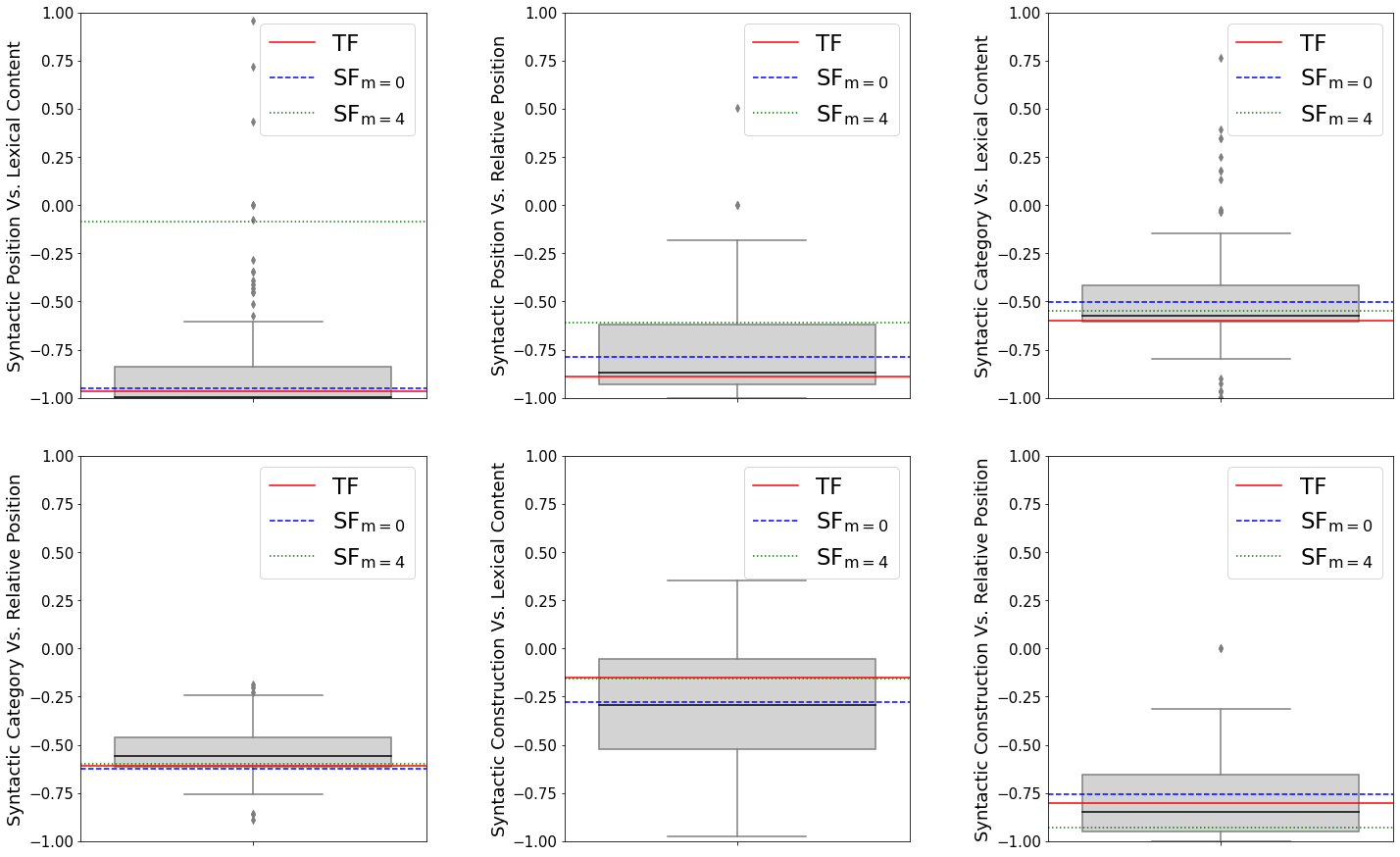}
\end{figure}

\begin{figure}[!htp]
    \caption[MSGS Aggregate Score for the BabyLM models]{Distribution of the BabyLM models aggregate scores on MSGS. The lower, medium, and top black lines in the box represent the distribution's 25\%, 50\%, and 75\% quartiles.}
    \label{fig:msgs_ag}
    \centering
    \includegraphics[width=0.4\linewidth]{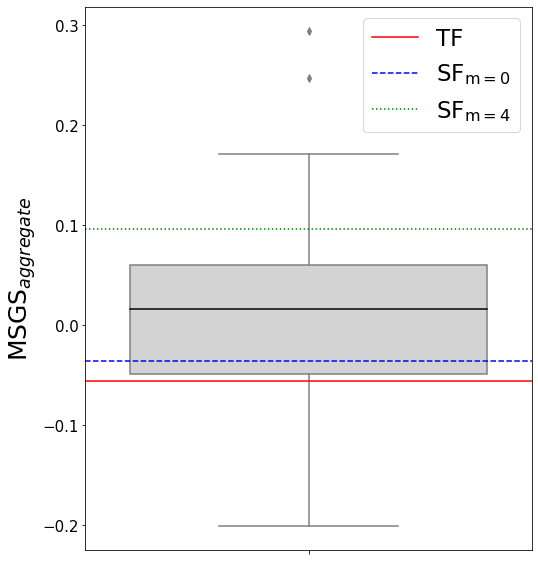}
\end{figure}

\textbf{MSGS:} In Figure~\ref{fig:msgs_cn_res}, we present the results of the control experiments of the MSGS tasks, encompassing all submitted models alongside our three variant models. Similarly, Figure~\ref{fig:msgs_pref_res} displays the outcomes from the preferential experiments of the MSGS tasks. These results were obtained after models were finetuned on the designated task datasets, with MCC as the evaluation metric. In the control experiments, the \(SF\) variants demonstrate a marginally stronger correlation with the feature of Syntactic Position in comparison to the \(TF\) model. Contrary to expectations, for Syntactic Category and Syntactic Construction features, the \(TF\) model exhibits a stronger correlation than the \(SF\) variants. Interestingly, the \(SF\) variants show a stronger correlation with both surface features (Lexical Content and Relative Position) than the \(TF\) model. \(SF_{m=4}\) outperforms \(SF_{m=0}\) in correlation strength for surface features, and the \(TF\) model shows a weaker correlation in comparison to the broader participant group. Within the preferential experiments, \(SF_{m=4}\) displays ambiguity in preference for Syntactic Position relative to Lexical content (observed as an outlier in the distribution of performances), albeit a moderate preference for Relative Position. Conversely, both \(SF_{m=0}\) and \(TF\) demonstrated a pronounced preference for both surface features over Syntactic Position. In evaluating Syntactic Category and Construction, all three variants mirrored the general participant trend, showing a moderate preference for surface features, with \(SF_{m=4}\) exhibiting a particularly strong inclination towards Relative Position against Syntactic Construction compared to most models. Figure~\ref{fig:msgs_ag} aggregates the MSGS scores for all participating models, including our variants. Here, \(SF_{m=4}\) is noted for a slightly positive correlation against most participating models. However, interpreting this aggregate score is challenging due to its amalgamation of outcomes from both control and preferential experiments across all examined linguistic and surface features. Despite these complexities, this score is reported as part of the comprehensive BabyLM evaluation framework.

\begin{figure}[!htp]
    \caption[Total Aggregate Score for the BabyLM models]{Distribution of the BabyLM models total aggregate score. The lower, medium, and top black lines in the box represent the distribution's 25\%, 50\%, and 75\% quartiles.}
    \label{fig:total_ag}
    \centering
    \includegraphics[width=0.4\linewidth]{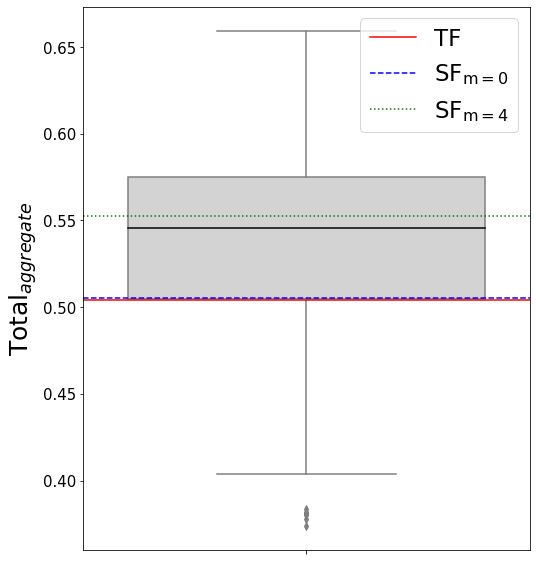}
\end{figure}

\textbf{Total Score:} In Figure~\ref{fig:total_ag}, the aggregate scores for all models are depicted. Among the models we evaluated, \(SF_{m=4}\) stands out, securing a score that surpasses over half of the participating models and beating our other two variants by 5 points. \(SF_{m=0}\) and \(TF\) attain comparable aggregate scores, outperforming merely 25\% of the models submitted.

%%%%%%%%%%%%%%%%%%%%%%%%%%%%%%%%%%%%%%%%%%%%%%%%%%%%%%%%%%%%%%%%%%%%%%%%%%%%%%%%%%%

\section{Discussion}
\label{evp_disc}

In this section, we analyze the results of the linguistic evaluation, focusing on the impact of syntactic inductive bias across various tasks. We also conclude the effect of the architecture modification we introduced in~\ref{sec:in_parser}. Through this critique, we seek to contribute to the ongoing discourse on enhancing the pretrained models' capabilities through architectural innovations and the concept of shifting the inductive bias of language models towards a linguistic aspect.

In Chapters \ref{ch:sf} and \ref{ch:sf_dev}, we demonstrated StructFormer's ability to induce syntactic trees and compared these induced trees against reference trees. We also confirmed the model's effectiveness in masked language modeling through perplexity measurements. StructFormer's authors suggest in~\cite{shen-etal-2021-structformer} that incorporating syntactic inductive bias could enhance the transformer model's performance. However, the specific areas of improvement were not detailed. In our study, we investigate whether StructFormer outperforms a standard transformer across a diverse range of linguistic tasks. The critical inquiry centers on whether StructFormer's design, which includes a syntactic inductive bias, impacts its linguistic capabilities, particularly in tasks demanding syntactic understanding.

The BLiMP task set provides the most direct response to our primary inquiry due to its apparent relation to syntactic features. From the results, it becomes clear that StructFormer does not outperform the vanilla transformer in the BLiMP tasks, with the notable exception being the "Irregular Forms" task where \(SF_{m=4}\) demonstrates superior performance. Irregular Forms examines morphological aspects. This outcome suggests that the ability of a model to generate meaningful syntactic trees, even with an embedded syntactic inductive bias, does not necessarily translate to enhanced performance on tasks demanding syntactic comprehension in a zero-shot context as in the BLiMP tasks.

Contrary to the results of BLiMP, StructFormer models demonstrate superior performance over the vanilla transformer across most SuperGLUE tasks. These tasks are designed to evaluate semantic understanding capabilities, except the CoLA task, which focuses on grammatical accuracy. StructFormer's architecture incorporates a hierarchical inductive bias that is fundamentally based on the relationships between tokens. However, the success of StructFormer in SuperGLUE tasks suggests that its inductive bias may extend beyond purely syntactic domains. It can be the case that it leans towards semantic inclinations, especially since semantic structures can also be hierarchical and rely on word dependencies, akin to syntax. We suggest a hypothesis that the inductive bias in StructFormer pushes towards a mixture of semantic and syntactic features. This hypothesis is supported by the fact that StructFormer's dependency mechanism, as demonstrated in Section~\ref{dep_funct}, primarily outlines how words converge to form larger units, indicative of constituent building, and how they interrelate, mirroring dependency relations. This mechanism, however, is not exclusive to syntax and could equally apply to semantic features. This hypothesis necessitates further investigation to decode the specific nature of the bias within StructFormer. \cite{arps-etal-2022-probing} proposes a perturbed dataset that separates the effect of semantics when evaluating a LM for syntactic features. We suggest that failing to induce syntactic trees of significantly lower quality on this dataset can further support the abovementioned hypothesis.

Beyond BLiMP, the MSGS tasks serve as another critical gauge for evaluating syntax representation within our pretrained models. As noted in~\cite{warstadt-etal-2020-learning}, MSGS task performance is significantly influenced by the size of the pretraining dataset. This presents an opportunity to assess whether the hierarchical inductive bias in transformer models can mitigate the limitations imposed by dataset size. StructFormer demonstrates an improved correlation with the Syntactic Position feature among three linguistic features in the control experiments but shows a weaker correlation with Syntactic Category and Construction, which are known to correlate poorly with models pretrained on datasets smaller than 100M words. Remarkably, StructFormer exhibits a strong correlation with both surface features, suggesting that the model's dependency relations during training could be leveraged more effectively to represent surface features compared to a conventional transformer. Despite this, like all the models participating in the shared task, StructFormer does not show a preference for linguistic features over surface features in the preferential experiments, except for a few outliers favoring Syntactic Position. These outlier models merit further exploration.

Drawing on prior insights regarding the syntactic representation within transformer models at specific layer positions, we explore whether repositioning the StructFormer's parser network after four standard attention layers, as detailed in~\ref{sec:in_parser}, would enhance its linguistic capabilities beyond its original configuration. Although this architectural adjustment does not yield marked improvements in the tree induction benchmark as shown in Section~\ref{sec:in_parser}, the modified StructFormer variant \(SF_{m=4}\) outperforms the original \(SF_{m=0}\) across most of the tasks within our evaluation framework. This finding aligns with prior assertions regarding the critical role of mid-layers for syntax representation~\cite{liu-etal-2019-linguistic,vig-belinkov-2019-analyzing,arps-etal-2022-probing}. This result also underscores the necessity for a more comprehensive analysis of the induced syntactic structures from the in-parser configuration, potentially comparing the induced trees against diverse syntactic annotation standards.

Adopting an overall view to evaluate our participation in the BabyLM challenge, the StructFormer model with the parser network positioned after four layers \(SF_{m=4}\) emerged as our top-performing submission. Notably, it surpassed the performance of over half the challenge's entries despite its relatively smaller model size than other submissions. This achievement suggests a potential for further enhancement through hyperparameter tuning and possibly adopting a curriculum learning strategy\footnote{A pretraining technique that orders the training samples in a sequence from easy to hard before introducing them to the model. This approach enables the model to initially learn from simpler features before converging on harder features~\cite{curriculum}.}, indicating that compact models can compete effectively in complex NLP tasks.

\chapter{Conclusion}
\label{ch:conc}

In this thesis, we present an exploratory study on linguistic structure induction from language models, aiming to understand and analyze the problem, the techniques utilized to solve it, and the challenges encountered. The scope is set to focus specifically on producing syntactic constituency and dependency trees from transformer-based language models without any kind of supervision.

We start by conducting an overall review of the literature, examining the evolution of methods used to address the problem, starting from early methods before the introduction of neural networks and arriving at the most recent state where transformer-based methods dominate the scene of addressing language problems including the particular problem of linguistic structure induction. The literature review highlights two main implicit objectives in the recent studies: 1) developing a representation of syntactic trees that aligns well with neural-based models, and 2) incorporating a syntactic inductive bias into the pretraining of language models. Based on these insights, our attention is drawn towards a recent approach that introduces a numerical representation (syntactic distance) for representing syntactic trees. A unique promising method (StructFormer~\cite{shen-etal-2021-structformer}) is selected as an example of a method that utilizes the syntactic distance representation to incorporate an inductive bias in a transformer-based model to induce both constituency and dependency trees from a pretrained model. 

We present a comprehensive theoretical and analytical study on StructFormer, breaking down its model architecture and examining each component in the architecture. We reproduce the implementation and evaluation procedure of the model, reporting both the original and reproduced results. We also try to understand how the model induces syntactic trees by analyzing an input/output scenario from a pretrained model supported by detailed illustrations of the entire process.

StructFormer's implementation and results show some limitations: the restrictions imposed by using a basic word-based tokenization algorithm, pretraining on limited small datasets, and the absence of a deep linguistic evaluation for the pretrained LMs. Towards extending the StructFormer's original implementation, we propose a potential improvement on the model architecture based on a previous finding about the importance of middle layers in transformer models for capturing linguistic knowledge~\cite{liu-etal-2019-linguistic,vig-belinkov-2019-analyzing,arps-etal-2022-probing}. We examine the impact of positioning the parser network in StructFormer at different positions within the attention blocks of a transformer model. We also attempt to address the observed limitation of tokenization in the problem of syntactic structure induction from language models by implementing a subword-based tokenization paradigm in pretraining and tree induction from StructFormer. By solving this limitation, we can pretrain StructFormer on a larger dataset that represents domains not usually used in syntax structure induction experiments. 

To test the linguistic capabilities of StructFormer and our proposed extensions, we participate in an NLP challenge (BabyLM Shared Task~\cite{warstadt-etal-2023-findings}) that aims to enhance cognitively plausible language models' pretraining. Three model variants representing the experimented concepts in the thesis are submitted to the challenge. The evaluation pipeline of the shared task is a comprehensive, thorough benchmarking framework, seen as an opportunity to verify the validity of several hypotheses and assumptions about the introduced architectures in the thesis. Hence, an introduction of the task's evaluation pipeline is presented, followed by a graphical report of the evaluation results that eventually support the insights and findings claimed. We have published a research paper~\cite{omar-2023} documenting our participation in this challenge.

\section{Findings}

For the sake of a clear closure, the findings of this thesis are summarized as follows:

\begin{enumerate}
    \item StructFormer is a masked language model that induces constituency and dependency structures that partially align with long-established linguistic theories of syntax. StructFormer is empirically demonstrated to be a robust architecture by performing fairly well in various settings such as:
    \begin{itemize}
        \item Utilizing a word frequency-based tokenizer, as shown in Section~\ref{sec:implem}.
        \item Employing a subword-based tokenizer, as shown in Section~\ref{sec:subword}.
        \item Being pretrained on datasets of varying sizes and from different domains, as shown in Section~\ref{sec:babylm}.
        \item Having its induced trees evaluated against either human-annotated or automatically parsed trees, as shown in Sections~\ref{sec:eval} and~\ref{sec:babylm}.
    \end{itemize}
     
    \item Positioning the parser network in StructFormer around the middle attention layers shows an improvement in the MLM objective by 3 points measured by the adapted MLM perplexity~\ref{eq:ppl_mlm}. However, the induced syntactic trees demonstrate inconsistent alignments across different runs per the same setting. Some runs achieve the best-ever parsing performance observed in all experiments, while others, under the same configuration, display some of the poorest parsing results. This is shown by the results in Section~\ref{sec:in_parser}.
    
    \item The induction of syntactic structures from language models using a subword-based approach is possible and empirically shown to function without significant issues. However, the absence of benchmarking references for subword-based syntactic structure induction may limit future research in this direction. This is shown by the results in Section~\ref{sec:subword}.
    
    \item It is not guaranteed that a model capable of producing meaningful syntactic trees will outperform a model that does not produce any trees on tasks requiring knowledge of syntactic structures of language. This is shown by the results in Section~\ref{evp_results}.
    
    \item While the inductive bias in StructFormer aids in inducing meaningful syntactic trees, its contribution to the model's ability to generalize language modeling based on structured syntax features remains unclear. Nonetheless, it may assist the model in generalizing based on structured semantic features or even surface relations between tokens. This is shown by the results in Section~\ref{evp_results}.
    
    \item The in-parser model variant consistently outperforms the original StructFormer implementation in most BLiMP, SuperGLUE, and MSGS tasks. It also shows an overall superior performance compared to the vanilla transformer model and more than 50\% of the models participating in the BabyLM shared task. This is shown by the results in Section~\ref{evp_results}.

\end{enumerate}

\section{Future Work}

Throughout the thesis, several potential tasks have emerged as inspirations for future work, briefly listed as follows:

\begin{enumerate}
    \item The need becomes apparent for an updated comparative study on methods that perform unsupervised syntax structure induction, with evaluations of the methods' performance based on standardized benchmarks and under identical settings.

    \item The efficacy of the benchmarking tools used, such as the training/evaluation datasets and the evaluation metrics (UF1, UAS), is questioned and hence worth investigation, and the introduction of new tools is considered if efficacy could not be demonstrated.

    \item Exploration of subword-based structure induction across other methods and the establishment of a subword-based structure induction benchmark is encouraged. An adaptation is required to reform word-based dependency trees into subword-based trees.

    \item The possibility that StructFormer's inductive bias is skewed towards semantic or surface features rather than syntactic features is to be empirically verified. Additionally, comparing the induced trees from StructFormer against semantic trees is suggested to confirm the hypothesis that the dependency mechanism in StructFormer is not exclusive to syntax. 

    \item Participation in future BabyLM challenges is possible using the best-performing architecture from recent participation \(SF_{m=4}\) after conducting hyperparameter tuning and possibly applying a curriculum learning approach during pretraining.

    \item Extending StructFormer to languages other than English is possible after introducing the subword-based paradigm in pretraining and evaluating StructFormer. 
\end{enumerate}

% Auskommentieren, falls kein Anhang notwendig
% !TeX root = master.tex
% !TeX spellcheck = de_DE

\appendix
\chapter{Appendix}
\label{sec:anhang}
\raggedbottom

\section{BabyLM Dataset}
\label{apx:babylm}

The BabyLM authors compiled and distributed a pretraining corpus inspired by the input that children receive, aimed at providing a developmentally cognitive plausible data set for language model training. The dataset is distinctively characterized by two principal properties: its size and its domain. These characteristics are meticulously controlled to ensure the data is more aligned with developmental plausibility than is typical for language model pretraining datasets. A scaled-down version of this dataset is also provided, which is approximately 10\% the size of the full corpus. The corpus comprises fewer than 100 million words, while the smaller version of this corpus contains fewer than 10 million words. The design of the corpus takes into account findings~\cite{childs} that suggest that children are exposed to approximately 2 million to 7 million words per year. By selecting the onset of adolescence (age 12) as a cutoff point, the total word exposure would theoretically range from 24 million to 84 million words. For the purposes of this dataset, the upper limit is rounded to 100 million words. The small version corresponds to the amount of input in the first two to five years of a child's development. The corpus predominantly features transcribed or scripted speech, accounting for approximately 56\% of the total content. This decision reflects the observation that speech constitutes a significant portion of a child's linguistic input, a trend that gradually shifts towards written media as the child ages. This approach contrasts with conventional language model (LM) training datasets, which primarily consist of texts intended for reading and are often edited. The emphasis on speech is crucial for exploring grammar acquisition, considering the variation in the frequency of certain grammatical constructions between spoken and written language, as noted by~\cite{Biber_1988}. Additionally, about 40\% of the pretraining corpus is derived from sources tailored for or suitable to children, including child-directed speech, children's literature, educational materials, and simplified English content.

\newpage

\section{Linguistic Evaluation Results}
\label{apx:res}

\begin{longtable}{l|cccc}
\caption[BabyLM Evaluation Results]{The BabyLM evaluation results for all the 39 tasks. 12 BLiMP tasks, 5 BLiMP-Supplement tasks, 11 SuperGLUE tasks, 5 MSGS control experiments, and 6 MSGS preferential experiments. Three model variants are evaluated. Vanilla Transformer baseline \(TF\), StructFormer with parser network placed in its original position \(SF_{m=0}\), and StructFormer with parser network placed after 4 transformer blocks \(SF_{m=4}\). The numerical values are multiplied by \(100\) to represent percentages for the sake of readability.}\\
\toprule
\textbf{Metric} & \textbf{\(TF\)} & \textbf{\(SF_{m=0}\)} & \textbf{\(SF_{m=4}\)} & \textbf{\(BabyLM_{median}\)} \\
\midrule
\endfirsthead

\multicolumn{5}{c}%
{\tablename\ \thetable\ -- \textit{Continued from previous page}} \\
\toprule
\textbf{Metric} & \textbf{\(TF\)} & \textbf{\(SF_{m=0}\)} & \textbf{\(SF_{m=4}\)} & \textbf{\(BabyLM_{median}\)}  \\
\midrule
\endhead

\hline \multicolumn{5}{r}{\textit{Continued on next page}} \\
\endfoot

\bottomrule
\endlastfoot
\multicolumn{5}{c}{\textbf{BLiMP}} \\
\midrule
Anaphor Agreement & 84.3 & 76.64 & 80.37 & \textbf{87.12} \\
Argument Structure & 67.76 & 66.8 & 66.66 & \textbf{70.39} \\
Binding & \textbf{69.23} & 66.87 & 67.28 & 68.02 \\
Control Raising & 65.33 & 63.72 & 65.58 & \textbf{67.45} \\
Determiner Noun Agreement & 90.57 & 91.12 & \textbf{91.2} & 91.04 \\
Ellipsis & \textbf{79.5} & 74.19 & 78.06 & 77.37 \\
Filler Gap & \textbf{71.97} & 70.7 & 71.43 & 71.62 \\
Irregular Forms & 88.55 & 86.46 & \textbf{92.37} & 85.8 \\
Island Effects & 49.44 & 48.24 & 49.18 & \textbf{51.2} \\
NPI Licensing & \textbf{62.72} & 56.83 & 61.78 & 60.14 \\
Quantifiers & 70.74 & \textbf{73.62} & 69.58 & 68.96 \\
Subject Verb Agreement & \textbf{79.1} & 73.66 & 77.83 & 76.53 \\
\hline
\multicolumn{5}{c}{\textbf{BLiMP Supplement}} \\
\midrule
Hypernym & 49.19 & 48.14 & 48.95 & \textbf{49.3} \\
QA Congruence Easy & 57.81 & 54.69 & 53.12 & \textbf{65.62} \\
QA Congruence Tricky & 26.67 & 27.88 & 35.76 & \textbf{40.0} \\
Subject Auxiliary Inversion & \textbf{79.53} & 77.9 & 79.41 & 77.41 \\
Turn Taking & 60.0 & 54.64 & 56.07 & \textbf{62.5} \\
\hline
\(BLiMP_{aggregate}\) & 67.79 & 65.42 & 67.33 & \textbf{68.27} \\
\hline
\multicolumn{5}{c}{\textbf{SuperGLUE}} \\
\midrule
BoolQ & 63.62 & \textbf{65.84} & 62.38 & 65.28 \\
CoLA & -0.35 & 13.08 & \textbf{26.71} & 26.26 \\
MNLI & 58.11 & 62.31 & 68.85 & \textbf{71.56} \\
MNLI-mm & 59.26 & 63.69 & 66.31 & \textbf{72.69} \\
MRPC & 76.92 & 77.32 & 77.34 & \textbf{79.42} \\
MultiRC & 56.63 & 58.71 & \textbf{60.13} & 59.69 \\
QNLI & 65.31 & 64.48 & \textbf{78.39} & 71.26 \\
QQP & 69.49 & 73.5 & 80.36 & \textbf{80.97} \\
RTE & \textbf{53.54} & 49.49 & 52.53 & 52.53 \\
SST-2 & 82.68 & 84.84 & 84.84 & \textbf{86.42} \\
WSC & 61.45 & 61.45 & \textbf{62.65} & 61.45 \\
\hline
\(SuperGLUE_{aggregate}\) & 58.79 & 61.34 & \textbf{65.54} & \textbf{65.55} \\
\hline
\multicolumn{5}{c}{\textbf{MSGS}} \\
\midrule
Syntactic Position (Control) & 92.81 & 97.71 & \textbf{99.3} & 98.99 \\
Syntactic Category (Control) & 47.86 & 42.94 & 40.11 & \textbf{53.17} \\
Syntactic Construction (Control) & \textbf{69.3} & 57.7 & 60.7 & 63.03 \\
Lexical Content (Control) & 48.21 & 61.17 & 99.32 & \textbf{99.91} \\
Relative Position (Control) & 82.34 & 91.65 & \textbf{98.85} & 94.66 \\
\hline
Syntactic Construction Vs. Lexical Content & -15.2 & -28.01 & -15.81 & -29.53 \\
Syntactic Construction Vs. Relative Position & -80.16 & -75.66 & -93.01 & -84.63 \\
Syntactic Position Vs. Lexical Content & -96.45 & -94.96 & -8.48 & -99.65 \\
Syntactic Position Vs. Relative Position & -89.0 & -78.74 & -60.82 & -87.08 \\
Syntactic Category Vs. Lexical Content & -59.84 & -50.31 & -54.74 & -57.54 \\
Syntactic Category Vs. Relative Position & -60.74 & -62.49 & -59.7 & -55.68 \\
\hline
\(MSGS_{aggregate}\) & -5.53 & -3.55 & \textbf{9.61} & 1.67 \\
\hline
\midrule
\(Total_{aggregate}\) & 50.42 & 50.4 & \textbf{55.24} & 54.57 \\
\end{longtable}

\backmatter

\printbibliography

\cleardoublepage

\printindex

% do not change this file
\begin{otherlanguage}{ngerman}

\chapter*{Ehrenwörtliche Erklärung}

Hiermit versichere ich, die vorliegende Masterarbeit selbstständig verfasst und keine anderen als die angegebenen Quellen und Hilfsmittel benutzt zu haben.
Alle Stellen, die aus den Quellen entnommen wurden, sind als solche kenntlich gemacht worden.
Diese Arbeit hat in gleicher oder ähnlicher Form noch keiner Prüfungsbehörde vorgelegen.

\vspace{3cm}
\noindent Düsseldorf, \thesissubmissionday{}. \DTMmonthname{\thesissubmissionmonth} \thesissubmissionyear{} \hfill \thesisauthor{}

\vspace{1em}

\hfill  \includegraphics[width=0.18\linewidth]{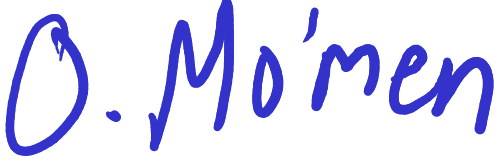}

\end{otherlanguage}

\cleardoublepage

\end{document}